\documentclass[pdflatex,sn-basic]{sn-jnl}

\usepackage{graphicx}%
\usepackage{multirow}%
\usepackage{amsmath,amssymb,amsfonts}%
\usepackage{amsthm}%
\usepackage{mathrsfs}%
\usepackage[title]{appendix}%
\usepackage{xcolor}%
\usepackage{textcomp}%
\usepackage{manyfoot}%
\usepackage{booktabs}%
\usepackage{algorithm}%
\usepackage{algorithmicx}%
\usepackage{algpseudocode}%
\usepackage{listings}%

\usepackage{booktabs}
\usepackage{tabularx}
\usepackage{array}
\usepackage{ragged2e}
\usepackage{colortbl}

\newcolumntype{L}[1]{>{\RaggedRight\arraybackslash}p{#1}}
\newcolumntype{C}[1]{>{\Centering\arraybackslash}p{#1}}
\newcolumntype{Y}{>{\RaggedRight\arraybackslash}X}

\newcolumntype{A}{>{\centering\arraybackslash}X}
\newcolumntype{B}{>{\raggedright\arraybackslash}X}

\newcolumntype{N}[1]{>{\raggedright\arraybackslash\hyphenpenalty=10000\exhyphenpenalty=10000}p{#1}}
\newcommand{\tablegroupgray}{\rowcolor{gray!15}}

\usepackage{placeins}

\usepackage{pifont}

\theoremstyle{thmstyleone}%
\theoremstyle{thmstyletwo}%

\theoremstyle{thmstylethree}%

\begin{document}

\title[DLMs for Mobile Edge Agentic AI]{Diffusion Language Models for Mobile Edge Agentic AI: Foundations, Applications, and Challenges}

% \author[1]{\fnm{Chenqi} \sur{Li}}\email{iauthor@gmail.com}

% \author[1]{\fnm{Minghui} \sur{Min}}\email{iiauthor@gmail.com}

% \author[2]{\fnm{Dusit} \sur{Niyato}}\email{iiiauthor@gmail.com}


% \affil*[1]{\orgdiv{Department}, \orgname{Organization}, \orgaddress{\street{Street}, \city{City}, \postcode{100190}, \state{State}, \country{Country}}}

% \affil[2]{\orgdiv{Department}, \orgname{Organization}, \orgaddress{\street{Street}, \city{City}, \postcode{10587}, \state{State}, \country{Country}}}

% \affil[3]{\orgdiv{Department}, \orgname{Organization}, \orgaddress{\street{Street}, \city{City}, \postcode{610101}, \state{State}, \country{Country}}}

\author[1]{\fnm{Chenqi} \sur{Li}}
\email{chenqi.li@cumt.edu.cn}

\author*[1]{\fnm{Minghui} \sur{Min}}
\email{minmh@cumt.edu.cn}

\author[2]{\fnm{Dusit} \sur{Niyato}}
\email{dniyato@ntu.edu.sg}

\author[3]{\fnm{Wei} \sur{Ni}}
\email{wei.ni@ieee.org}

\affil*[1]{%
  \orgdiv{School of Information and Control Engineering},
  \orgname{China University of Mining and Technology},
  \orgaddress{
    \city{Xuzhou},
    \postcode{221116},
    \state{Jiangsu},
    \country{China}%
  }
}

\affil[2]{%
  \orgdiv{College of Computing and Data Science},
  \orgname{Nanyang Technological University},
  \orgaddress{
    \city{Singapore},
    \postcode{639798},
    \country{Singapore}%
  }
}

\affil[3]{%
  \orgdiv{School of Engineering},
  \orgname{Edith Cowan University},
  \orgaddress{
    \city{Perth},
    \state{Western Australia},
    \postcode{6027},
    \country{Australia}%
  }
}


\abstract{Diffusion language models (DLMs) offer a non-autoregressive alternative for mobile edge agentic artificial intelligence (AI) by refining tokens through iterative denoising rather than left-to-right decoding. Compared with autoregressive Transformer-based large language models (LLMs), DLMs can update multiple uncertain tokens in parallel and exploit bidirectional context throughout the generation process, enabling more flexible quality-latency trade-offs beyond fixed sequential decoding. These properties are particularly attractive for edge agents, where partial refinement, early exit, and constraint-guided correction can reduce response delay and communication overhead while improving robustness under noisy, incomplete, or dynamic contexts. This survey reviews DLM foundations and analyzes their suitability for edge settings under latency, memory, energy, bandwidth, privacy, and reliability constraints. We cover resource-efficient architectures, training and inference acceleration, compression, edge/cloud deployment, communication-aware serving, Internet of Things (IoT)/wireless applications, and evaluation of DLM-based agents. We further discuss open issues in long-context state management, split inference, trustworthy execution, multimodal grounding, and reproducible benchmarking. The goal is to connect DLM modeling properties, including bidirectionality, parallel refinement, controllability, and quality-latency elasticity, with system-level requirements of future mobile edge intelligence.}




\maketitle

\section{Introduction}\label{sec1}
%\subsection{Motivation and Scope}

In recent years, artificial intelligence (AI) has been increasingly incorporated into the Internet of Things (IoT). The proliferation of smart devices, edge sensors, and cyber--physical infrastructures has generated large volumes of heterogeneous data that require intelligent processing and decision-making capabilities. At the same time, large language models (LLMs) have shown strong capabilities in natural language understanding (NLU), reasoning, and multimodal interaction~\cite{xiao2024efficient,saheed2025autonomous}. Integrating such capabilities into IoT environments may help future IoT systems become more autonomous, adaptive, and context-aware~\cite{karunanayake2025toward}. However, the use of LLMs in real-time IoT applications remains constrained by resource limitations~\cite{xiao2024efficient,saheed2025autonomous}. In addition, privacy and security requirements may prevent institutions from accessing commercial state-of-the-art LLM services, requiring local deployment within private networks~\cite{xiao2024efficient}. Thus, deploying LLMs directly on resource-constrained mobile edge IoT devices remains challenging, making LLM efficiency optimization a core research topic for mobile edge deployment~\cite{xiao2024efficient,saheed2025autonomous}. This highlights the growing need for adaptive and resilient systems that can operate efficiently at the mobile edge~\cite{karunanayake2025toward}.

%\subsection{Limitations of Autoregressive LLMs in Mobile and IoT Systems}

However, realizing this vision at the mobile edge remains nontrivial because modern generative models impose substantial computational, memory, and communication requirements. For example, GPT-4 has been estimated to contain approximately 1.76 trillion parameters~\cite{wang2024parameter}, a scale that far exceeds the hardware capacities of typical edge nodes~\cite{zheng2025review}. Most current LLMs rely on autoregressive architectures that generate tokens sequentially from left to right. Although this paradigm excels in natural language processing (NLP), it introduces inherent inefficiencies that limit its applicability in resource-constrained IoT environments. The sequential nature of autoregressive decoding leads to high inference latency and poor parallelization efficiency, presenting a major bottleneck for latency-sensitive edge applications~\cite{shi2026simpletool}. Furthermore, the intensive memory access patterns of these models place significant pressure on both on-device resources and wireless communication infrastructures. Specifically, processing long inputs, such as the 128,000-token context window supported by advanced GPT models, requires the maintenance of large Key-Value (KV) caches~\cite{sun2026hillinfer}. This operation can consume gigabytes (GB) of random-access memory (RAM) during inference, making deployment on memory-limited IoT devices difficult~\cite{zheng2025review,sun2026hillinfer}. As IoT systems typically operate under stringent constraints regarding computational capability, energy availability, and network bandwidth, exploring alternative generative paradigms that align with edge computing architectures is important~\cite{zheng2025review}.

Table~\ref{tab:llm_edge_comparison} provides a qualitative comparison of autoregressive LLMs, diffusion models (DMs), and diffusion language models (DLMs) from the perspective of modeling mechanisms. To complement this paradigm-level discussion, Table~\ref{tab:reported_dlm_ar_quantitative} summarizes representative quantitative evidence reported in recent technical papers. These results suggest that diffusion-based LLMs have begun to approach autoregressive LLMs on general reasoning and code benchmarks, while showing particular advantages in planning-oriented tasks and emerging faster-than-autoregressive inference settings. 

%%%%%%%%%%%%%%%%%%%%%%%%%%%%table1
% Required packages:
% \usepackage{booktabs}
% \usepackage{tabularx}
% \usepackage{array}
% \usepackage{ragged2e}
% Required packages in the preamble:
% \usepackage{booktabs}
% \usepackage{tabularx}
% \usepackage{array}
% \usepackage{ragged2e}

\begin{table}[!t]
\centering
\footnotesize
\caption{Representative LLM paradigms.}
\label{tab:llm_edge_comparison}
\renewcommand{\arraystretch}{1.08}
\setlength{\tabcolsep}{3pt}
\begin{tabularx}{\textwidth}{
@{}
L{0.14\textwidth}
L{0.16\textwidth}
L{0.10\textwidth}
L{0.25\textwidth}
Y
@{}
}
\toprule
\textbf{Model} &
\textbf{Paradigm} &
\textbf{Scale} &
\textbf{Backbone} &
\textbf{Inference Pattern / Edge Bottleneck} \\
\midrule

\tablegroupgray
\multicolumn{5}{@{}l}{\textit{Autoregressive Transformer-based LLMs}} \\
\midrule

\textbf{GPT-3}\par\cite{brown2020language} &
Autoregressive LM &
175B &
Transformer decoder with sparse attention &
Sequential next-token generation; KV cache grows with context. \\

\textbf{GPT-4}\par\cite{achiam2023gpt} &
Autoregressive multimodal LM &
1.76T &
Transformer-style model with closed technical details &
Frontier-scale capability; unsuitable for direct edge deployment. \\

\textbf{Llama~3.1}\par\cite{meta2024llama31} &
Autoregressive LM &
8B / 70B / 405B &
Optimized Transformer with GQA and 128K context &
Open-weight autoregressive baseline; long-context inference incurs substantial KV-cache pressure. \\

\textbf{Mistral~7B}\par\cite{jiang2023mistral7b} &
Efficient autoregressive LM &
7B &
Transformer with GQA, SWA, and rolling KV cache &
More deployment-friendly than larger autoregressive models, but generation remains token-by-token. \\

\textbf{Mixtral~8$\times$7B}\par\cite{jiang2024mixtralexperts} &
Sparse MoE autoregressive LM &
47B total / 13B active &
Transformer MoE with top-2 expert routing &
Reduces active compute per token, but requires large weight storage and routing overhead. \\

\midrule
\tablegroupgray
\multicolumn{5}{@{}l}{\textit{Diffusion-based Language Models}} \\
\midrule

\textbf{Diffusion-LM}\par\cite{li2022diffusion} &
Continuous diffusion LM &
Task-scale &
Transformer-based denoiser over continuous word vectors &
Non-left-to-right refinement; useful for control and infilling, but not yet a general large-scale LLM. \\

\textbf{LLaDA~8B}\par\cite{nie2025large} &
Masked diffusion LM &
8B &
Transformer mask predictor with full attention &
Predicts multiple masked tokens in parallel; avoids causal KV decoding but requires iterative denoising. \\

\textbf{Diffusion\hspace{0pt}Gemma 26B-A4B}\par\cite{google2026diffusiongemma_model,google2026diffusiongemma_blog} &
Masked discrete diffusion MoE LM &
26B total / about 4B active &
Gemma-family MoE backbone with block-diffusion decoding &
Open-weight DLM checkpoint; parallel refinement can improve local GPU throughput, but iterative denoising and model size remain edge bottlenecks. \\

\textbf{Dream~7B}\par\cite{ye2025dream7bdiffusionlarge} &
Discrete diffusion LLM &
7B &
Transformer-based DLM with autoregressive initialization &
Parallel iterative denoising with arbitrary-order generation and tunable quality--speed trade-offs. \\

\bottomrule
\end{tabularx}

\vspace{3pt}
\begin{minipage}{\textwidth}
%\scriptsize
\footnotesize
\RaggedRight
\textit{Note:} \textbf{B}: Billion (parameters); \textbf{GQA}: Grouped-Query Attention; \textbf{GPU}: Graphics Processing Unit; \textbf{SWA}: Sliding Window Attention; \textbf{MoE}: Mixture-of-Experts.
\end{minipage}
\end{table}
%%%%%%%%%%%%%%%%%%%%%%%%%%%%table2
\begin{table}[t]
\caption{Reported quantitative evidence comparing autoregressive LLMs and diffusion-based LLMs.}
\label{tab:reported_dlm_ar_quantitative}
\centering
\scriptsize
\renewcommand{\arraystretch}{1.18}
\setlength{\tabcolsep}{3pt}
\begin{tabularx}{\textwidth}{@{}
>{\RaggedRight\arraybackslash}p{0.1\textwidth}
>{\RaggedRight\arraybackslash}p{0.20\textwidth}
>{\RaggedRight\arraybackslash}p{0.14\textwidth}
>{\RaggedRight\arraybackslash}X
>{\RaggedRight\arraybackslash}p{0.24\textwidth}
@{}}
\toprule
\textbf{Study} &
\textbf{Compared Models} &
\textbf{Evaluation Focus} &
\textbf{Reported Results} &
\textbf{Implication} \\
\midrule

\cite{nie2025large} &
\textbf{LLaDA 8B Base} (DLM) vs. \textbf{Llama3 8B Base} (autoregressive) &
General, math, and code benchmarks &
\begin{minipage}[t]{\linewidth}
\RaggedRight
\textbf{MMLU}: 65.9 vs. 65.4\par
\textbf{GSM8K}: 70.3 vs. 48.7\par
\textbf{MATH}: 31.4 vs. 16.0\par
\textbf{HumanEval}: 35.4 vs. 34.8\par
\textbf{HumanEval-FIM}: 73.8 vs. 73.3
\end{minipage} &
DLMs can reach performance comparable to similarly sized autoregressive LLMs on several standard benchmarks. \\
\addlinespace[2pt]
\midrule
\addlinespace[2pt]

\cite{ye2025dream7bdiffusionlarge} &
\textbf{Dream 7B} (DLM) vs. \textbf{Qwen2.5 7B} / \textbf{Llama3 8B} (autoregressive) &
Math reasoning and code generation &
\begin{minipage}[t]{\linewidth}
\RaggedRight
\textbf{GSM8K}: 77.2 vs. 78.9 / 55.3\par
\textbf{MATH}: 39.6 vs. 41.1 / 18.0\par
\textbf{HumanEval}: 57.9 vs. 56.7 / 35.4
\end{minipage} &
DLMs can remain competitive with strong autoregressive baselines on reasoning and code tasks. \\
\addlinespace[2pt]
\midrule
\addlinespace[2pt]

\cite{ye2025dream7bdiffusionlarge} &
\textbf{Dream 7B} (DLM) vs. \textbf{Qwen2.5 7B} / \textbf{Llama3 8B} (autoregressive) &
Planning-oriented tasks &
\begin{minipage}[t]{\linewidth}
\RaggedRight
\textbf{Countdown}: 16.0 vs. 6.2 / 3.7\par
\textbf{Sudoku}: 81.0 vs. 21.0 / 0.0\par
\textbf{Trip planning}: 17.8 vs. 3.6 / 8.7
\end{minipage} &
Bidirectional refinement may benefit constraint satisfaction and long-horizon planning. \\
\addlinespace[2pt]
\midrule
\addlinespace[2pt]

\cite{wang2025diffusion} &
\textbf{D2F-Dream-Base-7B} / \textbf{D2F-LLaDA-Instruct-8B} vs. \textbf{autoregressive baselines} &
Inference throughput &
\begin{minipage}[t]{\linewidth}
\RaggedRight
\textbf{GSM8K}: 119.9 vs. 48.0 / 52.7 tokens/s\par
\textbf{HumanEval}: 81.6 vs. 50.9 / 55.4 tokens/s
\end{minipage} &
Recent DLM acceleration methods can enter a faster-than-autoregressive inference regime. \\

\bottomrule
\multicolumn{5}{@{}p{0.96\textwidth}@{}}{\textit{Note}: The numbers are reported from the original papers under their respective evaluation settings. Therefore, the table is intended as a compact literature-level summary rather than a controlled unified benchmark. \textbf{MMLU}: Massive Multitask Language Understanding; \textbf{GSM8K}: Grade School Math 8K; \textbf{FIM}: fill-in-the-middle.} \\
\end{tabularx}
\end{table}
%%%%%%%%%%%%%%%%%%%%%%%%%%%%

%\subsection{Potential of Diffusion Language Models}

\begin{figure}[!t]
    \centering
    \includegraphics[width=1\linewidth]{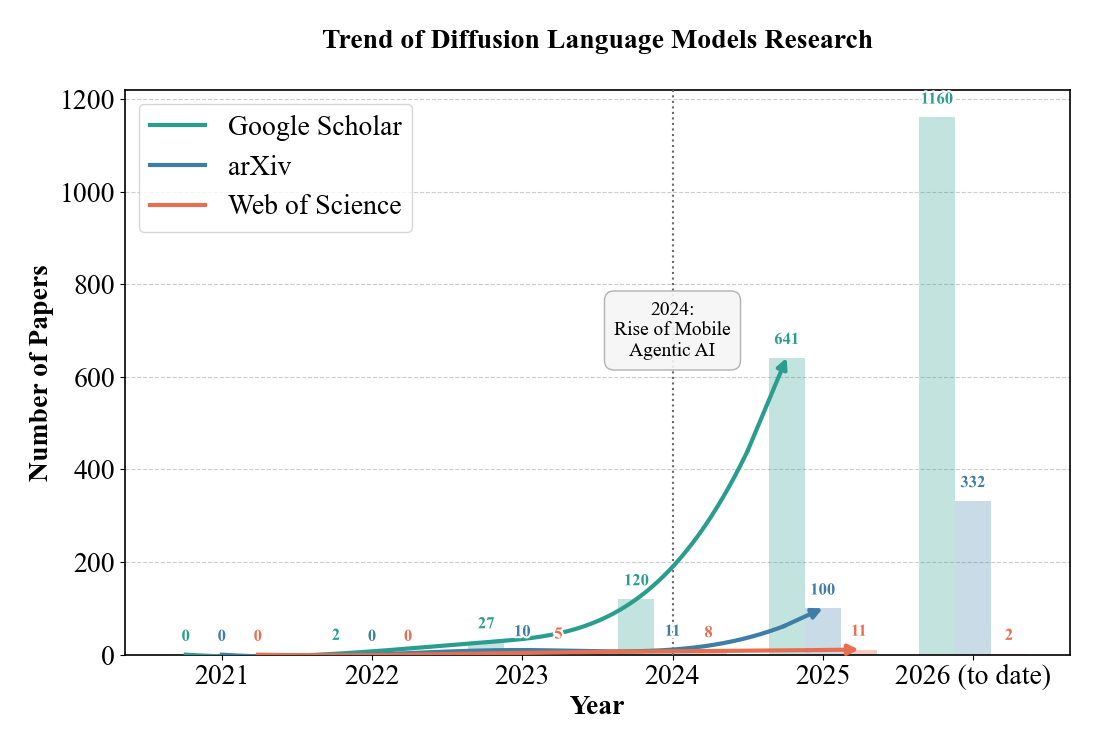}
    \caption{Research trends of DLMs across Google Scholar, arXiv, and Web of Science}
    \label{Fig.1.Trend of DLMs Research}
\end{figure}

DLMs have emerged as a non-autoregressive alternative to conventional autoregressive frameworks for edge intelligence~\cite{li2022diffusion,sahoo2024simple}. As shown in Fig.~\ref{Fig.1.Trend of DLMs Research}, research on DLMs has grown rapidly in recent years, especially since 2024, driven by mobile agentic AI and the need to migrate generative capabilities to resource-constrained edge environments. The trend serves as an indicative literature-level signal since the three sources differ in indexing scope and update policy.

Unlike token-by-token autoregressive generation, DLMs synthesize text through iterative denoising from a known prior, such as Gaussian noise in embedding space or an absorbing [MASK] state. This process updates multiple tokens in parallel and incorporates bidirectional context during generation. For edge intelligence, these properties enable better hardware utilization, flexible compute-quality trade-offs, inference-time controllability, and distributed inference across cloud-edge-device hierarchies~\cite{yang2025efficient}. They also align with decentralized IoT systems, where diffusion-based solution generators can support complex optimization in Multi-access Edge Computing (MEC) networks~\cite{liang2025gdsg}.

%\subsection{Motivation for Distributed DLMs in Mobile Edge Computing}

Distributed DLMs are suitable for mobile edge computing because their iterative denoising trajectory can be partitioned across device-edge-cloud tiers, reducing the local memory and computation burden of standalone mobile or IoT devices~\cite{sung2025memory}. Their fixed-length latent states are also more predictable than token-by-token autoregressive synchronization with expanding KV caches, which can reduce bandwidth overhead and transmission latency in device-edge collaboration~\cite{du2023exploring}. Moreover, DLMs support anytime decoding, allowing systems to adjust denoising depth and offloading ratios according to network conditions, latency targets, and battery status~\cite{miles2026test}. By keeping initial noising and final decoding on devices while offloading only obfuscated intermediate states, distributed DLMs may further support privacy-preserving collaborative generation~\cite{allmendinger2024collafuse}.

%\subsection{Comparison with Other Similar Surveys}

Existing diffusion-related surveys can be broadly grouped into three categories, as summarized in Table~\ref{tab:survey_comparison}. A first line of work reviews general DM foundations, model variants, and broad applications. For instance, ~\cite{croitoru2023diffusion} focus on denoising DMs in computer vision, covering denoising diffusion probabilistic models (DDPMs), noise conditional score networks (NCSNs), stochastic differential equations (SDEs), and vision tasks such as image generation, super-resolution, inpainting, and editing. ~\cite{1yang2023diffusion} provide a broader methodological survey of DMs, emphasizing efficient sampling, likelihood estimation, structured data, and applications across vision, NLP, temporal modeling, and scientific domains. ~\cite{ahsan2025comprehensive} further summarize DM foundations and cross-domain applications, including image synthesis, audio generation, healthcare, time-series forecasting, anomaly detection, and molecular dynamics. These surveys provide foundations for general diffusion modeling. Yet, they do not study DLMs as edge-deployable backbones for mobile agents.

A more recent line of work specifically investigates DLMs and discrete diffusion large models. In particular, ~\cite{li2025survey} review continuous, discrete, and hybrid DLMs, together with training, inference optimization, caching, multimodal extensions, and applications. ~\cite{yu2025discrete} focus on discrete diffusion large language and multimodal models, covering mathematical formulations, representative diffusion large language models (dLLMs) and diffusion multimodal large language models (dMLLMs), training, inference, quantization, privacy, safety, and applications. Although these works establish important DLM taxonomies, they do not systematically connect DLM structures with mobile edge deployment, communication-computation cooperation, IoT/wireless applications, or agent-oriented evaluation.

Another related line of work studies DMs in networks, wireless systems, and semantic communications. For example, ~\cite{luong2025diffusion} review applications of DMs in future networks, including channel modeling, signal reconstruction, integrated sensing and communication (ISAC), edge resource management, semantic communications, and security. ~\cite{qin2025generative} focus on DMs for generative semantic communications, emphasizing conditional diffusion, efficient diffusion, generalized diffusion, and semantic decoding as an inverse problem. ~\cite{fan2026generative} survey generative diffusion models (GDMs) for wireless networks from sensing, transmission, application, and security perspectives. 

Distinct from these works, this current survey centers on DLMs rather than continuous DMs/GDMs, and investigates how DLM properties---bidirectional context, parallel refinement, controllability, and quality-latency elasticity---can support mobile edge agentic AI under latency, memory, energy, bandwidth, privacy, and reliability constraints. Although DLMs have attracted growing attention, most studies still focus on model design, generation quality, and inference acceleration, while their role in mobile edge agentic AI remains underexplored. Mobile edge and IoT systems require low latency, small memory and computing footprint, energy efficiency, bandwidth awareness, privacy, reliability, and closed-loop interaction, calling for a system-level view beyond model-centric progress. The main contributions of this survey are summarized as follows:

\begin{table}[t]
\centering
\caption{Comparison between this survey and representative diffusion-related surveys.}
\label{tab:survey_comparison}
\footnotesize
%\scriptsize
\renewcommand{\arraystretch}{1.1}
\setlength{\tabcolsep}{4pt}
\begin{tabularx}{\textwidth}{
>{\raggedright\arraybackslash}p{0.12\textwidth}
>{\raggedright\arraybackslash}p{0.50\textwidth}
>{\raggedright\arraybackslash}X
}
\toprule
\textbf{Survey} &
\textbf{Main Focus} &
\textbf{Uncovered Aspects} \\
\midrule

\cite{croitoru2023diffusion} &
Denoising DMs in computer vision, including DDPMs, NCSNs, SDEs, relations to other generative models, and vision tasks such as generation, super-resolution, inpainting, editing, and translation. &
Lacks DLM-specific analysis and does not consider edge deployment, IoT/wireless constraints, or agentic execution. \\

\midrule

\cite{1yang2023diffusion} &
General DM methods and applications, including efficient sampling, likelihood estimation, structured data, connections to other generative models, and applications in vision, NLP, temporal data, multimodal learning, and science. &
Does not map diffusion modeling properties to mobile edge systems, resource-constrained deployment, or agent-oriented intelligence. \\

\midrule

\cite{ahsan2025comprehensive} &
Cross-domain DM foundations and applications, including image synthesis, image transformation, text-to-image generation, audio synthesis, healthcare, time-series forecasting, anomaly detection, and molecular dynamics. &
Provides broad application coverage but lacks DLM-centered inference, compression, edge serving, and agent evaluation perspectives. \\

\midrule

\cite{li2025survey} &
DLM ecosystem, including continuous and discrete DLMs, hybrid autoregressive-DMs, pre-training, post-training, inference optimization, caching, multimodal DLMs, and applications. &
Does not address edge/cloud deployment, wireless bandwidth limits, IoT scenarios, or structure-aware evaluation for DLM-based agents. \\

\midrule

\cite{yu2025discrete} &
Discrete diffusion in large language and multimodal models, including mathematical formulations, representative dLLMs/dMLLMs, training, inference, quantization, privacy, safety, and applications. &
Lacks a system-level treatment of mobile edge deployment, communication-computation co-design, and closed-loop agent execution. \\

\midrule

\cite{luong2025diffusion} &
DMs for future networks and communications, including channel modeling, signal reconstruction, ISAC, edge resource management, semantic communications, security, and other wireless applications. &
Focuses on DMs as wireless generation, reconstruction, and optimization tools, without studying token-level DLMs for edge-native agents. \\

\midrule

\cite{qin2025generative} &
DMs for generative semantic communications, including score-based foundations, conditional diffusion, efficient diffusion, generalized diffusion, semantic decoding as inverse problems, and human-/machine-/agent-centric scenarios. &
Centers on semantic reconstruction and transmission, leaving DLM-based agent architecture, split inference, and agent evaluation underexplored. \\

\midrule

\cite{fan2026generative} &
GDMs for wireless networks, organized around sensing, transmission, application, and security planes, with emphasis on channel modeling, semantic communications, network optimization, vertical applications, and security. &
Emphasizes continuous GDMs for wireless networks rather than DLM-native modeling, edge deployment, and agentic decision loops. \\

\midrule

\textbf{This survey} &
\textbf{DLMs for mobile edge agentic AI, covering DLMs foundations, resource-efficient DLMs, edge/cloud deployment, communication-efficient diffusion intelligence, IoT and wireless applications, and evaluation frameworks.} &
\textbf{Bridges the above gaps by connecting DLM modeling properties with latency, memory, energy, bandwidth, privacy, reliability, and agent-oriented system requirements.} \\

\bottomrule
\end{tabularx}
\end{table}
%%%%%%%%
%\subsection{Contributions and Organization of the Survey}

\begin{itemize}
    \item \textbf{Edge-oriented view of DLMs.}
    We frame DLMs as an edge-oriented generative paradigm for mobile agents by connecting their iterative denoising, bidirectional context modeling, parallel token refinement, controllable generation, and quality-latency elasticity with the requirements of mobile edge intelligence.

    \item \textbf{Unified review of foundations and efficiency techniques.}
    We review the modeling foundations of DLMs and summarize resource-efficient techniques for lightweight architectures, efficient training, accelerated inference, and model compression.

    \item \textbf{System-level deployment and communication perspective.}
    We organize edge DLM deployment from device-only execution to edge-assisted, cloud-edge collaborative, distributed, and adaptive serving paradigms, and discuss communication-computation co-optimization under dynamic wireless conditions.

    \item \textbf{Applications, evaluation, and open challenges.}
    We summarize the potential of DLMs in IoT and wireless systems, review structure-aware evaluation across capability, interaction, deployment, safety, and reproducibility, and identify open challenges toward DLM-native mobile edge intelligence.
\end{itemize}

\begin{figure}[htbp]
    \centering
    \includegraphics[width=1\linewidth]{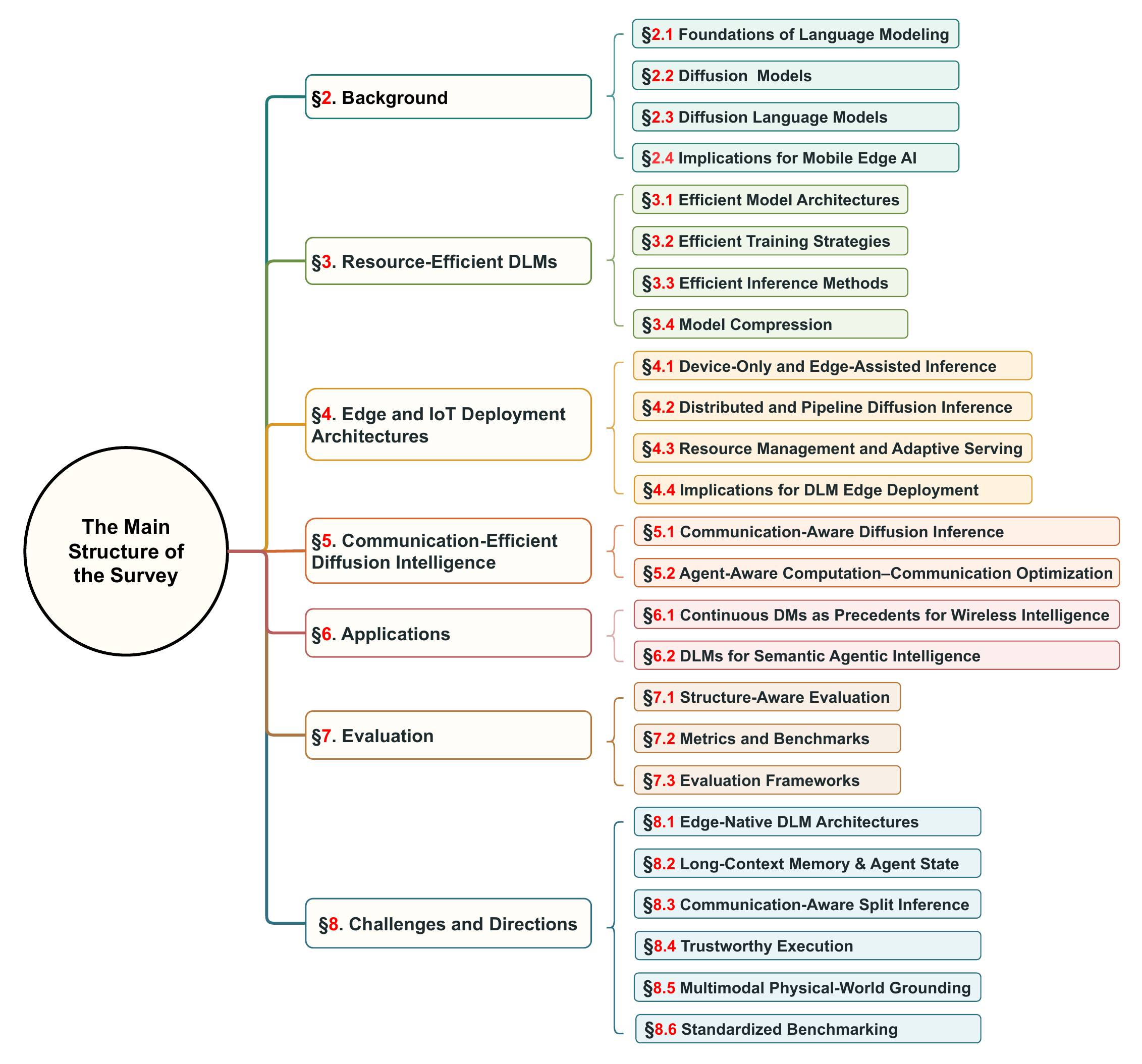}
    \caption{Survey framework for DLMs for mobile edge agentic AI, organized around foundations, edge-oriented applications, evaluation, and research challenges}
    \label{figure_survey_structure}
\end{figure}

The overall structure of this survey is illustrated in Fig.~\ref{figure_survey_structure}. The figure provides a roadmap from DLM foundations and resource-efficient techniques to edge deployment, communication-efficient diffusion intelligence, IoT and wireless applications, evaluation frameworks, and open challenges. For ease of reference, Table~\ref{tab:abbreviations} summarizes the
abbreviations and acronyms frequently used throughout this survey.

\FloatBarrier

%%%%%%%%%%%%%%%%%%%%%%%%%
\begin{table}[t]
\caption{Frequently used abbreviations.}
\label{tab:abbreviations}
\centering
\footnotesize
%\scriptsize
\renewcommand{\arraystretch}{1.3}
\setlength{\tabcolsep}{6pt}

\begin{tabularx}{0.8\textwidth}
{@{}>{\centering\arraybackslash}p{0.20\textwidth}B@{}}
\toprule
\textbf{Abbreviation} & \textbf{Full Term} \\
\midrule

AI    & Artificial Intelligence \\
DM    & Diffusion Model \\
DLM   & Diffusion Language Model \\
LLM   & Large Language Model \\
IoT   & Internet of Things \\
MEC   & Multi-access Edge Computing \\
NLP   & Natural Language Processing \\
MLM   & Masked Language Model \\
KV    & Key-Value \\
MoE   & Mixture-of-Experts \\
SDE   & Stochastic Differential Equation \\
D3PM  & Discrete Denoising Diffusion Probabilistic Model \\
CFG   & Classifier-Free Guidance \\
MLLM  & Multimodal Large Language Model \\
VLA   & Vision-Language-Action \\
UI    & User Interface \\
API   & Application Programming Interface \\
PEFT  & Parameter-Efficient Fine-Tuning \\
LoRA  & Low-Rank Adaptation \\
DP    & Differential Privacy \\
NFE   & Number of Function Evaluations \\
DRL   & Deep Reinforcement Learning \\

\bottomrule
\end{tabularx}
\end{table}
%%%%%%%%%%%%%%%%%%%%%%%%%

%
%
%
%%%%%%%%%%%%%%%%%%%%%%%%%%%%%%%%%%%%%%%%%%%%%%
\section{Background: Diffusion Language Models}\label{sec2}
\label{sec:background}

\subsection{Language Modeling Paradigms}

\subsubsection{Autoregressive Models}

Autoregressive models form the dominant paradigm of modern LLMs. They generate sequences through next-token prediction, where each token depends only on previous tokens~\cite{autoregressive_models2}. This factorizes the joint sequence probability as:

\begin{equation}
p(x_1, x_2, \ldots, x_T) = p(x_1) \prod_{t=2}^{T} p(x_t \mid x_1, x_2, \ldots, x_{t-1}),
\end{equation}
where \(x_t\) denotes the token at position \(t\), \(T\) denotes the sequence length, \(p(x_1)\) denotes the marginal probability of the first token, and \(p(x_t \mid x_1, x_2, \ldots, x_{t-1})\) denotes the conditional probability of \(x_t\) given all preceding tokens. Accordingly, this left-to-right factorization supports data compression, pattern extrapolation, and in-context learning~\cite{autoregressive_models2}, and has shaped several representative model families relevant to mobile edge agents.

Closed-source autoregressive Transformers, represented by GPT-3 and GPT-4, established the scaling paradigm by showing that larger parameter capacity and training data can enable emergent capabilities~\cite{autoregressive_models1}. Their strong zero-shot and few-shot generalization supports complex cognitive tasks without downstream fine-tuning~\cite{autoregressive_models2}, making them cloud-based upper-bound references for mobile edge agent intelligence. Open-source models, represented by Llama 1 and Llama 2, use optimized autoregressive Transformer designs with mechanisms such as grouped-query attention (GQA)~\cite{autoregressive_models1}. Llama 2 further combines large-scale self-supervised pre-training with Reinforcement Learning from Human Feedback (RLHF) to align with human preferences~\cite{autoregressive_models1}. Their openness and efficiency make them key foundations for on-device and lightweight edge research.

Beyond natural language, autoregressive modeling has been extended to time-series and visual generation. Time-LLM maps time-series data into textual prototypes through patch reprogramming, enabling frozen LLaMA or GPT-2 models to perform forecasting and spatiotemporal reasoning~\cite{autoregressive_models4}. This improves the ability of edge agents to process physical sensor data. In visual generation, traditional Vector-Quantized Generative Adversarial Network (VQGAN)-style raster-scan autoregressive generation is inefficient~\cite{autoregressive_models3}. Visual autoregressive modeling addresses this limitation by reformulating generation as next-scale prediction, using coarse-to-fine multi-scale parallel autoregressive modeling to surpass Diffusion Transformers (DiTs) in visual fidelity and scalability~\cite{autoregressive_models3}.

\subsubsection{Masked Language Models for Edge Intelligent Perception}

Masked Language Models (MLMs) have introduced bidirectional representation learning by masking part of an input sequence and training the model to reconstruct the missing tokens from surrounding contexts~\cite{devlin2019bert}. Unlike autoregressive models, which impose unidirectional causality, MLMs jointly exploit left and right contexts to learn deep semantic representations~\cite{devlin2019bert}. For mobile edge agents, this contextual reconstruction ability provides compact semantic perception for robust decision-making under limited computation, memory, and bandwidth~\cite{9052677}.

BERT has established the canonical MLM architecture through cloze-style masked prediction, showing that large-scale masked pre-training on unlabeled corpora yields transferable semantic embeddings for many downstream tasks~\cite{devlin2019bert}. Its variants, such as RoBERTa, verify the effectiveness of optimized masked pre-training for representation learning~\cite{devlin2019bert,liu2019roberta}. To address the low pre-training efficiency of standard MLMs, DeBERTaV3 introduces a sample-efficient Replaced Token Detection (RTD) objective and Gradient-Disentangled Embedding Sharing (GDES), which reduces conflicts between discriminative and generative objectives and improves NLU performance~\cite{masked_language_models2}.

MLM embeddings also support large-scale unstructured data analysis. BERTopic uses document embeddings from BERT or RoBERTa, combines density-based clustering with class-based Term Frequency-Inverse Document Frequency (c-TF-IDF), and enables coherent latent-topic discovery in large text corpora~\cite{masked_language_models1}. Beyond text, masked prediction has been extended to time-series analysis and visual generation. In mobile edge computing, this contextual completion mechanism can impute missing sensor readings and support environmental situational awareness, making MLMs an important precursor to DLMs that also rely on bidirectional reconstruction and iterative refinement.

\subsection{Diffusion Models}

DMs are generative latent variable models inspired by non-equilibrium thermodynamics. They gradually corrupt a complex data distribution into a simple prior through a parameterized Markov chain, and then synthesize data by learning the reverse denoising trajectory~\cite{Diffusion_Models1}. According to the topology of the data state space, DMs are commonly divided into continuous and discrete paradigms.

\subsubsection{Continuous Diffusion Models}
For continuous state spaces, such as images and audio, DMs define a forward process that incrementally injects Gaussian noise and a reverse process that reconstructs data through denoising. The forward corruption process can be formulated under the continuous-time framework of SDEs:
\begin{equation}
dx = f(x,t)dt + g(t)dw,
\end{equation}
where $f(\cdot, t)$ is the drift coefficient, $g(t)$ is the diffusion coefficient, and $w$ is the standard Wiener process~\cite{Diffusion_Models5}. The generative phase follows the reverse-time evolution of this SDE. By Anderson's theorem, the reverse SDE is given by
\begin{equation}
dx = \left[f(x,t) - g^2(t)\nabla_x \log p_t(x)\right]dt + g(t)d\bar{w}.
\end{equation}
Solving this process requires neural networks to estimate the score function $\nabla_x \log p_t(x)$ of the perturbed data distribution, commonly through objectives such as Denoising Score Matching~\cite{Diffusion_Models3}. To reduce the cost of score estimation in high-dimensional spaces, Latent Diffusion Models (LDMs) perform diffusion in the low-dimensional latent space of a pre-trained autoencoder, supporting lighter edge deployment~\cite{11rombach2022high}. Flow Matching (FM) further accelerates training and sampling by regressing the vector field of a fixed conditional probability path for Continuous Normalizing Flows (CNFs)~\cite{Diffusion_Models1}.

\subsubsection{Discrete Diffusion Models}

Because text is inherently discrete, continuous DMs are unsuitable for direct natural language modeling. Discrete Denoising Diffusion Probabilistic Models (D3PMs) extend diffusion to categorical spaces by replacing Gaussian noising with Markov transition matrices $Q_t$, which define probabilistic token-state transitions~\cite{Diffusion_Models2}. A major advantage of this paradigm is the ability to design structured transition matrices, including absorbing states, such as the [MASK] token. When the forward process iteratively masks text, the reverse process predicts masked tokens in a hierarchical and layer-wise manner. This creates a mathematical connection between DMs and conventional MLMs, forming the theoretical basis of modern DLMs~\cite{Diffusion_Models2}.

\subsection{Diffusion Language Models}

DLMs represent a non-autoregressive paradigm for language generation. Instead of generating tokens strictly from left to right, they synthesize text by iteratively reconstructing corrupted sequences. For mobile edge agents, this mechanism offers parallel token refinement, bidirectional context modeling, controllable generation, and flexible quality-latency adaptation under resource constraints.

\subsubsection{Generation Pipeline of Diffusion Language Models}

DLM architectures rely on the duality between forward noising and reverse generation. Due to the discrete nature of language, existing methods can commonly be grouped into three pipelines: Continuous Embedding Diffusion, Discrete State-Space Diffusion, and Masked Diffusion with Maximum Likelihood.

\noindent \textbf{Continuous Embedding Diffusion.}
Methods, such as Diffusion-LM, map discrete tokens into continuous embedding space, inject Gaussian noise in the forward process, denoise latent vectors through a learned reverse network, and finally project continuous variables back to tokens through rounding. This design enables fine-grained controllable generation with gradient-guided optimization, but may suffer from rounding errors, embedding sensitivity, and slow sampling~\cite{li2022diffusion}.

\noindent \textbf{Discrete State-Space Diffusion.}
D3PMs perform diffusion directly in discrete token space using structured Markov transition matrices. Tokens evolve according to categorical distributions, and absorbing states can stabilize selected token states. The reverse process recovers sequences through iterative categorical sampling. Unlike masked prediction, these absorbing placeholders are not designed as explicit masked-token targets. This pipeline avoids continuous-space quantization errors and provides a principled framework for discrete sequence modeling~\cite{Diffusion_Models2}.

% \paragraph{Masked Diffusion and Likelihood-Based Training}
\noindent \textbf{Masked Diffusion and Maximum Likelihood Estimation.}
Masked Diffusion Language Models (MDLMs) use a masking-based forward process, in which tokens of a clean sequence $x_0$ are corrupted according to a time-dependent masking probability. The reverse model predicts the original tokens at masked positions from the corrupted sequence $x_t$. Under the linear masking process adopted by LLaDA, where $t \sim \mathcal{U}(0,1]$ and each token is independently replaced with the \texttt{[MASK]} token with probability $t$, a principled masked-token training objective can be written as~\cite{nie2025large}:
\begin{equation}
\mathcal{L}_{\mathrm{mask}}(\theta)
=
-\mathbb{E}_{x_0,t,x_t}
\left[
\frac{1}{t}
\sum_{i \in \mathcal{M}_t}
\log p_\theta(x_0^i \mid x_t)
\right],
\label{eq:masked_diffusion_objective}
\end{equation}
where $x_0 \sim p_{\mathrm{data}}$ is a clean sequence sampled from the data distribution, $x_t \sim q_{t|0}(\cdot \mid x_0)$ is sampled from the forward masking distribution, and $\mathcal{M}_t$ denotes the set of masked positions at time step $t$. This objective computes the cross-entropy loss only over masked tokens and upper-bounds the negative log-likelihood of the induced generative model, thereby providing a structured approximate maximum-likelihood training framework~\cite{nie2025large}. From a complementary variational perspective, ~\cite{sahoo2024simple} derive a continuous-time Rao--Blackwellized negative Evidence Lower Bound (ELBO) under a substitution-based reverse-process parameterization, showing that masked diffusion training can be expressed as a schedule-weighted average of masked language modeling losses.

% \paragraph{Masked Diffusion and Maximum Likelihood Estimation}
% Masked Diffusion Language Models (MDLMs) use a masking-based forward process, where tokens are randomly or \modified{meticulously} replaced with [MASK] tokens. \modified{The reverse model predicts masked positions using a masked-token reconstruction objective, closely related to the Rao-Blackwellized likelihood formulations of masked diffusion models~\cite{sahoo2024simple,nie2025large}. A simplified form can be written as:}
% \begin{equation}
% \mathcal{L}_{\mathrm{MDLM}}
% =
% -\mathbb{E}_{t,x_t}
% \sum_{i\in M_t}
% \log p_\theta(x_0^i \mid x_t),
% \end{equation}
% % The reverse model predicts masked positions through a Rao-Blackwellized maximum likelihood estimation (MLE) objective, equivalent to maximizing the conditional probability of original tokens given the corrupted input:
% % \begin{equation}
% % \mathcal{L}_{MDLM} = - \mathbb{E}_{t, x_t} \sum_{i \in \mathcal{M}_t} \log p_\theta(x_0^i \mid x_t),(\textcolor{red}{add a reference})
% % \end{equation}
% \modified{where $M_t$ denotes the set of masked positions at time step $t$.} This formulation supports efficient semi-autoregressive generation of variable-length sequences. Large-scale models, such as LLaDA, adopt this mechanism for scratch pre-training with probabilistic inference optimized through the evidence lower bound (ELBO)~\cite{sahoo2024simple, nie2025large}.

Beyond these training formulations, DiffusionGemma further demonstrates that masked discrete diffusion is moving from research prototypes toward open-weight checkpoints in the Gemma family. The public DiffusionGemma-26B-A4B-it model combines masked diffusion decoding with a mixture-of-experts (MoE) backbone, providing a recent reference point for studying scalable DLM inference in practical deployment settings~\cite{google2026diffusiongemma_model,google2026diffusiongemma_blog}.

Table~\ref{tab:dlm_pipeline_summary} compares these three pipelines in terms of structure, forward and reverse processes, advantages, and limitations, providing a compact reference for later discussions on DLM design and deployment.
%%%%%%%%%%%%%%%%%%%%%%%%%%%%
% Place this command before the table, e.g., in the preamble or before \begin{table*}
% Required packages in the preamble:
% \usepackage{graphicx}
% \usepackage{array}
% \usepackage{ragged2e}
% Required packages in the preamble:
% \usepackage{graphicx}
% \usepackage{array}
% \usepackage{ragged2e}

\newcolumntype{P}[1]{>{\centering\arraybackslash}p{#1}}

\newcommand{\pipelinefig}[2][]{%
\makebox[\linewidth][c]{%
\includegraphics[
width=0.94\linewidth,
height=0.20\textheight,
keepaspectratio,
#1
]{#2}}%
}
\newcommand{\pipelinehead}[1]{\rule{0pt}{3ex}\textbf{#1}}

\begin{table}[!t]
%\scriptsize
\footnotesize
\centering
\caption{Comparison of generation pipelines in Diffusion Language Models.}
\label{tab:dlm_pipeline_summary}
\renewcommand{\arraystretch}{1.05}
\setlength{\tabcolsep}{3.5pt}

\begin{tabular}{
|P{0.305\textwidth}
|P{0.305\textwidth}
|P{0.305\textwidth}|
}
\hline
\pipelinehead{Continuous Embedding Diffusion} &
\pipelinehead{Discrete State-Space Diffusion} &
\pipelinehead{Masked Diffusion \& Maximum Likelihood} \\
\hline

\pipelinefig[trim=5 5 5 5,clip]{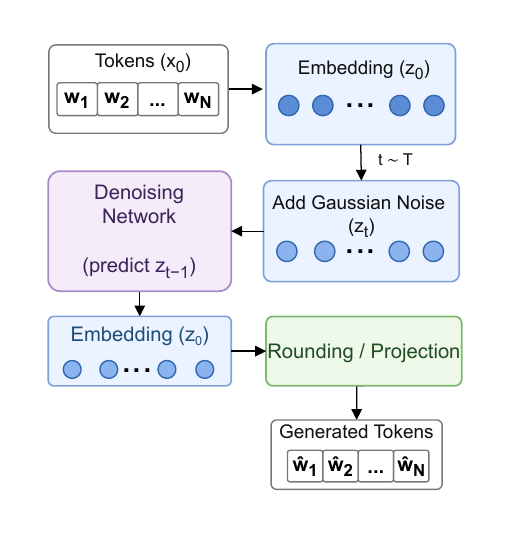} &
\pipelinefig[trim=5 5 5 5,clip]{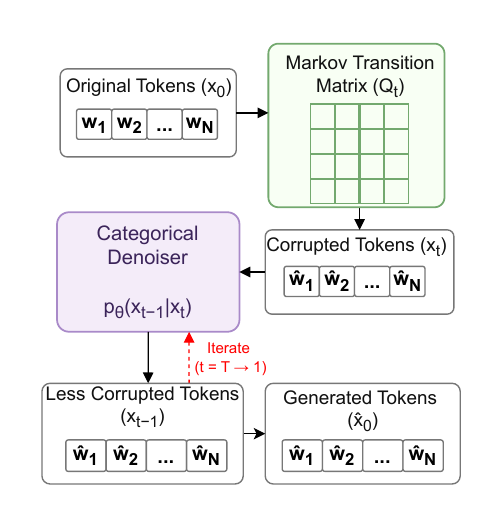} &
\pipelinefig[trim=5 5 5 5,clip]{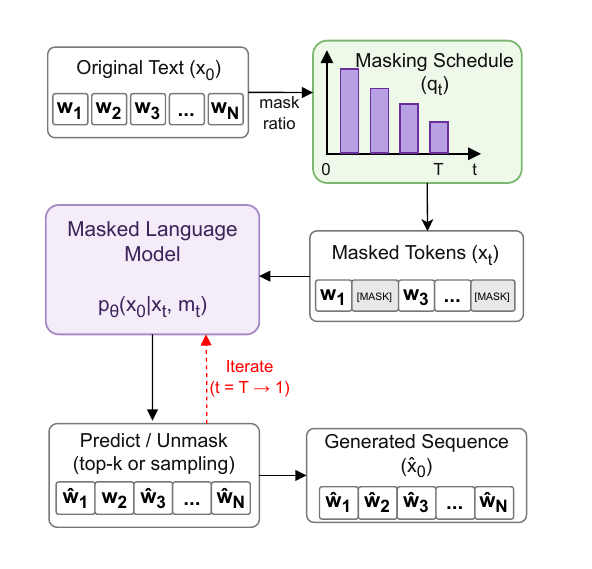} \\
\hline

\begin{minipage}[t]{0.96\linewidth}
\vspace{0pt}
\RaggedRight
\textbf{Structure:} Tokens $\rightarrow$ embeddings $\rightarrow$ Gaussian noising $\rightarrow$ denoising $\rightarrow$ rounding.

\medskip
\textbf{Description:} Maps tokens into continuous embeddings, performs Gaussian diffusion in latent space, and projects denoised embeddings back to tokens~\cite{li2022diffusion}.

\medskip
\textbf{Pros:} Gradient-based control; fine-grained guidance.

\medskip
\textbf{Cons:} Rounding errors; embedding sensitivity; slow sampling.
\end{minipage}
&
\begin{minipage}[t]{0.96\linewidth}
\vspace{0pt}
\RaggedRight
\textbf{Structure:} Tokens $\rightarrow$ Markov corruption $\rightarrow$ categorical denoising $\rightarrow$ output tokens.

\medskip
\textbf{Description:} Performs diffusion directly in token space using structured Markov transition matrices and reverse categorical sampling~\cite{Diffusion_Models2}.

\medskip
\textbf{Pros:} Avoids rounding; models discrete states directly.

\medskip
\textbf{Cons:} Transition design complexity; costly for large vocabularies.
\end{minipage}
&
\begin{minipage}[t]{0.96\linewidth}
\vspace{0pt}
\RaggedRight
\textbf{Structure:} Tokens $\rightarrow$ masking $\rightarrow$ masked-token prediction $\rightarrow$ iterative unmasking.

\medskip
\textbf{Description:} Uses a masking-based forward process and Rao-Blackwellized MLM-style MLE for semi-autoregressive generation~\cite{sahoo2024simple,nie2025large}.

\medskip
\textbf{Pros:} Efficient sampling; variable-length generation; scalable.

\medskip
\textbf{Cons:} Masking-policy dependence; multi-step refinement; costly pretraining.
\end{minipage}
\\
\hline

\end{tabular}
\end{table}

\subsubsection{Comparison with Autoregressive LLMs}

\begin{figure}[t]
    \centering
    \includegraphics[width=\textwidth]{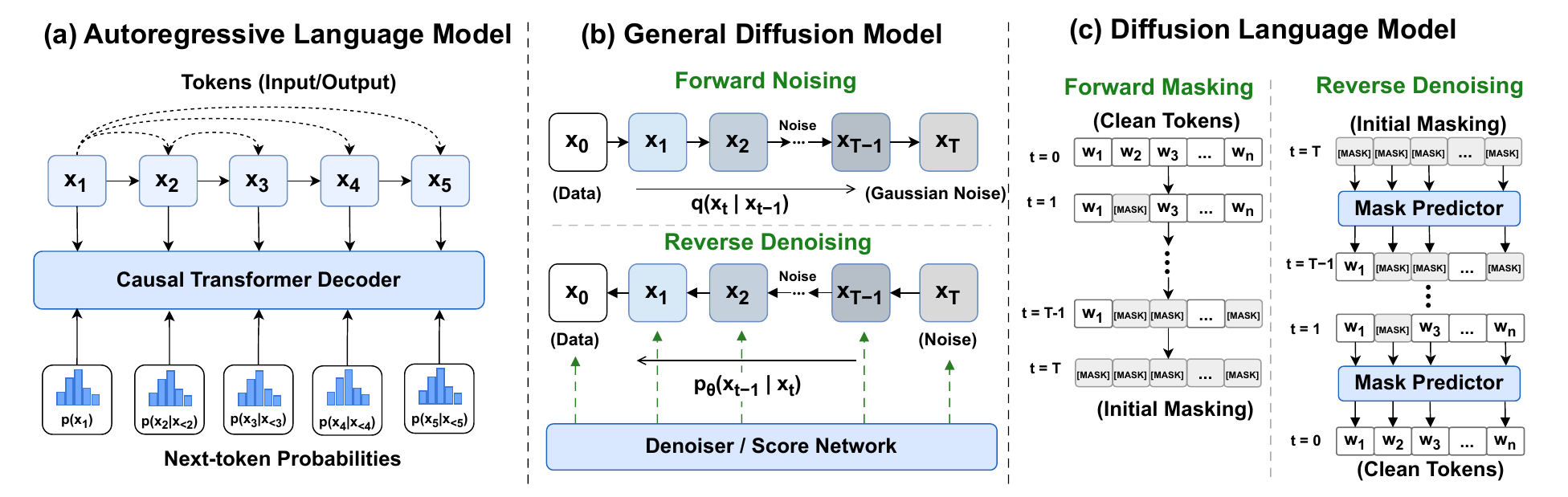}
   \caption{Generation mechanisms of autoregressive language models, general DMs, and DLMs}
    \label{fig:ar_dm_dlm_comparison}
\end{figure}

Although autoregressive LLMs remain dominant, DLMs address several structural limitations of left-to-right generation. To clarify this structural shift, Fig.~\ref{fig:ar_dm_dlm_comparison} contrasts autoregressive language modeling, general diffusion modeling, and diffusion language modeling from the perspective of their generation trajectories. Panel (a) represents left-to-right causal decoding, panel (b) shows continuous forward noising and reverse denoising, and panel (c) maps diffusion-style generation to token masking and iterative token denoising.

\noindent \textbf{Generation Direction and Contextual Modeling.}
Autoregressive models follow temporal causality and unidirectional decoding, which can cause error accumulation and limit global structural planning. In contrast, DLMs use bidirectional refinement. LLaDA, for example, improves global contextual reasoning and mitigates the Reversal Curse, namely the difficulty of inferring B is A after learning A is B~\cite{nie2025large}.

\noindent \textbf{Compute-Quality Trade-off.}
Autoregressive generation is constrained by parameter scale and fixed decoding trajectories. The Score Entropy Discrete Diffusion models (SEDD) shows that DMs can trade computation for quality by adjusting reverse sampling steps. At a comparable scale, it outperforms autoregressive counterparts, such as GPT-2, and maintains high-fidelity generation with far fewer network evaluations~\cite{Language_diffusion5, 15feng2025theoretical}.

\noindent \textbf{Controllability and Infilling.}
Because of unidirectionality, autoregressive models struggle with infilling and zero-shot constraint satisfaction. DLMs can incorporate conditioning guidance during inference. Diffusion-LM uses gradient-based guidance, while SEDD uses classifier-free guidance (CFG), enabling controllable generation without task-specific retraining~\cite{li2022diffusion, Language_diffusion5}.

\subsubsection{Multimodal Large Diffusion Language Models}

Recent studies have extended masked and discrete diffusion from text-only generation to multimodal understanding, reasoning, and generation. As summarized in recent surveys, a growing body of dMLLMs has explored diverse architectures for integrating visual and textual information within diffusion-based generation frameworks~\cite{li2025survey,yu2025discrete}. Representative examples include Dimple, which combines a vision encoder with a discrete DLM backbone and adopts an autoregressive-then-diffusion training strategy together with its Confident Decoding mechanism for parallel generation~\cite{yu2025dimplediscretediffusionmultimodal}; LaViDa, which equips DLMs with visual encoders and introduces complementary masking and Prefix-DLM caching for multimodal understanding~\cite{NEURIPS2025_975affbe}; and LLaDA-V, which extends a large language DM to multimodal understanding through visual instruction tuning and bidirectional attention within a purely diffusion-based multimodal large language model (MLLM) framework~\cite{You_2026_CVPR}. Beyond these understanding-oriented dMLLMs, MMaDA exemplifies a more unified multimodal diffusion direction by modeling text and image tokens within a shared discrete diffusion framework, enabling textual reasoning, multimodal understanding, and text-to-image generation under a unified denoising formulation~\cite{yang2025mmada}. 

Rather than exhaustively enumerating all multimodal DLMs, this survey highlights these representative models to illustrate major architectural directions relevant to mobile edge agents. Such developments suggest a flexible token-level interface for processing instructions, visual observations, user interface (UI) states, sensor streams, and action-related representations under tight latency, memory, and energy budgets. However, practical edge deployment still depends on efficient visual tokenization, denoising acceleration, cache optimization, model quantization, and adaptive device-edge offloading.

\subsection{Implications for Mobile Edge System Design}

Integrating DLMs into the cognitive architecture of mobile edge agents provides system-level implications along three complementary dimensions: compute--quality elasticity, memory-bandwidth efficiency, and deterministic and safe on-device execution.

\subsubsection{Elastic Compute Adaptation and Deployment}
Edge devices are characterized by substantial heterogeneity and temporal dynamics in computing resources (e.g., GPU/NPU availability) and power budgets. Exploiting the compute-quality trade-off validated by SEDD~\cite{Language_diffusion5}, edge agents can elastically adjust diffusion sampling steps through early stopping based on real-time energy thresholds. Under favorable resource conditions, the agent can execute multi-step denoising for complex reasoning; under battery constraints, it can truncate the sampling trajectory to deliver core semantics with ultra-low latency, offering flexibility not available in the same form in standard autoregressive models.

\subsubsection{Alleviating Memory Bandwidth Bottlenecks via Parallel Decoding}
Mobile edge devices (e.g., smartphones and IoT nodes) are generally bottlenecked by memory bandwidth rather than raw floating-point operations (FLOPs). The pervasive KV-caching and serial token generation of autoregressive models impose substantial memory access overhead. In contrast, DLMs (e.g., MDLM and LLaDA) can predict multiple tokens in parallel during the reverse diffusion phase~\cite{sahoo2024simple,nie2025large}. This parallelization improves the utilization of edge AI accelerators and reduces end-to-end perception-to-action latency for intelligent agents.

\subsubsection{Deterministic and Safe On-Device Execution}
Edge agents frequently translate cloud-issued directives into physical actions via local operating systems or actuators (e.g., smart home control and application programming interface (API) invocation). This requires reliable format control and strict safety constraints. By using gradient-based or gradient-free controllable generation in DLMs~\cite{li2022diffusion,Language_diffusion5}, generation can be constrained to follow predefined syntactic structures and security boundaries during action generation, thereby improving the robustness and safety of on-device AI systems.

The above discussion shows that DLMs are relevant to mobile edge agents, not only because they avoid strictly left-to-right decoding, but also because their denoising dynamics can be translated into concrete system-level functions. Table~\ref{tab:dlm_edge_agent_mapping} summarizes this mechanism-to-requirement mapping from edge-agent perspectives.

\begin{table}[t]
\caption{Mapping DLM properties to edge-native agent requirements.}
\label{tab:dlm_edge_agent_mapping}
\centering
\footnotesize
%\scriptsize
\renewcommand{\arraystretch}{1.18}
\setlength{\tabcolsep}{3pt}
\begin{tabular}{@{}N{0.22\textwidth}L{0.45\textwidth}L{0.29\textwidth}@{}}
\toprule
\textbf{DLM Property} &
\textbf{Edge-Agent Implication} &
\textbf{Deployment Concern} \\
\midrule

Bidirectional context &
Supports command repair, global planning, and context completion by using both left and right contexts~\cite{nie2025large}. &
Long-context memory and full-sequence refinement cost. \\
\addlinespace[2pt]

Parallel refinement &
Updates multiple uncertain tokens together for low-latency structured generation and action-token synthesis~\cite{wu2025fast}. &
Memory access, cache behavior, and synchronization overhead. \\
\addlinespace[2pt]

Adaptive denoising &
Adjusts refinement steps for early exit, anytime response, and quality-latency trade-offs~\cite{zhu2026esdllm}. &
Semantic drift or insufficient correction after truncation. \\
\addlinespace[2pt]

Structured controllability &
Guides generation toward valid formats, tool calls, API commands, and safe executable plans~\cite{li2022diffusion}. &
Need for constraint checking before final action execution. \\
\addlinespace[2pt]

Noising and split states &
Enables privacy-aware offloading by transmitting obfuscated intermediate states, instead of raw prompts~\cite{allmendinger2024collafuse}. &
Possible leakage from latent states, token distributions, or reasoning traces. \\
\addlinespace[2pt]

Quality-latency elasticity &
Adapts generation to battery level, wireless condition, task urgency, and interaction risk~\cite{peng2025efficient}. &
Requires joint reporting of latency, energy, memory, communication, and hardware settings. \\

\bottomrule
\end{tabular}
\end{table}
% section 2
\subsection{Lessons Learned}
Autoregressive LLMs remain preferable for open-ended and long-form generation, but their sequential decoding and KV-cache dependence create latency and memory bottlenecks at the edge~\cite{li2024survey,peng2025efficient}. DLMs better support infilling, structured generation, global planning, and constraint-aware action synthesis; yet, their practical gains depend on stable sampling, cache compatibility, and hardware-aware parallel decoding~\cite{nie2025large,wu2025fast}. Thus, edge systems should treat autoregressive models and DLMs as complementary backbones selected by task structure, latency budget, and ecosystem maturity.

\FloatBarrier

%
%
%
%
%%%%%%%%%%%%%%%%%%%%%%%%%%%%%%%%%%%%%%%%%%%%%%%%%%%%%%%%%%%%%%%%%%%%%%%%%%%%%%%%%%%%%%%%%%%%%%%%%%%%%%%%%%%%%%%%%%%%%%%%%%%%%%%%%%%%%
\section{Resource-Efficient Diffusion Language Models}\label{sec3}

\subsection{Efficient Model Architectures}

Deploying agentic AI in MEC is limited by edge devices' constrained computation, memory bandwidth, and power budgets. Traditional DMs incur high latency and redundancy because of iterative multi-step denoising. Recent work targets sub-second mobile inference through architectural redesign and lightweight optimization~\cite{zhao2024mobilediffusion, li2023snapfusion}. This section reviews efficient architectures in three directions: lightweight diffusion Transformers, block and hybrid autoregressive-DMs, and sparse MoE diffusion models.

\subsubsection{Lightweight Diffusion Transformers}
Conventional DMs rely on computation-heavy convolutional U-Nets. DiT, built entirely on attention, has become a simpler foundation for on-device diffusion. Token merging and pruning further reduce the attention memory wall, improving throughput on resource-limited hardware while preserving fidelity~\cite{bolya2023token}. Table~\ref{tab:efficient_architecture_comparison} summarizes representative efficient diffusion and DLM architectures in terms of model scale, application scenarios, and mobile/IoT-related deployment characteristics.

\begin{sidewaystable}[p]
\centering
\caption{Representative efficient diffusion and DLM architectures for edge deployment.}
\label{tab:efficient_architecture_comparison}
\scriptsize
\renewcommand{\arraystretch}{1.2} % 鎻愬崌琛岄珮锛岀敤鐣欑櫧浠ｆ浛妯嚎
\setlength{\tabcolsep}{4pt}       % 绋嶅井璋冩暣鍒楅棿璺?
\begin{tabularx}{\textheight}{
>{\raggedright\arraybackslash}p{0.14\textwidth}
>{\raggedright\arraybackslash}p{0.15\textwidth} % 鍑忓皬浜嗕唬琛ㄦā鍨嬪垪鐨勫搴?
>{\raggedright\arraybackslash}p{0.20\textwidth} % 澧炲姞浜嗘ā鍨嬪昂瀵稿垪鐨勫搴?
>{\raggedright\arraybackslash}X
>{\raggedright\arraybackslash}X
}
\toprule
\textbf{Category} &
\textbf{Model} &
\textbf{Model Size} &
\textbf{Application Scenario} &
\textbf{Edge/Mobile Implication} \\
\midrule

% =========== Group 1: Lightweight & Compressed Models ===========
\tablegroupgray
\multicolumn{5}{l}{\textit{\textbf{Lightweight and Compressed Diffusion Architectures}}} \\
\addlinespace
Lightweight mobile DM & MobileDiffusion\par\cite{zhao2024mobilediffusion} & $<$400M parameters; one-step generation & Text-to-image generation, controllable generation, style/object LoRA, and inpainting & On-device latency: 0.2 s; enables instant local generation. \\

Lightweight mobile DM & SnapFusion\par\cite{li2023snapfusion} & Text encoder: 123M; UNet: 848M; decoder: 13M & Mobile text-to-image generation at $512 \times 512$ resolution & Mobile latency: 1.96 s; reduces diffusion to 8 denoising steps. \\

Lightweight DiT backbone & U-ViT\par\cite{Efficient_Model1} & U-ViT-M: 131M; U-ViT-L: 287M; U-ViT-H: 501M $+$ 84M autoencoder & Unconditional, class-conditional, and text-to-image generation & Simplified tokenized backbone; easier hardware-friendly deployment. \\

Edge-oriented DiT compression & SnapGen++\par\cite{Efficient_Model2} & On-device model: 0.4B; full model: 1.6B & High-fidelity text-to-image generation at approximately $1024^2$ resolution & Mobile latency: 1.8 s; elastic sub-models fit heterogeneous devices. \\

Post-training quantized DM & PTQD\par\cite{he2023ptqd} & LDM-4: 1603.35 MB FP32; 430.06 MB W8A8; 234.51 MB mixed / W4A4 & Compressed image diffusion inference & Low-bit quantization; lowers memory footprint and compute cost. \\

\midrule
% =========== Group 2: DLMs & Hybrids ===========
\tablegroupgray
\multicolumn{5}{l}{\textit{\textbf{DLMs and Hybrids}}} \\
\addlinespace
Block / hybrid DLM & Block Diffusion\par\cite{Efficient_Model3} & Backbone-dependent; block size $L' \in \{4,8,16,128\}$ & Flexible-length text generation and language modeling & Block-wise decoding; supports KV cache and variable-length text. \\

Training-free DLM acceleration & Fast-dLLM\par\cite{wu2025fast} & No extra parameters; applied to LLaDA and Dream & Diffusion LLM inference for reasoning, code, and text generation & KV cache + parallel decoding; accelerates practical DLM inference. \\

MoE DLM & Diffusion\hspace{0pt}Gemma\par\cite{google2026diffusiongemma_model,google2026diffusiongemma_blog} & 26B / about 4B active & Text generation & MoE lowers active compute; memory-bound. \\

Hybrid autoregressive-DM language model & Evo\par\cite{Efficient_Model4} & Evo-8B & Text generation, reasoning, code generation, and language understanding & Adaptive autoregressive-diffusion balance; avoids unnecessary refinement steps. \\

\midrule
% =========== Group 3: Sparse & Distributed ===========
\tablegroupgray
\multicolumn{5}{l}{\textit{\textbf{Sparse MoE and Distributed Architectures}}} \\
\addlinespace
Sparse MoE DM & Switch-DiT\par\cite{Efficient_Model5} & DiT-B: 131M; Switch-DiT-B: 137M--151M; common setting: 144M & Unconditional and class-conditional image generation & Timestep-aware sparse routing; activates fewer experts per step. \\

Billion-scale sparse DiT & DiT-MoE\par\cite{Efficient_Model6} & 199M/71M active to 16.5B/3.1B active parameters & Large-scale class-conditional image generation & Large total capacity; only sparse experts are activated during inference. \\

Distributed diffusion serving & DistriFusion\par\cite{li2024distrifusion} & Distributed execution across multiple GPUs/devices; no retraining & High-resolution image generation with SDXL-style DMs & Patch-parallel inference; distributes high-resolution generation across devices. \\

\bottomrule
\end{tabularx}
\vspace{4pt}
\raggedright
\footnotesize
%\scriptsize
\end{sidewaystable}

\noindent \textbf{U-ViT Architecture.}
For feature mapping in visual and multimodal generation, U-ViT proposes a streamlined universal Transformer. It treats timesteps, conditional signals, and noised sequence patches as uniform tokens. Long skip connections between shallow and deep layers replace the convolutional downsampling and upsampling modules of U-Nets, reducing complexity while maintaining feature representation~\cite{Efficient_Model1}.

\noindent \textbf{On-Device Extreme Compression (SnapGen++).}
For computation-limited mobile devices, SnapGen++ demonstrates strong lightweight generation. Architecture-level optimization and efficient diffusion training reduce DiT parameters to 0.4 billion (0.4B), achieving high-fidelity generation with 1.8-second latency~\cite{Efficient_Model2}. Diffusion-specific post-training quantization (PTQ) further compresses weights to low-bit representations, fitting low-memory edge gateways~\cite{he2023ptqd}.

\subsubsection{Block Diffusion and Hybrid Autoregressive Diffusion Models}
% \added{Lightweight diffusion transformers reduce the cost of individual backbones, but they do not fully resolve the structural mismatch between parallel diffusion and flexible-length language generation. Block and hybrid autoregressive-diffusion models address this issue by changing the decoding organization itself, combining diffusion-style parallelism with autoregressive-style length flexibility and cache reuse.}

% Autoregressive models capture long-range dependencies but suffer linear decoding latency. DMs enable parallel decoding but struggle with variable-length generation and exact likelihood modeling~\cite{li2022diffusion}. Hybrid designs reconcile efficiency and flexibility, while MDLMs improve language generation through masking loss matrices~\cite{sahoo2024simple}.

Lightweight diffusion Transformers reduce the cost of individual backbones, but they do not fully resolve the structural mismatch between parallel diffusion and flexible-length language generation. Autoregressive models capture long-range dependencies and support flexible-length decoding but suffer from sequential generation latency, whereas DMs enable parallel decoding but face challenges in variable-length generation and exact likelihood modeling~\cite{li2022diffusion}. Block and hybrid autoregressive-DMs therefore address this mismatch by reorganizing the decoding process to combine diffusion-style parallelism with autoregressive-style length flexibility and cache reuse.

\noindent \textbf{Block Diffusion Architecture.}
Block Diffusion addresses fixed-length limits by interpolating discrete denoising diffusion with autoregressive mechanisms. It supports flexible-length generation and combines KV caching with adaptive parallel token sampling during inference~\cite{wu2025fast}. Block-wise parallel decoding preserves diffusion acceleration, removes redundant long-sequence computation, and lowers single-pass memory footprint~\cite{Efficient_Model3}. Fully masked DMs, such as LLaDA, validate scalable parallel prediction on large datasets~\cite{nie2025large}.

\noindent \textbf{Evolution-Balanced Hybrid Latent Trajectory (Evo).}
Evo mathematically unifies autoregressive and diffusion paradigms in a Duality Latent Trajectory Model. It views text generation as a continuous evolutionary flow, assigning each token $t_i \in [0,1]$ to represent semantic maturity. Low-maturity tokens receive autoregressive-style high-confidence local refinement, while high-maturity tokens trigger diffusion-style global parallel planning. This balance adaptively trades off inference uncertainty and computational efficiency~\cite{Efficient_Model4}.

\subsubsection{Sparse and MoE Diffusion Models}
% \added{While block and hybrid designs improve decoding flexibility, they still rely on relatively fixed model capacity during each inference pass. Sparse and MoE diffusion models provide another architectural direction by increasing total model capacity while activating only a subset of parameters for each timestep, token group, or expert route.}

% Diffusion timesteps differ greatly in task complexity, so static dense networks waste computation. Sparse activation and MoE decouple total parameters from per-step inference FLOPs. Token-partitioned distributed inference also supports collaborative computing across edge devices~\cite{li2024distrifusion}.

While block and hybrid designs improve decoding flexibility, they still rely on relatively persistent model capacity during each inference pass. Sparse and MoE diffusion models provide another architectural direction by increasing the total model capacity while activating only a subset of parameters for each timestep, token group, or expert route. Because diffusion timesteps differ in task complexity, static dense networks may waste computation, whereas sparse activation decouples the total parameter capacity from per-step inference FLOPs~\cite{Efficient_Model5, Efficient_Model6}. Token-partitioned distributed inference can further support collaborative computing across edge devices~\cite{li2024distrifusion}.

\noindent \textbf{Sparse Routing for Collaborative Denoising (Switch-DiT).}
Switch-DiT treats denoising timesteps as independent tasks and inserts Sparse Mixture-of-Experts (SMOE) into each Transformer block. Its Diffusion Prior Loss routes similar denoising tasks through shared experts while isolating conflicting parameters. This timestep-aware routing improves complex multimodal generation without increasing single-pass computational overhead~\cite{Efficient_Model5}.

\noindent \textbf{Billion-Scale Sparse Diffusion (DiT-MoE).}
DiT-MoE enables large-model emergence under bounded inference overhead through Shared Expert Routing and Expert-level Balance Loss. Empirical results show temporal expert preferences: early denoising queries low-frequency global-feature experts, whereas late denoising uses high-frequency detail experts. For mobile edge agents, such sparse activation is potentially beneficial because only a subset of experts is activated for each input, reducing inference computation and offering a configurable trade-off between model capacity and memory requirements~\cite{Efficient_Model6}.

\subsection{Efficient Training Strategies}

Mobile edge agent systems face strict computation, memory, and communication constraints, making resource-efficient training and fine-tuning essential for scalable deployment. Current research addresses resource-efficient DLM training at three complementary levels: (a) autoregressive-to-diffusion adaptation reduces the cost of model initialization by reusing pretrained knowledge, (b) parameter-efficient fine-tuning (PEFT) limits the number of parameters updated during adaptation, and (c) federated diffusion coordinates training across distributed and privacy-sensitive data sources. Together, these strategies target initialization cost, local update efficiency, and cross-device training coordination, respectively.

\subsubsection{Autoregressive-to-Diffusion Adaptation and Joint Training}

Training large DLMs from scratch is costly. Since pre-trained autoregressive models already encode rich sequential priors, adapting them to diffusion offers a resource-efficient path~\cite{13gong2024scaling}.

A key strategy unifies discrete token prediction and continuous denoising. Transfusion jointly trains text and image generation in one Transformer backbone by combining next-token prediction with diffusion losses, improving scalability without heterogeneous networks~\cite{Efficient_Training1}. Block Diffusion addresses fixed-length limits through interpolation-based adaptation between discrete denoising and autoregressive generation, improving optimization with KV-caching and parallel sampling~\cite{Efficient_Model3}. Evo dynamically switches between high-confidence autoregressive refinement and global diffusion planning through evolutionary variational inference~\cite{Efficient_Model4}. These methods reuse foundation weights and avoid the energy cost of de novo training.

\subsubsection{PEFT Mechanisms}
% \added{Reusing pretrained autoregressive weights reduces the need for de novo DLM training, but full-parameter adaptation can still be prohibitively expensive on resource-constrained devices. PEFT therefore targets the next bottleneck by restricting parameter updates to lightweight trainable components.}

% Low-Rank Adaptation (LoRA) freezes pre-trained weights and inserts trainable low-rank matrices into the Transformer bypass, reducing trainable parameters by orders of magnitude while matching or surpassing full fine-tuning~\cite{Efficient_Training4}. DiffFit targets diffusion spatio-temporal dynamics by fine-tuning only bias terms and lightweight spatial scaling factors instead of dense layer approximations, reducing storage and gradient costs for edge domain transfer~\cite{Efficient_Training5}. Combining PEFT with training-free acceleration solvers, such as DPM-Solver, or Consistency Models enables high-quality generation in 1 to 10 steps, supporting ultra-low-latency interactive agents~\cite{lu2022dpmsolver, song2023consistency}.

Reusing pretrained autoregressive weights reduces the need for de novo DLM training, but full-parameter adaptation can still be prohibitively expensive on resource-constrained devices. PEFT targets the next bottleneck by restricting parameter updates to lightweight trainable components. Low-Rank Adaptation (LoRA) freezes pre-trained weights and inserts trainable low-rank matrices into the Transformer bypass, reducing trainable parameters by orders of magnitude while matching or surpassing full fine-tuning~\cite{Efficient_Training4}. DiffFit targets diffusion spatio-temporal dynamics by fine-tuning only bias terms and lightweight spatial scaling factors, instead of dense layer approximations, reducing storage and gradient costs for edge domain transfer ~\cite{Efficient_Training5}.

\subsubsection{Federated Diffusion Training Strategies}
% \added{PEFT and federated diffusion address resource-efficient training at different levels. While PEFT reduces the number of parameters updated during local or domain-specific adaptation, federated training coordinates learning across distributed and privacy-sensitive data sources. The two directions are therefore orthogonal but complementary: the former targets local adaptation cost, whereas the latter addresses cross-device coordination, data locality, and communication overhead.}

% In mobile edge networks, agent data are distributed and privacy-sensitive. Federated Learning (FL) with DMs supports privacy-preserving training, but non-independent and identically distributed (non-IID) data, communication latency, and edge energy remain major barriers.

PEFT and federated diffusion address resource-efficient training at different levels. While PEFT reduces the number of parameters updated during local or domain-specific adaptation, federated training coordinates learning across distributed and privacy-sensitive data sources. The two directions are distinct but complementary: the former targets local adaptation cost, whereas the latter addresses cross-device coordination, data locality, and communication overhead~\cite{sun2024exploring,li2025edge}. In mobile edge networks, however, federated diffusion still faces major challenges arising from non-independent and identically distributed (non-IID) data, communication overhead, and heterogeneous edge resource constraints~\cite{Efficient_Training7}.

To handle data heterogeneity, recent frameworks locally fine-tune DMs to synthesize high-quality data, aligning global distributions without exposing client data and reducing typical federated learning (FL) performance gaps~\cite{Efficient_Training6}. FedDiff introduces one-shot federated diffusion to reduce eavesdropping risks and multi-round synchronization latency. With Differential Privacy (DP) and Fourier Magnitude Filtering (FMF), it extracts task features in heterogeneous settings, accelerates convergence, and provides rigorous privacy bounds~\cite{Efficient_Training7, dockhorn2022differentially}. On-demand quantization further adapts weight compression and wireless transmission under battery and bandwidth limits, enabling energy-efficient federated aggregation without degrading model precision~\cite{Efficient_Training8}.

\subsection{Efficient Inference Methods}

To deploy DLMs on mobile edge agents, inference must meet strict latency and computation limits. Diffusion generation offers acceleration opportunities distinct from autoregressive models. This section organizes DLM inference acceleration along three complementary dimensions. Parallel decoding reduces serial dependence across token positions by updating multiple tokens jointly; few-step diffusion shortens the denoising trajectory by reducing refinement rounds or function evaluations; and speculative draft--verify strategies seek to reduce expensive target-model computation through lightweight proposal and verification.

\subsubsection{Parallel Decoding}

Unlike autoregressive language models that generate tokens left to right, DLMs can update multiple tokens simultaneously during reverse denoising~\cite{wu2023ar,gong2022diffuseq}. MaskGIT establishes this paradigm with a bidirectional Transformer and dynamic mask scheduling, generating all tokens in a constant number of steps~\cite{chang2022maskgit}. Extending this non-autoregressive design, the autoregressive-diffusion framework uses multi-level diffusion with position-dependent denoising steps~\cite{wu2023ar}. Left tokens undergo fewer steps, emerge earlier, and guide right-token generation, preserving parallel decoding while implicitly modeling language dependencies~\cite{wu2023ar}. SSD-LM further adopts semi-autoregressive block-wise generation, enabling parallel context updates within each block and bridging full parallel generation with serial decoding~\cite{han2023ssd}.

\subsubsection{Few-Step Diffusion}
% \added{Parallel decoding increases the amount of sequence progress made within each model evaluation by updating multiple token positions jointly. However, generation may still require many iterative refinement rounds. Few-step diffusion therefore targets a different bottleneck by shortening the denoising trajectory itself. These two directions are complementary: parallel decoding reduces serial dependence across token positions, whereas few-step diffusion reduces the number of denoiser evaluations required for generation.}

% Standard DMs require hundreds or thousands of Markov-chain steps for one high-quality sample, creating a major inference bottleneck~\cite{song2020denoising,salimans2022progressive}. Since each denoising step usually requires one forward pass of the denoising network, inference cost is commonly characterized by the number of function evaluations (NFEs)~\cite{lu2022dpmsolver}. Few-step diffusion reduces this cost by compressing the sampling trajectory and decreasing the number of denoiser forward passes. 

Parallel decoding increases the amount of sequence progress made within each model evaluation by updating multiple token positions jointly. However, generation may still require many iterative refinement rounds, and standard DMs can require hundreds or thousands of Markov-chain steps for one high-quality sample~\cite{song2020denoising,salimans2022progressive}. Few-step diffusion targets a different bottleneck by shortening the denoising trajectory itself. Since each denoising step usually requires one forward pass of the denoising network, inference cost is commonly characterized by the number of function evaluations (NFEs)~\cite{lu2022dpmsolver}. These two directions are complementary: parallel decoding reduces serial dependence across token positions,
whereas few-step diffusion reduces the number of denoiser evaluations required for generation.

Denoising Diffusion Implicit Models (DDIMs) generalizes the forward process to a non-Markovian form while keeping the standard training objective, enabling deterministic generation with 10x to 50x wall-clock acceleration~\cite{song2020denoising}. Progressive distillation iteratively distills a deterministic sampler into models needing half as many sampling steps, reducing samplers from up to 8192 steps to as few as 4 steps with limited perceptual degradation~\cite{salimans2022progressive}.

Beyond DDIM and progressive distillation, solver- and consistency-based approaches provide additional routes to few-step generation. DPM-Solver reduces sampling cost through high-order ordinary differential equation (ODE) solvers, enabling high-quality generation with a small number of NFEs, while Consistency Models learn mappings that support one-step or few-step generation~\cite{lu2022dpmsolver, song2023consistency}.

For discrete text generation, data-distribution ratio estimation removes dependence on autoregressive distribution annealing, enabling flexible compute-quality trade-offs and high-fidelity generation with fewer sampling steps~\cite{Language_diffusion5}. FM provides simulation-free training with CNFs and can use Optimal Transport (OT) displacement interpolation to form straight conditional probability paths, improving generation and sampling efficiency~\cite{Diffusion_Models1}. Latent Consistency Models (LCMs) directly predict solutions of augmented probability flow ODEs in latent spaces, achieving high-fidelity generation in 2 to 4 steps or even one step~\cite{luo2023latent}.

\subsubsection{Emerging Speculative Decoding Directions}

% \added{Few-step diffusion compresses the denoising trajectory globally, but it may require retraining, distillation, or carefully designed solvers. A complementary inference-time strategy is speculative decoding, which accelerates generation by using a lightweight draft process to propose candidate tokens or intermediate states before verification by a stronger model. Thus, few-step diffusion reduces the length of the trajectory, whereas speculative decoding reduces the cost of traversing or verifying that trajectory.}

% Speculative decoding accelerates inference by letting a small draft model generate multiple token suggestions in parallel and then using a larger target model for simultaneous verification~\cite{leviathan2023fast}. It preserves the target model's output distribution~\cite{leviathan2023fast}. Although designed for autoregressive transformers, this draft-and-verify principle can extend to diffusion by approximating simpler subtasks to skip or accelerate selected denoising steps without weakening the primary generator~\cite{leviathan2023fast}. 

Few-step diffusion compresses the denoising trajectory globally, but may require retraining, distillation, or carefully designed solvers. A complementary emerging inference-time direction is speculative decoding, where a lightweight draft model proposes multiple candidate tokens or intermediate states and a stronger target model performs parallel verification~\cite{leviathan2023fast}. In autoregressive settings, this draft-and-verify strategy can preserve the target model's output distribution while reducing expensive target-model computation. While originally designed for autoregressive Transformers, the principle may be extended to diffusion by approximating simpler subtasks or selected intermediate states to accelerate parts of the denoising process. Thus, few-step diffusion reduces the length of the trajectory, whereas speculative approaches seek to reduce the cost of traversing or verifying that trajectory.

For distributed edge systems, lightweight drafting and initial trajectory prediction can run on IoT devices, while costly parallel verification is offloaded to edge servers, reducing end-to-end latency.

\subsection{Model Compression}
Despite strong generation quality, DLMs remain difficult to deploy on mobile edge agents because of their large parameter scales, memory-bandwidth demand, and costly iterative denoising. Model compression reduces storage, computation, and runtime memory costs with limited quality loss, making it essential for edge deployment. Since model compression has been extensively studied in diffusion and language models, this subsection only outlines its relevance to edge DLMs, while detailed techniques can be found in dedicated surveys on efficient DMs, diffusion quantization, and LLM compression~\cite{shen2025efficient, ma2025efficient, zeng2025diffusion, zhu2024survey}.

\textbf{Quantization} compresses weights and activations into low-bit formats, such as INT8 and INT4, but DMs require timestep-aware treatment because activation distributions vary along the denoising trajectory~\cite{li2023q, he2023ptqd}. 

\textbf{Distillation} transfers sampling trajectories or guidance behavior from large teacher models to lightweight students, reducing denoising steps or merging CFG passes for faster conditional generation~\cite{salimans2022progressive, meng2023distillation}. 

\textbf{Pruning and sparsity} remove redundant parameters, substructures, or denoising steps; methods exploiting spatial-frequency or temporal-gradient redundancy indicate useful directions for compressing Transformer-based DLMs under edge resource constraints~\cite{yang2023diffusion, fang2023structural}.

% section 3
\subsection{Lessons Learned}
% Section 3
Resource-efficient DLMs must balance denoising depth, memory footprint, and generation quality, rather than optimize parameter count alone~\cite{peng2025efficient,shen2025efficient}. Lightweight blocks, sparse routing, compression, and adaptive decoding can reduce deployment cost, but may introduce semantic drift or hardware complexity~\cite{Efficient_Model3,Efficient_Model5,wu2025fast}. Future edge DLMs should prioritize memory-bandwidth reduction, cache reuse, adaptive steps, and quality reports under fixed latency, energy, and device settings.

\FloatBarrier

%
%
%
%
%%%%%%%%%%%%%%%%%%%%%%%%%%%%%%%%%%%%%%%%%%%%%%%%%%%%%%%%%%%%%%%%%%%%%%%%%%%%%%%%%%%%%%%%%%%%%%%%%%%%%%%%%%%%%%%%%%%%%%%%%%%%%%%
\section{DLM-Native Edge and IoT Agent Deployment}\label{sec4}

Although DLMs are increasingly relevant to mobile edge agentic AI, direct studies on DLM deployment in edge and IoT systems remain scarce. This section takes a DLM-centered perspective: it uses system-level schemes developed mainly for standard DMs as deployment references and discusses how they should be adapted for future DLM agent execution. The key insight is that device-only execution and lightweight on-device diffusion provide useful principles for reducing local latency and memory cost~\cite{zhao2024mobilediffusion,zheng2025diffusion}. 
Meanwhile, edge-assisted inference, cloud--edge collaboration, distributed denoising, and adaptive resource management suggest architectural directions for partitioning iterative generation across heterogeneous devices and edge servers~\cite{li2024distrifusion,xu2025large}. 
On the other hand, DLMs introduce additional challenges, such as discrete token generation, long-context memory, KV-cache reuse, multimodal inputs, and multi-round agent interactions, which require DLM-specific deployment designs rather than direct reuse of visual diffusion serving pipelines~\cite{zheng2025review}. Accordingly, wireless and MEC conditions are treated here as deployment constraints rather than the main conceptual focus; the main focus remains how DLM mechanisms affect inference, memory, privacy, and interaction for edge agents.

\subsection{Device-Only and Edge-Assisted DLM Agent Inference}

Deploying DLM-based agents on mobile and IoT devices is limited, not only by computation, memory, energy, and wireless bandwidth, but also by the iterative token-refinement process itself. Standard diffusion systems mainly follow three patterns: local execution, device--edge workload splitting, and cloud--edge--device collaboration. Although not yet mature for DLMs, these patterns provide practical directions for edge DLM serving.

Device-only inference removes network dependence and enables local generation on smartphones, wearables, and standalone IoT nodes. Existing diffusion systems rely on architecture pruning, hardware-aware redesign, operator approximation, dynamic loading, and step reduction. To achieve device-only execution, researchers have explored optimizations across different system levels. At the architectural level, frameworks, such as SnapFusion~\cite{li2023snapfusion} and EdgeDiT~\cite{kodavanti2026edgedit}, focus on simplifying the backbone (e.g., UNet or DiT) through rigorous pruning and distillation techniques to meet mobile latency constraints. At the operator level, solutions, such as LUT-Diff~\cite{wang2025efficient}, bypass costly matrix multiplications altogether, relying instead on lookup tables and mobile co-scheduling to maintain efficiency. MobileDiffusion further enables single-step generation with a hybrid Diffusion-GAN design, and on-device video generation uses dynamic loading for large backbones under mobile memory limits~\cite{zhao2024mobilediffusion,kim2025device}. For DLMs, these schemes suggest lightweight token backbones, cache-aware loading, low-bit execution, fewer denoising steps, and early-exit strategies for local agent reasoning and generation.

For DLM agents, edge-assisted inference should be viewed not only as offloading heavy computation to edge servers, but also as deciding which denoising stages, token blocks, or reasoning states should remain on devices. Standard diffusion systems split workloads by denoising stages, intermediate activations, or feature extractors. Early denoising or shared semantic extraction can run at the edge, with personalized refinement on devices~\cite{yang2025efficient}. Dynamic splitting further adapts partition points to modality complexity, network state, and device capacity~\cite{liu2025reinforcement}. For DLMs, similar designs may support token-block, multimodal-feature, or reasoning-trajectory splitting, letting edge servers handle heavy denoising or fusion while devices preserve private prompts, local states, permission-sensitive context, and final actions.

Cloud-edge-device collaboration forms a three-tier system: the cloud provides high-capacity generation, the edge supports low-latency caching or preview inference, and devices handle interaction and personalization. For DLM agents, this hierarchy should be organized around agent-level objectives, such as time-to-first-action, context reuse, and safe final action decoding. DiffusionX uses lightweight edge preview generation before invoking cloud models for final rendering~\cite{wei2025diffusionx}. Other frameworks route tasks across local, edge, and cloud nodes according to latency, energy, and quality of service (QoS) constraints, while edge caching reduces redundant generation for repeated prompts~\cite{hu2024cloud,yao2025enhancing}. For DLM agents, this architecture supports iterative interaction: preliminary reasoning, cached context, or draft responses can run at the edge, while complex reasoning or multimodal synthesis is escalated to the cloud.

\subsection{Distributed and Pipeline DLM Agent Inference}
% \added{The deployment modes discussed above determine which computing tiers participate in DLM agent inference. Building on this placement perspective, distributed and pipeline inference further specify how denoising steps, token blocks, model components, or reasoning states can be scheduled and executed across these tiers.}

% Distributed diffusion inference partitions generation across heterogeneous nodes, including devices, edge servers, and device clusters. It addresses the mismatch between large models, iterative denoising, and limited edge resources. 

The deployment modes discussed above determine which computing tiers participate in DLM agent inference. Building on this placement perspective, distributed and pipeline inference specify how denoising steps, token blocks, model components, or reasoning states can be scheduled and executed across these tiers. Distributed diffusion inference partitions generation across heterogeneous nodes, including devices, edge servers, and device clusters, to address the mismatch between large models, iterative denoising, and limited edge resources.

Although most existing systems target standard DMs, their principles can guide DLM deployment for long sequences, multimodal inputs, and multi-agent reasoning. For DLM agents, the split boundary should be considered at token blocks, denoising rounds, reasoning states, and action-related outputs, rather than only at model layers.

Collaborative denoising splits diffusion steps between devices and edge servers. Typically, early compute-intensive denoising runs at the edge, and compact latent representations are sent to devices for task-specific refinement~\cite{du2023exploring}. This asymmetric allocation reduces device energy use, avoids full-data offloading, and supports privacy because devices share intermediate activations rather than raw prompts, conditions, or final outputs, as in CollaFuse~\cite{allmendinger2024collafuse}. Adaptive controllers tune local and offloaded denoising steps based on bandwidth, device compute capacity, and latency-quality preferences~\cite{kong2025distributed}. For DLMs, this suggests splitting denoising iterations, token blocks, reasoning traces, or multimodal latent states across device and edge nodes.

Pipeline diffusion inference raises throughput by overlapping computation and communication. Recent systems use patch-level or sequence-level parallelism, rather than simple layer-wise partitioning, to distribute high-resolution generation across nodes~\cite{fang2024pipefusion}. Since adjacent diffusion timesteps are often similar, systems can reuse stale activations while asynchronously transmitting updated features, hiding communication latency within the pipeline~\cite{li2024distrifusion}. For DLMs, similar designs may exploit redundancy across denoising steps, token refinement rounds, or intermediate reasoning states. This is especially useful when stable tokens or reusable agent states can be cached across interaction turns.

Distributed systems may also adapt parallelism across inference stages. Under CFG, conditional and unconditional branches can show time-varying discrepancies, motivating dynamic switching among data parallelism, pipeline parallelism, and serial execution~\cite{jung2026accelerating}. Operator-level optimizations, such as pipelined attention, double buffering, quantized communication, and graph compilation, reduce synchronization and kernel-launch overhead in high-speed edge-cloud environments~\cite{li2026fastusp}. These techniques matter for DLMs because long contexts and KV caches make communication and memory movement comparable to computation.

For reasoning-oriented DLMs, distributed inference can support test-time scaling. Edge servers can generate candidate reasoning trajectories in parallel, evaluate or stitch them with reward models, and return compact reasoning states to devices for final response generation~\cite{miles2026test}. This separates exploration from synthesis and may enable low-latency agentic reasoning under edge constraints.

\subsection{Agent-State-Aware Resource Management and Adaptive Serving}
% \added{Distributed and pipeline inference provide the structural basis for splitting DLM execution, but mobile edge environments require these splits to change over time. Resource management and adaptive serving therefore move from static partitioning to runtime decisions that account for wireless variation, battery status, task urgency, and agent-state reuse.}

% For DLM agents, mobile resource management must handle fluctuating wireless channels, user demands, battery budgets, heterogeneous devices, and the variable cost of iterative token refinement. For diffusion-based agents, scheduling balances latency, energy, privacy, and generation quality across device, edge, and cloud resources.

Distributed and pipeline inference provide the structural basis for splitting DLM execution, but static partitions are insufficient in mobile edge environments where
wireless channels, user demands, battery budgets, device capabilities, and agent states vary over time. Resource management and adaptive serving therefore move from static partitioning to runtime decisions over offloading, scheduling, caching, and refinement allocation. For diffusion-based agents, these decisions must balance latency, energy, privacy, and generation quality across device, edge, and cloud resources.

Diffusion offloading differs from binary offloading because generation can be split by denoising steps, model blocks, or intermediate representations. Mobile Edge Generation (MEG) keeps some denoising stages on devices and offloads heavier stages to edge servers or nearby devices~\cite{feng2024exploring,xu2025large}. These decisions adapt to wireless signal-to-noise ratio (SNR), latency, and energy constraints, while diffusion-based reinforcement learning supports offloading under complex mobile trajectories~\cite{xu2025large,xu2024phd}. For DLMs, similar mechanisms may decide whether token refinement, multimodal fusion, long-context reasoning, or final response synthesis runs locally or remotely.

Compute scheduling extends offloading to multi-user mobile scenarios. In vehicle-to-everything (V2X) and dense edge networks, diffusion-based reinforcement learning can schedule serverless containers across roadside units and mobile terminals~\cite{huang2025diffusion,11193881}. Two-timescale designs combine long-term model-state caching across base stations with short-term computation allocation for dynamic requests~\cite{liu2025two,yang2025diffusion}. For DLM agents, interactive continuity may require migrating model states, cached contexts, stable token states, or partial reasoning trajectories across edge nodes.

Adaptive serving adds runtime elasticity. Visual diffusion frameworks, such as FlexGen, adjust network width, inference steps, or execution paths according to real-time network and device conditions~\cite{li2024flexgen}. Energy-aware edge inference combines lightweight compression and adaptive quantization to avoid service disruption on low-power IoT devices~\cite{xie2025energy}. Future DLM serving should be cache-aware, privacy-aware, and context-aware because agentic DLMs must process persistent user context, multimodal observations, and multi-turn decisions under tight resource budgets. Thus, adaptive serving should be evaluated not only by communication cost, but also by interaction success, state reuse, and safe action completion.

\subsection{Implications for Edge-Native DLM Agent Deployment}

Existing edge diffusion systems provide useful but indirect guidance for DLM deployment~\cite{zheng2025diffusion}. The main adaptation is to reinterpret these systems through DLM-specific generation dynamics rather than treating DLMs as ordinary visual diffusion backbones. Their core mechanisms, including device-only compression, edge-assisted split inference, cloud--edge routing, collaborative denoising, pipeline parallelism, caching, and adaptive scheduling, can be transferred to DLMs at the architectural level. Nevertheless, DLMs differ from visual DMs on several important aspects. DLMs operate over discrete or hybrid token spaces and interact heavily with KV caches~\cite{li2025survey}. Additionally, they often rely on masked-token refinement~\cite{yu2025discrete} and must support long-context memory for multi-round reasoning and action planning~\cite{li2024survey}. To this end, future edge DLM systems cannot simply inherit legacy image-based diffusion deployment schemes. Instead, they must be redesigned to handle discrete token-level dynamics~\cite{yu2025discrete}. This adaptation must also account for the memory overhead of caching reasoning trajectories~\cite{li2025survey}, while ensuring secure device-edge-cloud state management for privacy-sensitive prompts~\cite{zheng2025diffusion}. This gap indicates a potential research direction: building deployment frameworks specifically designed for DLM-based mobile edge agents.

% section 4
\subsection{Lessons Learned}
% Section 4
Edge DLM deployment should choose among device-only, edge-assisted, and cloud-edge collaboration according to privacy, latency, energy, and bandwidth constraints~\cite{zheng2025review,zheng2025diffusion}. Existing diffusion offloading offers useful primitives, but visual-generation schemes cannot be directly reused for discrete tokens, long-context states, reasoning traces, and tool calls~\cite{du2023exploring,li2025survey}. A practical design keeps private prompts and final actions on-device, offloads heavy intermediate refinement when necessary, and adapts partitioning to network conditions, task risk, token uncertainty, and interaction latency.

\FloatBarrier

%
%
%
%
%%%%%%%%%%%%%%%%%%%%%%%%%%%%%%%%%%%%%%%%%%%%%%%%%%%%%%%%%%%%%%%%%%%%%%%%%%%%%%%%%%%%%%%%%%%%%%%%%%%%%%%%%%%%%%%%%%%%%%%%%%%%%%%%%%%%%%%%%%%%%%%%%%
\section{Adaptive DLM Agent Serving under Communication Constraints}\label{sec5}

This section treats communication efficiency as one deployment condition for DLM agents, rather than as the core objective by itself. The focus is on how DLM serving can adapt denoising depth, split execution, privacy exposure, and interaction latency under edge constraints.

\subsection{Communication-Aware DLM Agent Inference}

For edge DLM serving, limited device resources and stochastic wireless channel conditions primarily affect how iterative denoising, split inference, and interaction latency should be managed. Traditional centralized inference often incurs high communication latency and bandwidth consumption. To address these challenges, communication-aware inference leverages model partitioning and collaborative generation to optimize performance within wireless environments while preserving agent-level goals, such as time-to-first-action, privacy exposure, and action reliability.

\subsubsection{DLM-Oriented Model Partitioning}

Partitioning strategies reconfigure DMs to align with heterogeneous mobile hardware and fluctuating wireless states. To minimize communication costs and satisfy dynamic latency constraints, inference frameworks optimize end-edge collaboration through fine-grained subgraph partitioning and adaptive, bandwidth-aware model width scaling~\cite{huo2025multi,li2024flexgen}. Furthermore, spatial and temporal techniques, such as patch-based diffusion and split learning, facilitate elastic model distribution across mobile edge nodes~\cite{li2026toward}. For Transformer-based architectures (e.g., U-ViT), multi-exit mechanisms allow mobile devices to adaptively truncate the denoising process during network congestion, mitigating wireless channel pressure by selecting optimal exit points~\cite{11075534}. For DLMs, partitioning should further consider token blocks, denoising rounds, reasoning states, and tool/action boundaries rather than only the model layers.

\subsubsection{Collaborative DLM Agent Generation}

DLM-oriented model partitioning defines the execution boundaries, at which computation and intermediate states can be placed across devices, edge servers, and cloud resources. Building on these boundaries, collaborative generation coordinates when and what each tier executes and exchanges during the generation process. Thus, partitioning primarily addresses the placement of computation, whereas collaborative generation focuses on cross-tier orchestration of partial denoising results, intermediate states, and semantic outputs.

In practice, such collaborative generation can optimize task allocation across the mobile-edge-cloud hierarchy to reduce wireless data traffic. Decoupling the diffusion process into cloud-side semantic planning and edge-side visual refinement reduces round-trip communication~\cite{yan2024hybridsdedgecloudcollaborative}. In scenarios involving multi-round prompt evolution, DiffusionX employs lightweight on-device DMs for rapid previews while invoking cloud-based high-fidelity refinement only after the prompt is finalized, thereby reducing repeated cloud-side generation and avoiding unnecessary device--cloud interactions during iterative prompt refinement~\cite{wei2025diffusionx}. For dynamic environments, generative AI-assisted reinforcement learning (GAIRL) and transfer learning enable real-time optimization of offloading strategies and local denoising ratios~\cite{11075534}. Additionally, multi-stage relay diffusion facilitates efficient task migration across heterogeneous mobile nodes under stringent communication and energy constraints~\cite{li2026toward}. For DLM agents, these methods should be adapted to support draft-and-revise interaction and to avoid exposing permission-critical outputs before verification.

% In scenarios involving multi-round prompt evolution, DiffusionX employs lightweight on-device DMs for rapid previews while offloading high-fidelity finalization to the cloud, reducing latency from repetitive wireless transmissions~\cite{wei2025diffusionx}. 

\subsection{Agent-Aware Computation--Communication Co-Optimization}

Deploying DLMs within mobile agentic ecosystems requires a paradigm shift from static resource allocation to DLM-aware adaptive serving under edge constraints. Unlike traditional autoregressive models, DLMs operate under a multi-step iterative paradigm in which decoding complexity scales with the number of denoising rounds $T$ and sequence length $N$. ~\cite{peng2025efficient} show that this iterative generation is fundamentally memory-bound rather than purely compute-bound; repeated memory accesses across $T$ steps increase bandwidth overhead, and standard parallel decoding strategies exhibit diminishing returns when scaled. Consequently, deployment on resource-constrained edge devices creates a combined latency-memory constraint. Strict latency deadlines and the stochastic convergence of semantic tokens require adaptive denoising schedules that accelerate or truncate generation steps when appropriate. At the same time, the trade-off between mobile battery lifetime and memory-intensive high-fidelity decoding requires energy-aware inference mechanisms, such as elastic task offloading and distributed architectures, to reduce the local hardware burden.

\begin{figure}[!t]
\centering
\includegraphics[width=\linewidth]{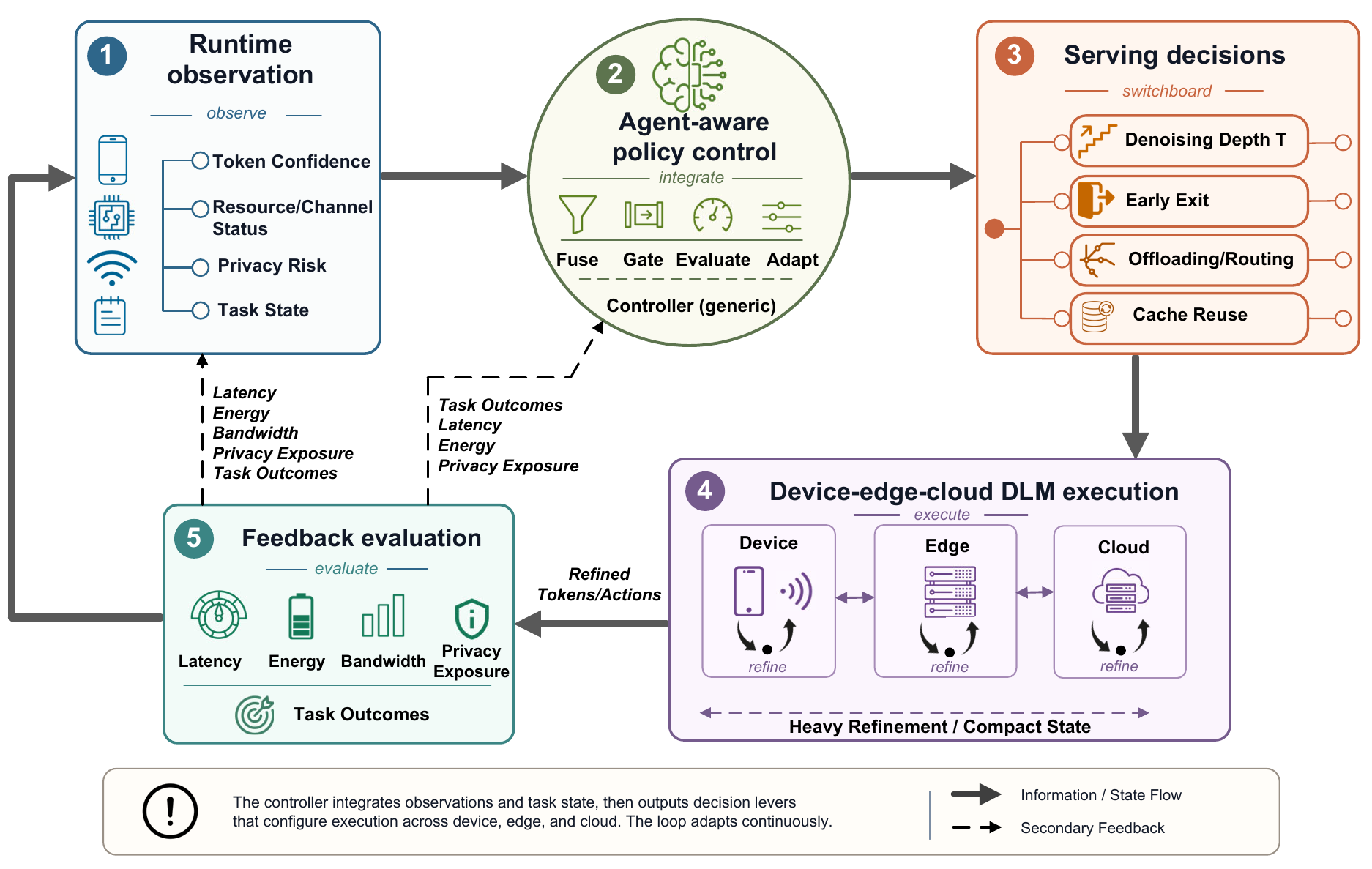}
\caption{Agent-aware adaptive DLM serving loop under communication constraints. Modules (1)--(5) denote runtime observation, policy control, serving decisions, tiered execution, and feedback evaluation}
\label{fig:adaptive_dlm_serving_loop}
\end{figure}

To clarify this dynamic serving process, Fig.~\ref{fig:adaptive_dlm_serving_loop} illustrates an agent-aware adaptive DLM serving loop under communication constraints. Instead of treating computation offloading, denoising scheduling, and cache reuse as independent decisions, the loop organizes runtime observation, policy control, serving decisions, device-edge-cloud execution, and feedback evaluation into a closed process. This closed-loop view explains how DLM agent serving can continuously adapt denoising depth, early exit, offloading/routing, and cache reuse, according to token confidence, resource/channel status, privacy risk, and task outcomes, thereby providing the basis for adaptive denoising schedules and energy-aware DLM agent inference, as will be discussed in the following subsections.

\subsubsection{Adaptive Denoising Schedules}

In edge environments, balancing interaction latency with generation quality is critical. For agents, adaptive denoising also determines how much refinement should be allocated to uncertain tokens, high-risk actions, or complex reasoning steps. Efficient DLM inference manipulates denoising depth based on the bits-to-rounds principle, which lower-bounds decoding rounds by the target sequence's total information content~\cite{fu2025bits}. To overcome the low per-round information gain of traditional decoding, ~\cite{fu2025bits} propose a training-free Explore-Then-Exploit (ETE) mechanism. By actively exploring high-information tokens to trigger confidence cascades and exploiting the induced high-confidence tokens, ETE reduces decoding rounds without compromising fidelity.

To navigate this information-theoretic boundary, modern frameworks employ semantic-aware truncation strategies that trade some refinement accuracy for lower latency. Since the latent variables in DLMs converge toward discrete word embeddings early in the reverse process, the system can terminate denoising once the entropy of the token distribution falls below a task-specific threshold. For example, the end-of-text prediction (EoTP) and learnable filtering models enable the system to halt computation on padding tokens or redundant iterations without compromising the final semantic output~\cite{bao2025learning}. Similarly, early-skipping mechanisms leverage intermediate tensor variations to bypass computationally expensive layers for low-importance tokens, reallocating saved cycles to meet strict interaction deadlines~\cite{zhu2026esdllm}. By integrating communication-aware guidance, the inference process can be guided toward outputs that are robust to packet loss or bandwidth fluctuations, helping the agent remain responsive even under degraded network conditions~\cite{he2025communication}. This adaptive scheduling transforms inference from fixed-length execution into a flexible process conditioned on token confidence, task urgency, and network state~\cite{liu2024joint}.

\subsubsection{Energy-Aware DLM Agent Inference}
% \added{Adaptive denoising schedules control how much refinement is performed for a given task, but they do not determine where this refinement should be executed. In mobile edge settings, the same denoising budget may lead to different energy and latency costs depending on whether it is processed locally, offloaded to an edge server, or distributed across nearby devices. Energy-aware DLM inference therefore complements adaptive scheduling by jointly deciding the depth and placement of denoising computation.}

% Multi-step DLM inference can increase energy consumption on mobile terminals, especially when repeated denoising causes high memory access and computation cost. Addressing this trade-off requires architectural elasticity, which allows the generative workload to be decoupled and distributed across the device-edge continuum. 

Adaptive denoising schedules control how much refinement is performed for a given task, but they do not determine where this refinement should be executed. In mobile edge settings, the same denoising budget may lead to different energy and latency costs, depending on whether it is processed locally, offloaded to an edge server, or distributed across nearby devices. This issue is particularly important because multi-step DLM inference can increase mobile energy consumption through repeated memory access and computation. Energy-aware DLM inference complements adaptive scheduling by jointly considering denoising depth, computation placement, and device power budgets. Addressing this trade-off requires architectural elasticity across the device-edge continuum.

A key strategy for addressing the power-compute conflict is split diffusion, where the edge device only executes the lightweight forward noising phase, while the energy-intensive reverse denoising is offloaded to a proximal edge server~\cite{ai2026cross}.

This offloading decision is not a simple binary choice but a multi-objective optimization involving remaining battery life, transmission power, and privacy requirements. Under strict energy budgets, systems can use deep reinforcement learning (DRL) to solve mixed-integer nonlinear programming problems, dynamically determining the optimal split points, which, in the context of DLMs, dictate how much local noising, token refinement, or final decoding should remain on the device to balance communication overhead, data obfuscation, and system lifetime~\cite{zhou2021deep}. However, as deployments scale to heterogeneous IoT environments, this single-server paradigm becomes a bottleneck. To address this issue, multimodal parallel offloading can be implemented through content-aware semantic partitioning, which minimizes transmission overhead by isolating critical region proposals~\cite{zeng2025generative}. To navigate the larger action space of multi-node routing, integrating a generative diffusion module into the DRL policy enables dynamic allocation under fluctuating edge resources. For DLM agents, the routing policy should also account for token uncertainty, action risk, and privacy exposure.

\begin{figure}[!t]
    \centering
    \includegraphics[width=1.0\textwidth]{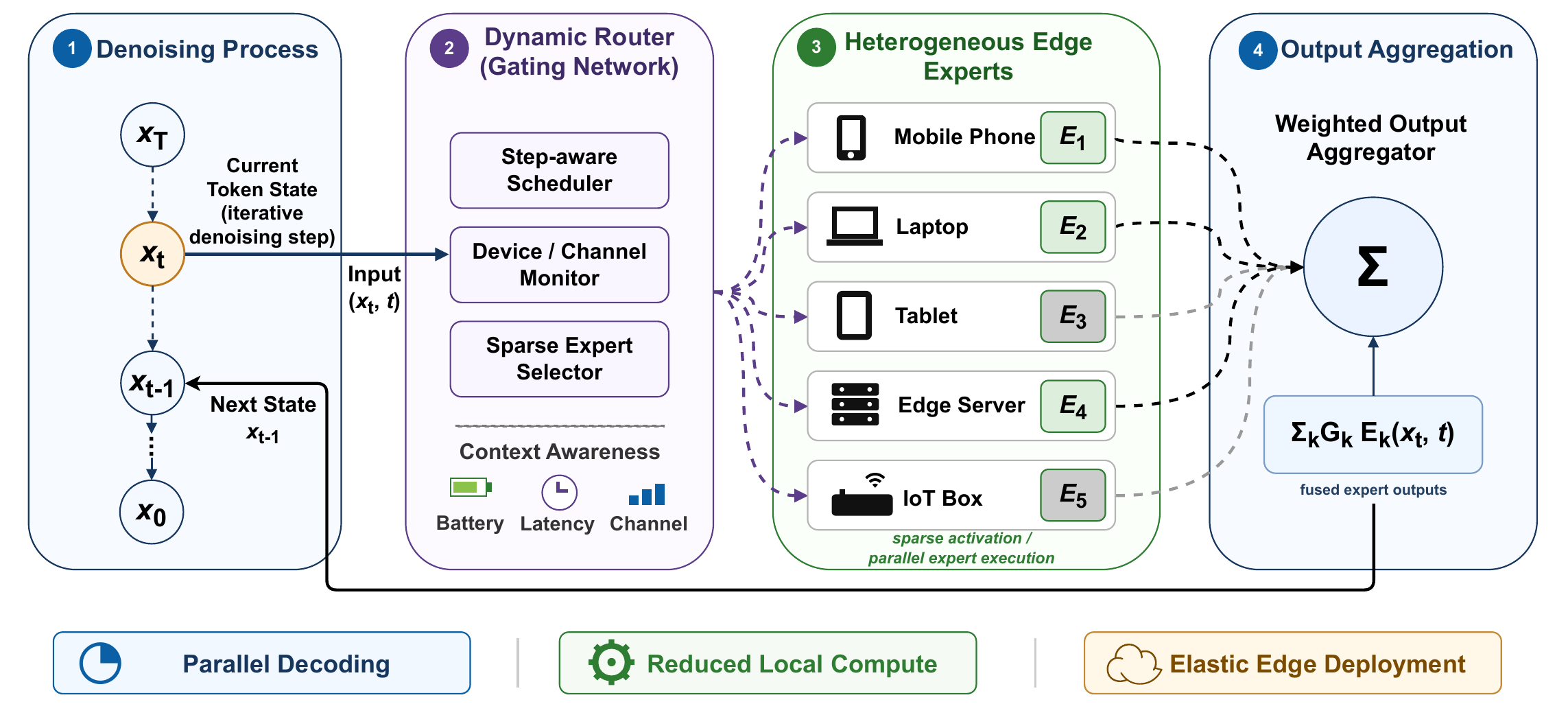}
    \caption{Distributed DLM-MoE architecture for sparse edge-agent collaborative decoding. Numbers (1)--(4) denote token-state input, dynamic routing, parallel edge experts, and output aggregation} 
    \label{fig:MoE_for_diffusion}
\end{figure}

Building on this multi-node foundation, the paradigm is shifting from simple task offloading toward multi-device collaborative decoding. While intra-device optimizations, such as the ETE mechanism, mitigate local compute limits, deploying the large parameter space of advanced DLMs requires architectural elasticity at the model level. A relevant approach is to apply the MoE mechanism directly to DMs. Previous studies have broadly explored the integration of MoE mechanisms with mobile edge computing~\cite{qin2025optimal,jin2026moe,gao2025towards}, proposing foundational solutions, such as hierarchical expert selection for device heterogeneity~\cite{jin2026moe}, joint optimization of routing and radio resources~\cite{qin2025optimal}, and privacy-preserving federated inference~\cite{gao2025towards}. Extending this paradigm to DLMs allows different specialized ``expert'' modules to be deployed across distinct proximal devices. By decoupling the dense generative backbone, the system can dynamically route intermediate token representations, or specific iterative denoising steps, to sparsely activated experts, enabling parallel processing without overloading single terminals. Fig.~\ref{fig:MoE_for_diffusion} provides a general technical architecture for sparse DLM execution through multi-device collaborative decoding. In one reverse denoising round, the current token state $(x_t,t)$ is routed by a gating module to heterogeneous edge experts, and the weighted expert outputs are aggregated to estimate $x_{t-1}$.

Beyond efficiency, elastic partitioning of DLMs creates a privacy-energy-performance trade-off. By performing the initial noising stages locally, the device can provide a noising-based obfuscation layer that reduces the exposure of the original discrete text before transmission~\cite{ai2026cross}. This can help ensure that the agent uses server-side compute without the high energy overhead of heavy encryption or the risk of exposing sensitive prompts. Thus, energy-aware inference in the DLM context requires a resource-allocation strategy in which the split point is continuously repositioned to balance device lifetime, privacy exposure, interaction latency, and generation quality~\cite{liu2024joint}.

% section 5
\subsection{Lessons Learned}
% Section 5
Adaptive DLM serving must jointly manage offloading, denoising depth, interaction latency, energy use, and privacy exposure~\cite{liu2024joint,xu2025large}. Fixed split points are fragile because mobile systems face channel fluctuation, mobility, contention, packet loss, and heterogeneous hardware~\cite{li2026toward,zheng2025diffusion}. Future systems should use communication-aware partitioning, adaptive denoising, and energy-aware collaboration to optimize time-to-first-action, selectively refine uncertain or risky tokens, and retain permission-critical steps locally.

\FloatBarrier

%
%
%
%
%%%%%%%%%%%%%%%%%%%%%%%%%%%%%%%%%%%%%%%%%%%%%%%%%%%%%%%%%%%%%%%%%%%%%%%%%%%%%%%%%%%%%%%%%%%%%%%%%%%%%%%%%%%%%%%%%%%%%%%%%%%%%%%%%%%%%%%%%%%%%%%%%%%
\section{Applications and Emerging Use Cases in IoT and Wireless Systems}\label{sec6}

Since direct DLM applications in IoT and wireless systems remain limited, this section first reviews continuous DMs as established precedents for diffusion-based wireless intelligence, and then discusses how DLMs may extend this progress toward discrete semantic and agentic tasks.

\subsection{Continuous DMs as Precedents for Wireless Intelligence}

% Continuous DMs have already shown the value of diffusion-style iterative refinement in wireless and IoT systems, especially for continuous signal modeling, resource optimization, trajectory generation, and network control. Existing studies have adapted DMs across diverse wireless and IoT domains. For network security, DMs have been used to synthesize scarce attack samples and improve intrusion detection systems~\cite{sanchez2026latent}. For physical and aerial operations, diffusion-based methods enable tasks such as forecasting electricity loads, restoring aerial imagery, and generating optimal unmanned aerial vehicle (UAV) trajectories~\cite{pan2025diffusion}. At the infrastructure level, these models have also been shown to be effective for complex radio resource management, particularly in Open Radio Access Network (O-RAN) and network slicing scenarios~\cite{yan2025xdiff}. These applications mainly operate on numerical features, physical dynamics, or resource variables rather than language tokens. Therefore, they should be viewed as system-level precedents for diffusion intelligence at the edge.

Continuous DMs have already shown the value of diffusion-style iterative refinement in wireless and IoT systems, especially for continuous signal modeling, resource optimization, trajectory generation, and network control. Recent surveys have systematically reviewed GDMs for wireless networks across sensing, transmission, application, and security dimensions, providing a broader context for the representative continuous-DM precedents summarized below~\cite{fan2026generative}. Against this broader landscape, representative studies have adapted DMs across diverse wireless and IoT domains. For network security, DMs have been used to synthesize scarce attack samples and improve intrusion detection systems~\cite{sanchez2026latent}. For physical and aerial operations, diffusion-based methods enable tasks such as forecasting electricity loads, restoring aerial imagery, and generating optimal unmanned aerial vehicle (UAV) trajectories~\cite{pan2025diffusion}. At the infrastructure level, these models have been shown to be effective for complex radio resource management, particularly in Open Radio Access Network (O-RAN) and network slicing scenarios~\cite{yan2025xdiff}. These applications operate on numerical features, physical dynamics, or resource variables rather than language tokens. They should be viewed as system-level precedents for diffusion intelligence at the edge.

\subsection{DLMs for Semantic Agentic Intelligence}
% \added{Continuous DMs mainly demonstrate how diffusion-style refinement can optimize signals, trajectories, and resource variables in wireless systems. DLMs extend this principle to discrete semantic and executable representations, making them more suitable for instruction repair, action-token synthesis, protocol configuration, and agent-level decision making.} As mobile edge environments evolve from data-driven optimization toward intent-driven semantic control, this discrete token modeling capability becomes particularly relevant for semantic reasoning, command repair, action-token synthesis, executable plan generation, and structured decision making.

Building on these precedents, DLMs extend diffusion-style refinement from continuous signals and resource variables to discrete semantic and executable representations. As mobile edge environments evolve from data-driven optimization toward intent-driven semantic control, this token-level modeling capability becomes particularly relevant for instruction repair, action-token synthesis, protocol configuration, executable plan generation, and agent-level decision making.

\begin{figure}[!t]
    \centering
    \includegraphics[width=\linewidth]{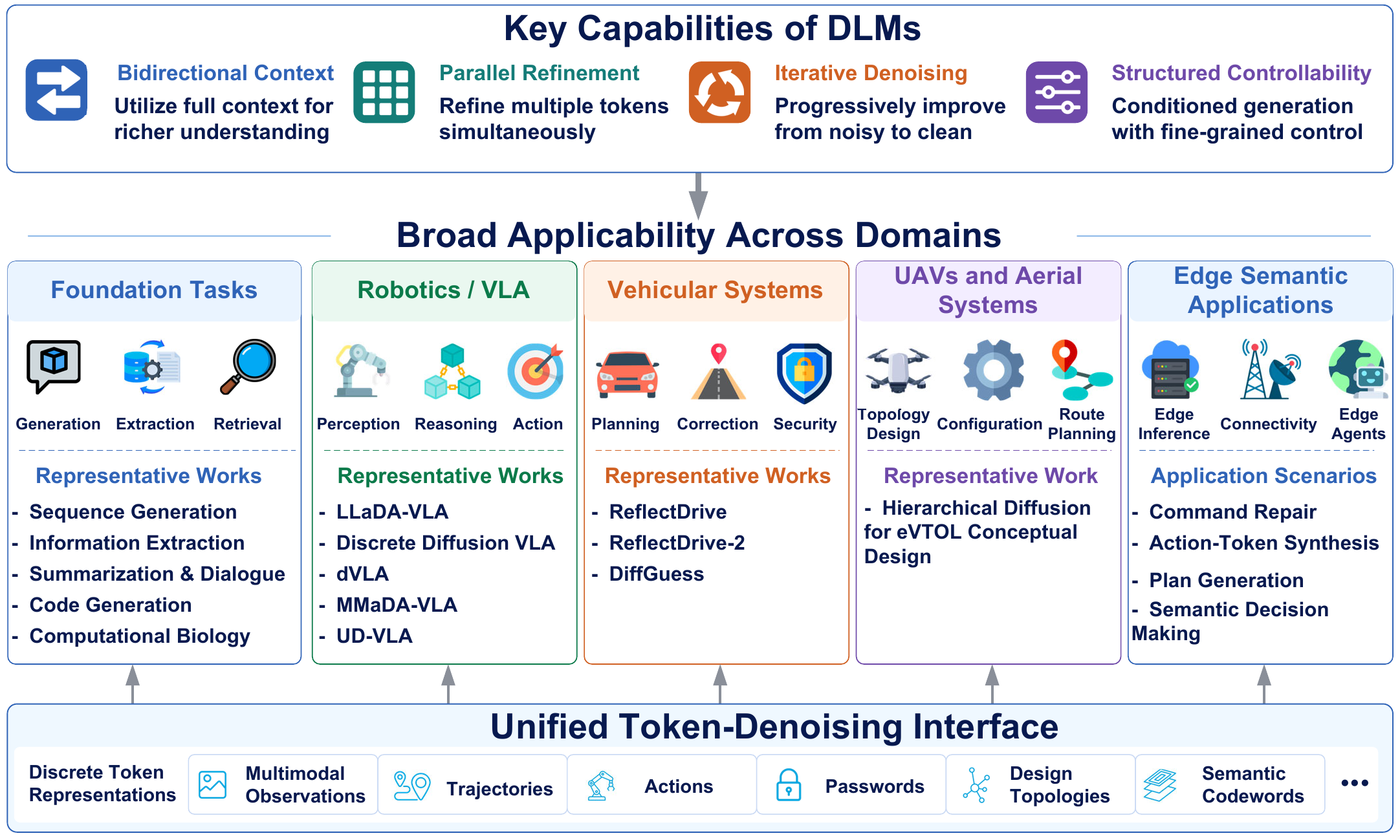}
    \caption{Applications of DLMs for semantic agentic intelligence through a unified token-denoising interface}
    \label{fig:dlm_applications}
\end{figure}

Existing DLMs studies have demonstrated broad applicability in conventional NLP tasks, code generation, and computational biology, covering sequence generation, information extraction, summarization, dialogue, retrieval, translation, molecular modeling, and protein design~\cite{li2025survey,reyes2024improving}. In addition, discrete language modeling and discrete diffusion offer a unified token-denoising interface that can extend diffusion-based generation beyond language-centric tasks. By representing multimodal observations, trajectories, actions, passwords, design topologies, or semantic codewords as discrete token sequences, DLM-style iterative refinement can support semantic reasoning, command repair, action-token synthesis, executable plan generation, and structured decision making for mobile edge agents under latency and resource constraints. Fig.~\ref{fig:dlm_applications} summarizes this capability--interface--application relationship. It first highlights four key DLM capabilities, namely bidirectional context, parallel refinement, iterative denoising, and structured controllability. These capabilities are then connected to a unified token-denoising interface over discrete token representations, through which multimodal observations, trajectories, actions, passwords, design topologies, and semantic codewords can be refined for different semantic agentic tasks. The resulting application map covers foundation tasks, robotics/vision-language-action (VLA), vehicular systems, UAV and aerial systems, and edge semantic applications.

In robotics, recent VLA studies provide direct evidence that DLMs can serve as semantic-to-action engines. LLaDA-VLA builds on pretrained diffusion-based vision-language models (VLMs) and adapts masked diffusion to robotic manipulation through localized special action-token classification and hierarchical action-structured decoding, enabling coherent action-sequence generation in simulation and real-robot tasks~\cite{wen2025llada_vla}. Discrete Diffusion VLA formulates action decoding as discrete diffusion over tokenized action chunks inside a unified Transformer, where high-confidence action tokens are fixed first and uncertain ones are re-masked for later refinement~\cite{liang2025discrete_vla}. Similarly, dVLA adopts a discrete diffusion language backbone with separate tokenizers for images, text, and actions, and introduces multimodal chain-of-thought generation to jointly optimize visual subgoals, textual reasoning, and robot actions~\cite{wen2025dvla}. More recent unified VLA models, such as MMaDA-VLA and UD-VLA, extend this direction by mapping language, images, and continuous robot controls into a shared discrete token space and synchronously denoising future visual observations and action chunks~\cite{liu2026mmada_vla,chen2026ud_vla}. These studies suggest that DLMs can transform robotic control from sequential action prediction into parallel, revisable, and semantically grounded action refinement.

In vehicular networks and autonomous driving, DLMs have begun to support trajectory-level semantic planning and safety-aware correction. ReflectDrive discretizes the two-dimensional driving space into an action codebook, fine-tunes pretrained DLMs for trajectory planning, and introduces a reflection mechanism that identifies unsafe trajectory tokens and regenerates them through gradient-free inpainting~\cite{li2025reflectdrive}. ReflectDrive-2 extends this direction with a decision--draft--reflect pipeline, where masked discrete diffusion generates editable trajectory drafts and AutoEdit rewrites selected tokens within the same discrete action space; it further couples drafting and editing through reinforcement learning and improves runtime efficiency with shared-prefix KV reuse and fused on-device unmasking~\cite{wang2026reflectdrive2}. Beyond driving control, DiffGuess applies a DiffuSeq-based DLM to V2X and intelligent transportation systems (ITS) password strength assessment by modeling password modification as an iterative diffusion process, showing that DLM-style refinement can also enhance authentication security in ITS~\cite{xiong2025diffguess}. These works demonstrate that DLMs can support vehicular agents, not only in planning and correction but also in semantic security and access-control reasoning.

For UAVs and aerial edge systems, existing DLM-related evidence is still preliminary but technically relevant. In eVTOL aircraft conceptual design, a hierarchical diffusion framework combines a Riemannian DLM with masked diffusion to sample discrete aircraft topologies and corresponding continuous parameters under simulation-based inference~\cite{ghiglino2026diffusion}. Although this work focuses on aircraft design rather than real-time UAV networking, it illustrates how DLM-like discrete diffusion can model topology-level design choices, variable-dimensional structures, and constraint-conditioned generation. This suggests a possible direction for future UAV edge agents, where mission plans, swarm formations, component configurations, and route constraints can also be represented as discrete structured tokens and refined through iterative denoising.

% Section 6
\subsection{Lessons Learned}

Continuous DMs have already demonstrated the value of diffusion-style refinement for numerical signals, physical dynamics, trajectory generation, resource allocation, and network optimization in IoT and wireless systems~\cite{pan2025diffusion,liang2025diffusion}. DLMs extend this paradigm toward discrete semantic and executable structures, making them potentially useful for command repair, structured API and tool generation, semantic coordination, protocol configuration, and intent translation~\cite{li2025survey,huh2026markov,habib2026generative}. Since current DLM applications in IoT and wireless systems remain early and often indirect, they should be viewed as semantic and action-level complements to continuous DMs, and practical edge DLM agents should integrate safety constraints, execution verification, closed-loop feedback, and real-device validation.

\FloatBarrier

%
%
%
%
%%%%%%%%%%%%%%%%%%%%%%%%%%%%%%%%%%%%%%%%%%%%%%%%%%%%%%%%%%%%%%%%%%%%%%%%%%%%%%%%%%%%%%%%%%%%%%%%%%%%%%%%%%%%%%%%%%%%%%%%%%%%%%%%%%%%%%%%%%%%
\section{Evaluation Metrics, Benchmarks, and Frameworks}
\label{sec:evaluation}

Existing evaluation protocols for LLMs traditionally emphasize static foundation capabilities. For instance, Massive Multitask Language Understanding (MMLU) and Holistic Evaluation of Language Models (HELM) baseline general reasoning and knowledge~\cite{1hendrycks2020measuring,2liang2022holistic}, while frameworks like AgentBench measure instruction following and coding proficiency~\cite{5liu2023agentbench}. However, mobile edge agentic AI requires a broader perspective that jointly considers generation mechanisms, interaction latency, resource constraints, and autonomous task execution. Consequently, recent benchmarks have introduced more interactive setups: Mobile-Bench evaluates multi-application workflows~\cite{7deng2024mobile}, AndroidWorld tests executable mobile UI states~\cite{8rawles2024androidworld}, and OSWorld incorporates process-aware rewards for open-ended environments~\cite{6xie2024osworld}. Furthermore, MLPerf Mobile highlights the importance of hardware-aware and reproducible deployment metrics~\cite{9janapa2022mlperf}. This section organizes evaluation around structure-aware comparison among autoregressive models, general DMs, and DLMs.

\subsection{Structure-Aware Evaluation with Emphasis on DLMs}

Structure-aware evaluation links generation mechanisms to deployability. Autoregressive models require sequential decoding and KV-cache management, so their edge performance depends on prefill latency, decoding latency, cache growth, and time-to-first-action. General DMs generate through iterative denoising, which makes sampling steps, function evaluations, energy per sample, and downstream utility important for mobile deployment~\cite{10ho2020denoising,11rombach2022high}. DLMs combine language generation with iterative refinement, bidirectional token prediction, and mask scheduling, so they should not be evaluated as ordinary LLM backbones.

For DLMs, high language accuracy alone is insufficient. Models, such as LLaDA and DiffuLLaMA, show strong language and reasoning potential~\cite{nie2025large,13gong2024scaling}, but practical efficiency can be weakened by repeated denoising, remasking instability, limited cache compatibility, and inconsistent latency protocols~\cite{peng2025efficient}. To address this issue, a structure-aware evaluation protocol for DLMs should jointly report model-level quality and generation-process behavior, including denoising-step efficiency, mask-schedule sensitivity, parallel decoding gain, early-exit quality, cache compatibility, and quality-latency-energy trade-offs.

A recent inference-only analysis of the public DiffusionGemma checkpoint illustrates why such process-level reporting is necessary: its token commitment is neither strictly parallel nor strictly sequential, but exhibits a task- and granularity-dependent left-to-right bias, large commit batches, remasking events, and entropy-based commitment behavior~\cite{asaria2026diffusiongemma}. For edge evaluation, this suggests that DLM reports should include commit-order statistics, same-call commit ratios, entropy or confidence signals, denoising-step budgets, and task-regime differences, rather than relying only on aggregate accuracy or throughput.

\subsection{Metrics and Benchmarks}
% \added{Given these structure-specific differences, evaluation metrics should not be selected as isolated accuracy scores. Instead, they should answer practical deployment questions: whether the model can reason, whether the generation process is stable, whether interaction succeeds, and whether the system remains efficient under edge constraints.}

% Rather than enumerating metrics separately from benchmarks, we group them by the practical questions they answer. Capability benchmarks measure whether the backbone can reason, follow instructions, code, or remain factual. Agent and mobile benchmarks measure whether the model can ground UI states, call tools, complete multi-step tasks, recover from errors, and handle dynamic feedback. Edge benchmarks measure whether these abilities remain usable under latency, memory, energy, thermal, and communication constraints.

Given these structure-specific differences, evaluation metrics should not be selected as isolated accuracy scores. Instead, they should answer practical deployment questions: whether the model can reason, whether the generation process is stable, whether interaction succeeds, and whether the system remains efficient under edge constraints. Accordingly, rather than enumerating metrics separately from benchmarks, we group them by the practical questions they answer. Capability benchmarks measure whether the backbone can reason, follow instructions, code, or remain factual~\cite{4chen2021evaluating}; agent and mobile benchmarks assess UI grounding, tool use, multi-step task completion, and error recovery; and edge benchmarks evaluate whether these capabilities remain usable under latency, memory, energy, thermal, and communication constraints.

For DLM-based mobile agents, benchmark reports should also specify generation-specific and system-specific settings, including denoising steps, mask schedule, output length, prompt length, batch size, cache strategy, quantization level, runtime, accelerator, and power measurement method. Without these details, claims about DLM speed, energy efficiency, or deployability are difficult to compare in a fair fashion, especially when similar aggregate scores may hide sample-level instability caused by diffusion steps, sampling strategies, batch size, hardware, or numerical precision~\cite{16fang2026dataset}.

\subsection{Evaluation Framework}
% \added{The metrics discussed above cover heterogeneous aspects of model capability, generation behavior, interaction, deployment, and safety. To make these aspects easier to apply in practice, we organize them into a layered evaluation framework.}

% Table~\ref{tab:evaluation_framework} summarizes the above discussion into a five-layer framework. The table is intended as a compact checklist rather than a fixed benchmark: it separates backbone capability, generation dynamics, mobile interaction, edge deployment, and safety, while keeping these dimensions comparable across autoregressive models, general DMs, and DLMs.

The metrics discussed above cover heterogeneous aspects of model capability, generation behavior, interaction, deployment, and safety. To make these aspects easier to apply in practice, we organize them into a layered evaluation framework. Table~\ref{tab:evaluation_framework} summarizes the above discussion into five layers covering backbone capability, generation dynamics, mobile interaction, edge deployment, and safety and reliability. The table is intended as a compact checklist, rather than a fixed benchmark, while keeping these dimensions comparable across autoregressive models, general DMs, and DLMs.

\begin{table}[t]
\centering
\caption{A structure-aware evaluation framework for mobile edge agentic AI.}
\label{tab:evaluation_framework}
\renewcommand{\arraystretch}{1.1}
\setlength{\tabcolsep}{4pt}
\footnotesize

\begin{tabularx}{\linewidth}{
    >{\hsize=0.85\hsize\raggedright\arraybackslash}X 
    >{\hsize=0.85\hsize\raggedright\arraybackslash}X 
    >{\hsize=1.3\hsize\raggedright\arraybackslash}X
}
\toprule
\textbf{Evaluation Dimension} & \textbf{Main Focus} & \textbf{Representative Elements} \\
\midrule

\textbf{Backbone Capability} &
Evaluate the intrinsic capability of the foundation model~\cite{1hendrycks2020measuring,2liang2022holistic,3lin2022truthfulqa} &
Knowledge, reasoning, coding, factuality, instruction following \\

\midrule

\textbf{Generation Dynamics} &
Evaluate how outputs are generated under different model structures~\cite{10ho2020denoising,nie2025large,peng2025efficient} &
\textbf{Autoregressive:} decoding latency, KV-cache cost \newline
\textbf{Diffusion:} denoising cost, sampling efficiency \newline
\textbf{DLMs:} step count, mask schedule, cache compatibility, early exit \\

\midrule

\textbf{Mobile Interaction} &
Evaluate closed-loop task execution in mobile environments~\cite{6xie2024osworld,7deng2024mobile,8rawles2024androidworld} &
UI grounding, tool/API use, planning, action execution, feedback, recovery \\

\midrule

\textbf{Edge Deployment} &
Evaluate deployability on mobile and edge platforms~\cite{9janapa2022mlperf} &
Latency, energy, memory, communication cost, thermal behavior, offloading \\

\midrule

\textbf{Safety and Reliability} &
Evaluate trustworthiness and robustness of mobile edge agents~\cite{3lin2022truthfulqa,8rawles2024androidworld,16fang2026dataset} &
Unsafe actions, permission violations, privacy leakage, adversarial UI robustness, non-determinism, reproducibility \\

\bottomrule
\end{tabularx}
\end{table}

Recent DLM acceleration studies further justify this layered framework. Consistency Diffusion Language Models (CDLM) improves sampling with consistency modeling, block-wise causal attention, multi-token finalization, and cache compatibility~\cite{17kim2025cdlm}, while R2-dLLM reduces redundant spatio-temporal decoding by finalizing confident and stable tokens~\cite{18du2026r}. These studies show that comprehensive DLM evaluation must jointly report accuracy, denoising trajectory, caching behavior, latency, and hardware efficiency. Overall, Table~\ref{tab:evaluation_framework} provides a concise basis for comparing autoregressive models, general DMs, and DLMs under the practical requirements of mobile edge agentic AI.

% section 7

\subsection{Lessons Learned}
% Section 7
DLM-based mobile edge agents should be evaluated beyond model accuracy, with generation dynamics, interaction reliability, system efficiency, and safety measured jointly~\cite{peng2025efficient,16fang2026dataset}. High benchmark scores do not imply deployability unless latency, memory, energy, communication, cache behavior, and tool execution are reported on real hardware~\cite{8rawles2024androidworld,9janapa2022mlperf}. Future benchmarks should standardize fixed-latency and fixed-energy settings, time-to-first-action, invalid-action rates, wireless dynamics, and reproducible hardware measurements.

\FloatBarrier

%
%
%
%
%%%%%%%%%%%%%%%%%%%%%%%%%%%%%%%%%%%%%%%%%%%%%%%%%%%%%%%%%%%%%%%%%%%%%%%%%%%%%%%%%%%%%%%%%%%%%%%%%%%%%%%%%%%%%%%%%%%%%%%%%%%%%%%%%%%%%%%%%%%%%%%
\section{Open Challenges and Future Directions}\label{sec8}

DLMs offer useful properties for mobile edge agentic AI, including non-autoregressive generation, bidirectional context modeling, parallel token refinement, and adaptive denoising~\cite{li2025survey,yu2025discrete}. However, most current studies remain model-centric and cloud-oriented, while edge systems require tight control of latency, memory, energy, bandwidth, privacy, and reliability~\cite{zheng2025review,peng2025efficient}. This section summarizes the main challenges and directions toward edge-native DLM systems.

\subsection{Edge-Native DLM Architectures}

A core challenge is that iterative denoising repeatedly accesses memory and may offset the parallelism gained from simultaneous token updates~\cite{peng2025efficient,wu2025fast}. This issue is more severe on mobile and IoT devices, where memory bandwidth, cache behavior, and accelerator support often dominate latency and energy. Future DLM architectures should jointly optimize denoising depth, token update ratio, mask schedule, cache compatibility, and hardware-friendly operators. Lightweight backbones, block-wise diffusion, sparse routing, early exit, and anytime generation are potential directions for adapting generation quality to real-time resource budgets~\cite{Efficient_Model3,zhu2026esdllm}.

\subsection{Long-Context Memory and Agent State Management}

Mobile edge agents must process multi-turn instructions, UI histories, sensor observations, tool calls, and previous actions. While autoregressive LLMs mainly suffer from KV-cache growth, DLMs face full-sequence refinement, repeated remasking, and multi-step token updates, which increase memory-bandwidth pressure on constrained devices~\cite{li2024survey,peng2025efficient}. Future systems should avoid refining all tokens at every step. Instead, they should identify stable tokens, selectively remask uncertain regions, compress historical context, and cache reusable agent states. Cache-aware and confidence-aware decoding also provides a useful starting point for efficient long-context DLM agents~\cite{wu2025fast,18du2026r}.

\subsection{Communication-Aware Split and Distributed DLM Inference}

Existing edge diffusion deployment primarily targets image or video generation, where latent features or denoising stages are split across devices and servers~\cite{du2023exploring,zheng2025diffusion}. DLMs are harder to distribute because they involve discrete tokens, reasoning traces, tool-call structures, and multimodal semantic states. Future distributed DLM systems should assign denoising steps, token blocks, or expert modules across devices, edge servers, and cloud nodes according to bandwidth, channel quality, latency, and privacy. A practical direction is to keep private prompt encoding and final action decoding on-device, while offloading heavy intermediate refinement or sparse experts to nearby edge nodes~\cite{qin2025optimal,jin2026moe,liu2024joint}.

\subsection{Privacy, Security, and Trustworthy Edge DLM Agents}

DLMs may support privacy-preserving split inference because forward noising and intermediate denoising states can obscure raw prompts before offloading~\cite{allmendinger2024collafuse,dockhorn2022differentially}. However, token distributions, latent states, reasoning traces, and tool-call intentions may still leak sensitive information. The risk is amplified when edge agents control real applications, devices, or physical environments. Future research should integrate privacy-preserving noising, secure split inference, trusted execution, and formal leakage analysis into distributed DLM pipelines. Iterative refinement also enables step-wise safety checking, constraint enforcement, and rollback before final action execution, which should be evaluated in realistic mobile-agent environments~\cite{7deng2024mobile,8rawles2024androidworld}.

\subsection{Multimodal and Physical-World DLM Agents}

Future mobile edge agents must handle camera inputs, speech, sensor streams, wireless signals, maps, UI states, and executable actions. Recent multimodal DLMs show that text reasoning, multimodal understanding, and text-to-image generation can be unified within diffusion architectures~\cite{yang2025mmada,tian2025mmada}. Yet, real-time deployment on mobile and IoT platforms remains underexplored. A potential direction is to represent instructions, observations, sensor readings, and actions as structured tokens refined through shared denoising. Such systems may support semantic communication, mobile assistants, vehicular coordination, UAV swarms, smart homes, and intent-driven network management, but require strict latency control, robustness, safety guarantees, closed-loop evaluation, and real-device validation~\cite{huh2026markov,habib2026generative}.

\subsection{Standardized Evaluation and Reproducibility}

Current benchmarks evaluate language ability, mobile interaction, or hardware efficiency separately, but DLM-based edge agents require joint evaluation of generation dynamics and system constraints~\cite{8rawles2024androidworld,9janapa2022mlperf,peng2025efficient}. Future benchmarks should report denoising steps, mask schedules, remasking stability, parallel decoding gain, cache behavior, energy use, communication overhead, and closed-loop task success. They should also include time-to-first-action, invalid action rate, privacy leakage risk, and safety failures. Reproducibility requires fixed-latency and fixed-energy protocols, together with hardware platform, accelerator type, quantization level, prompt length, output length, batch size, and power measurement details~\cite{16fang2026dataset,9janapa2022mlperf}.

\FloatBarrier

%
%
%
%
%%%%%%%%%%%%%%%%%%%%%%%%%%%%%%%%%%%%%%%%%%%%%%%%%%%%%%%%%%%%%%%%%%%%%%%%%%%%%%%%%%%%%%%%%%%%%%%%%%%%%%%%%%%%%%%%%%%%%%%%%%%%%%%%%%%%%%%%%%%%%%%%%%%
%Section 9
\section{Conclusion}\label{sec9}

DLMs offer new control dimensions for mobile edge agents, including parallel token refinement, adaptive denoising, constrained generation, and privacy-aware split execution. This survey has reviewed their modeling foundations, resource-efficient techniques, edge deployment architectures, communication-aware serving, IoT/wireless applications, and evaluation criteria. Our analysis suggests that DLMs should complement rather than replace autoregressive LLMs: they are suitable for structured, constraint-aware, and latency-adaptive tasks, but still face challenges in long-context state management, hardware-efficient inference, distributed collaboration, safety, multimodal grounding, and reproducibility. This survey aims to connect diffusion language modeling, edge computing, wireless networking, and trustworthy agentic AI.

\backmatter

\section*{Declarations}

\begin{itemize}
\item \textbf{Funding:} This work was supported by the National Natural Science Foundation of China under Grants 62571529 and U25A20388.
\item \textbf{Competing interests:} The authors have no competing interests to declare that are relevant to the content of this article.
\item \textbf{Author contributions:} Conceptualization, C.L., M.M., D.N. and W.N.; methodology, C.L., M.M. and D.N.; investigation, C.L.; data curation, C.L.; visualization, C.L.; writing---original draft preparation, C.L.; writing---review and editing, C.L., M.M., D.N. and W.N.; supervision, M.M., D.N. and W.N.; project administration, M.M.; funding acquisition, M.M. All authors have read and approved the final manuscript.
\item \textbf{Data availability:} No datasets were generated or analysed during the current study.
\item \textbf{Ethics approval and consent to participate:} Not applicable.
\item \textbf{Consent for publication:} Not applicable.
\item \textbf{Materials availability:} Not applicable.
\item \textbf{Code availability:} Not applicable.
\end{itemize}

\bibliography{References}% common bib file

\begin{thebibliography}{162}
\providecommand{\natexlab}[1]{#1}
\providecommand{\url}[1]{{#1}}
\providecommand{\urlprefix}{URL }
\providecommand{\doi}[1]{\url{https://doi.org/#1}}
\providecommand{\eprint}[2][]{\url{#2}}
 \bibcommenthead

\bibitem[{Ahsan et~al.(2025)Ahsan, Raman, Liu, and Siddique}]{ahsan2025comprehensive}
Ahsan MM, Raman S, Liu Y, et~al (2025) A comprehensive survey on diffusion models and their applications. Applied Soft Computing 181:113470. \doi{10.1016/j.asoc.2025.113470}

\bibitem[{Ai et~al.(2026)Ai, Chen, Wen, and Bennis}]{ai2026cross}
Ai Y, Chen Q, Wen D, et~al (2026) Cross-modal collaborative diffusion models for distributed {AI}-generated content. IEEE Transactions on Cognitive Communications and Networking 12:5552--5565. \doi{10.1109/TCCN.2026.3656389}

\bibitem[{Allmendinger et~al.(2026)Allmendinger, Zipperling, Struppek, and Kühl}]{allmendinger2024collafuse}
Allmendinger S, Zipperling D, Struppek L, et~al (2026) {CollaFuse}: Collaborative diffusion models. \urlprefix\url{https://arxiv.org/abs/2406.14429}, {\href{https://arxiv.org/abs/2406.14429}{{arXiv:2406.14429}}}

\bibitem[{Arriola et~al.(2025)Arriola, Gokaslan, Chiu, Yang, Qi, Han, Sahoo, and Kuleshov}]{Efficient_Model3}
Arriola M, Gokaslan A, Chiu J, et~al (2025) Block diffusion: Interpolating between autoregressive and diffusion language models. In: International Conference on Learning Representations, pp 50726--50753

\bibitem[{Asaria et~al.(2026)Asaria, Salomone, and Gandhi}]{asaria2026diffusiongemma}
Asaria A, Salomone T, Gandhi D (2026) Neither parallel nor sequential: How {DiffusionGemma} actually commits tokens. \urlprefix\url{https://arxiv.org/abs/2606.14620}, {\href{https://arxiv.org/abs/2606.14620}{{arXiv:2606.14620}}}

\bibitem[{Austin et~al.(2021)Austin, Johnson, Ho, Tarlow, and van~den Berg}]{Diffusion_Models2}
Austin J, Johnson DD, Ho J, et~al (2021) Structured denoising diffusion models in discrete state-spaces. In: Advances in Neural Information Processing Systems, pp 17981--17993

\bibitem[{Bao et~al.(2023)Bao, Nie, Xue, Cao, Li, Su, and Zhu}]{Efficient_Model1}
Bao F, Nie S, Xue K, et~al (2023) All are worth words: A {ViT} backbone for diffusion models. In: Proceedings of the IEEE/CVF Conference on Computer Vision and Pattern Recognition (CVPR), pp 22669--22679

\bibitem[{Bao et~al.(2025)Bao, Chen, Xu, and Shang}]{bao2025learning}
Bao W, Chen Z, Xu D, et~al (2025) Learning to parallel: Accelerating diffusion large language models via learnable parallel decoding. \urlprefix\url{https://arxiv.org/abs/2509.25188}, {\href{https://arxiv.org/abs/2509.25188}{{arXiv:2509.25188}}}

\bibitem[{Bolya and Hoffman(2023)}]{bolya2023token}
Bolya D, Hoffman J (2023) Token merging for fast {Stable Diffusion}. In: Proceedings of the IEEE/CVF Conference on Computer Vision and Pattern Recognition (CVPR) Workshops, pp 4599--4603

\bibitem[{Brown et~al.(2020)Brown, Mann, Ryder, Subbiah, Kaplan, Dhariwal, Neelakantan, Shyam, Sastry, Askell, Agarwal, Herbert-Voss, Krueger, Henighan, Child, Ramesh, Ziegler, Wu, Winter, Hesse, Chen, Sigler, Litwin, Gray, Chess, Clark, Berner, McCandlish, Radford, Sutskever, and Amodei}]{brown2020language}
Brown T, Mann B, Ryder N, et~al (2020) Language models are few-shot learners. In: Advances in Neural Information Processing Systems, vol~33. Curran Associates, Inc., pp 1877--1901

\bibitem[{Chang et~al.(2022)Chang, Zhang, Jiang, Liu, and Freeman}]{chang2022maskgit}
Chang H, Zhang H, Jiang L, et~al (2022) {MaskGIT}: Masked generative image transformer. In: Proceedings of the IEEE/CVF conference on computer vision and pattern recognition, pp 11315--11325

\bibitem[{Chen et~al.(2026)Chen, Song, Ding, Zhou, Zhao, Tang, Wang, and Li}]{chen2026ud_vla}
Chen J, Song W, Ding P, et~al (2026) Unified diffusion {VLA}: Vision-language-action model via joint discrete denosing diffusion process. In: The Fourteenth International Conference on Learning Representations

\bibitem[{Chen et~al.(2021)Chen, Tworek, Jun, Yuan, de~Oliveira~Pinto, Kaplan, Edwards, Burda, Joseph, Brockman et~al.}]{4chen2021evaluating}
Chen M, Tworek J, Jun H, et~al (2021) Evaluating large language models trained on code. \urlprefix\url{https://arxiv.org/abs/2107.03374}, {\href{https://arxiv.org/abs/2107.03374}{{arXiv:2107.03374}}}

\bibitem[{Croitoru et~al.(2023)Croitoru, Hondru, Ionescu, and Shah}]{croitoru2023diffusion}
Croitoru FA, Hondru V, Ionescu RT, et~al (2023) Diffusion models in vision: A survey. IEEE Transactions on Pattern Analysis and Machine Intelligence 45(9):10850--10869. \doi{10.1109/TPAMI.2023.3261988}

\bibitem[{Deng et~al.(2020)Deng, Zhao, Fang, Yin, Dustdar, and Zomaya}]{9052677}
Deng S, Zhao H, Fang W, et~al (2020) Edge intelligence: The confluence of edge computing and artificial intelligence. IEEE Internet of Things Journal 7(8):7457--7469. \doi{10.1109/JIOT.2020.2984887}

\bibitem[{Deng et~al.(2024)Deng, Xu, Sun, Liu, Tan, Liu, Li, Luan, Wang, Yan, and Shang}]{7deng2024mobile}
Deng S, Xu W, Sun H, et~al (2024) {Mobile-Bench}: An evaluation benchmark for {LLM}-based mobile agents. In: Proceedings of the 62nd Annual Meeting of the Association for Computational Linguistics (Volume 1: Long Papers), pp 8813--8831, \doi{10.18653/v1/2024.acl-long.478}

\bibitem[{Devlin et~al.(2019)Devlin, Chang, Lee, Toutanova, Doran, and Solorio}]{devlin2019bert}
Devlin J, Chang MW, Lee K, et~al (2019) {BERT}: Pre-training of deep bidirectional transformers for language understanding. In: Proceedings of the 2019 Conference of the North {A}merican Chapter of the Association for Computational Linguistics: Human Language Technologies, Volume 1 (Long and Short Papers), pp 4171--4186, \doi{10.18653/v1/N19-1423}

\bibitem[{Dockhorn et~al.(2023)Dockhorn, Cao, Vahdat, and Kreis}]{dockhorn2022differentially}
Dockhorn T, Cao T, Vahdat A, et~al (2023) Differentially private diffusion models. Transactions on Machine Learning Research

\bibitem[{Du et~al.(2024)Du, Zhang, Niyato, Kang, Xiong, Kim, Shen, and Poor}]{du2023exploring}
Du H, Zhang R, Niyato D, et~al (2024) Exploring collaborative distributed diffusion-based {AI}-generated content ({AIGC}) in wireless networks. IEEE Network 38(3):178--186. \doi{10.1109/MNET.006.2300223}

\bibitem[{Du et~al.(2026)Du, Xia, Zhong, Fu, Oswald, Ji, Khailany, Molchanov, and Lin}]{18du2026r}
Du Z, Xia K, Zhong X, et~al (2026) {R2-dLLM}: Accelerating diffusion large language models via spatio-temporal redundancy reduction. \urlprefix\url{https://arxiv.org/abs/2604.18995}, {\href{https://arxiv.org/abs/2604.18995}{{arXiv:2604.18995}}}

\bibitem[{Fan et~al.(2026)Fan, Meng, Xu, Liu, Nan, Feng, Han, Gao, Xu, Niyato, Quek, and Zhang}]{fan2026generative}
Fan D, Meng R, Xu X, et~al (2026) Generative diffusion models for wireless networks: Fundamental, architecture, and state-of-the-art. IEEE Communications Surveys \& Tutorials 28:5632--5677. \doi{10.1109/COMST.2026.3671110}

\bibitem[{Fang et~al.(2023)Fang, Ma, and Wang}]{fang2023structural}
Fang G, Ma X, Wang X (2023) Structural pruning for diffusion models. In: Thirty-seventh Conference on Neural Information Processing Systems

\bibitem[{Fang et~al.(2025)Fang, Pan, Li, Sun, and Jiannan}]{fang2024pipefusion}
Fang J, Pan J, Li A, et~al (2025) {PipeFusion}: Patch-level pipeline parallelism for diffusion transformers inference. In: Advances in Neural Information Processing Systems, pp 98434--98455

\bibitem[{Fang et~al.(2026)Fang, Jiang, Chen, Zhang, Li, Tang, Li, and Li}]{16fang2026dataset}
Fang Z, Jiang Z, Chen H, et~al (2026) Dataset-level metrics attenuate non-determinism: A fine-grained non-determinism evaluation in diffusion language models. \urlprefix\url{https://arxiv.org/abs/2604.13413}, {\href{https://arxiv.org/abs/2604.13413}{{arXiv:2604.13413}}}

\bibitem[{Fei et~al.(2024)Fei, Fan, Yu, Li, and Huang}]{Efficient_Model6}
Fei Z, Fan M, Yu C, et~al (2024) Scaling diffusion transformers to 16 billion parameters. \urlprefix\url{https://arxiv.org/abs/2407.11633}, {\href{https://arxiv.org/abs/2407.11633}{{arXiv:2407.11633}}}

\bibitem[{Feng et~al.(2025{\natexlab{a}})Feng, Geng, Guan, Wu, Wang, and He}]{15feng2025theoretical}
Feng G, Geng Y, Guan J, et~al (2025{\natexlab{a}}) Theoretical benefit and limitation of diffusion language model. In: Advances in Neural Information Processing Systems, pp 24415--24459

\bibitem[{Feng et~al.(2025{\natexlab{b}})Feng, Zhang, Zhu, Wang, Sun, Zhu, Li, and Taleb}]{feng2024exploring}
Feng W, Zhang R, Zhu Y, et~al (2025{\natexlab{b}}) Exploring collaborative diffusion model inferring for {AIGC}-enabled edge services. IEEE Transactions on Cognitive Communications and Networking 11(2):946--960. \doi{10.1109/TCCN.2024.3519320}

\bibitem[{Fu et~al.(2025)Fu, Huang, Adams, Wang, Srinivasan, and Jiao}]{fu2025bits}
Fu H, Huang B, Adams V, et~al (2025) From bits to rounds: Parallel decoding with exploration for diffusion language models. \urlprefix\url{https://arxiv.org/abs/2511.21103}, {\href{https://arxiv.org/abs/2511.21103}{{arXiv:2511.21103}}}

\bibitem[{Gao et~al.(2026)Gao, Zhang, Jing, Zhang, Zhou, Wang, and Cai}]{gao2025towards}
Gao S, Zhang S, Jing S, et~al (2026) Towards efficient federated learning of networked mixture-of-experts for mobile edge computing. \urlprefix\url{https://arxiv.org/abs/2511.01743}, {\href{https://arxiv.org/abs/2511.01743}{{arXiv:2511.01743}}}

\bibitem[{Ghiglino et~al.(2026)Ghiglino, Elenius, Roy, Kaur, Acharya, Samplawski, Matejek, Jha, Alonso, and Cobb}]{ghiglino2026diffusion}
Ghiglino A, Elenius D, Roy A, et~al (2026) Do diffusion models dream of electric planes? {D}iscrete and continuous simulation-based inference for aircraft design. \urlprefix\url{https://arxiv.org/abs/2603.13284}, {\href{https://arxiv.org/abs/2603.13284}{{arXiv:2603.13284}}}

\bibitem[{Gong et~al.(2023)Gong, Li, Feng, Wu, and Kong}]{gong2022diffuseq}
Gong S, Li M, Feng J, et~al (2023) {DiffuSeq}: Sequence to sequence text generation with diffusion models. \urlprefix\url{https://arxiv.org/abs/2210.08933}, {\href{https://arxiv.org/abs/2210.08933}{{arXiv:2210.08933}}}

\bibitem[{Gong et~al.(2025)Gong, Agarwal, Zhang, Ye, Zheng, Li, An, Zhao, Bi, Han, Peng, and Kong}]{13gong2024scaling}
Gong S, Agarwal S, Zhang Y, et~al (2025) Scaling diffusion language models via adaptation from autoregressive models. In: International Conference on Learning Representations, pp 5046--5073

\bibitem[{{Google}(2026{\natexlab{a}})}]{google2026diffusiongemma_model}
{Google} (2026{\natexlab{a}}) {DiffusionGemma 26B-A4B-IT Model Card}. \url{https://huggingface.co/google/diffusiongemma-26B-A4B-it}, accessed: 2026-06-26

\bibitem[{{Google}(2026{\natexlab{b}})}]{google2026diffusiongemma_blog}
{Google} (2026{\natexlab{b}}) {DiffusionGemma}: 4x faster text generation. \url{https://blog.google/innovation-and-ai/technology/developers-tools/diffusion-gemma-faster-text-generation/}, accessed: 2026-06-26

\bibitem[{Grootendorst(2022)}]{masked_language_models1}
Grootendorst M (2022) {BERTopic}: Neural topic modeling with a class-based {TF-IDF} procedure. \urlprefix\url{https://arxiv.org/abs/2203.05794}, {\href{https://arxiv.org/abs/2203.05794}{{arXiv:2203.05794}}}

\bibitem[{Gruver et~al.(2023)Gruver, Finzi, Qiu, and Wilson}]{autoregressive_models2}
Gruver N, Finzi M, Qiu S, et~al (2023) Large language models are zero-shot time series forecasters. In: Advances in Neural Information Processing Systems, pp 19622--19635

\bibitem[{Habib et~al.(2026)Habib, Elsayed, Ozcan, Iturria-Rivera, Bavand, and Erol-Kantarci}]{habib2026generative}
Habib MA, Elsayed M, Ozcan Y, et~al (2026) Generative {AI} for intent-driven network management in {6G} {RAN}: A case study on the {Mamba} model. IEEE Wireless Communications pp 1--8. \doi{10.1109/MWC.2026.3667666}

\bibitem[{Han et~al.(2023)Han, Kumar, Tsvetkov, Boyd-Graber, and Okazaki}]{han2023ssd}
Han X, Kumar S, Tsvetkov Y, et~al (2023) {SSD}-{LM}: Semi-autoregressive simplex-based diffusion language model for text generation and modular control. In: Proceedings of the 61st Annual Meeting of the Association for Computational Linguistics (Volume 1: Long Papers). Association for Computational Linguistics, pp 11575--11596, \doi{10.18653/v1/2023.acl-long.647}

\bibitem[{He et~al.(2025)He, Chen, Wang, and Wang}]{he2025communication}
He J, Chen K, Wang B, et~al (2025) Communication-aware diffusion models for multi-agent trajectory forecasting in connected and autonomous vehicles. In: Proceedings of the Tenth ACM/IEEE Symposium on Edge Computing, \doi{10.1145/3769102.3774636}

\bibitem[{He et~al.(2023{\natexlab{a}})He, Gao, and Chen}]{masked_language_models2}
He P, Gao J, Chen W (2023{\natexlab{a}}) {DeBERTaV3}: Improving {DeBERTa} using {ELECTRA}-style pre-training with gradient-disentangled embedding sharing. In: The Eleventh International Conference on Learning Representations

\bibitem[{He et~al.(2023{\natexlab{b}})He, Liu, Liu, Wu, Zhou, and Zhuang}]{he2023ptqd}
He Y, Liu L, Liu J, et~al (2023{\natexlab{b}}) {PTQD}: Accurate post-training quantization for diffusion models. Advances in Neural Information Processing Systems 36:13237--13249

\bibitem[{Hendrycks et~al.(2021)Hendrycks, Burns, Basart, Zou, Mazeika, Song, and Steinhardt}]{1hendrycks2020measuring}
Hendrycks D, Burns C, Basart S, et~al (2021) Measuring massive multitask language understanding. In: International Conference on Learning Representations

\bibitem[{Ho et~al.(2020)Ho, Jain, and Abbeel}]{10ho2020denoising}
Ho J, Jain A, Abbeel P (2020) Denoising diffusion probabilistic models. In: Advances in Neural Information Processing Systems, pp 6840--6851

\bibitem[{Hoefler et~al.(2024)Hoefler, Mazouka, Mueller, and Samek}]{Efficient_Training6}
Hoefler MA, Mazouka T, Mueller K, et~al (2024) Boosting federated learning with diffusion models for non-{IID} and imbalanced data. In: 2024 IEEE International Conference on Big Data (BigData), IEEE, pp 7790--7799

\bibitem[{Hu et~al.(2026)Hu, Gupta, Gabidolla, Sahni, Coskun, Li, Idelbayev, Mahmood, Lebedev, Lahiri, Goyal, Hu, Gong, Tulyakov, and Kag}]{Efficient_Model2}
Hu D, Gupta A, Gabidolla M, et~al (2026) {SnapGen++}: Unleashing diffusion transformers for efficient high-fidelity image generation on edge devices. \urlprefix\url{https://arxiv.org/abs/2601.08303}, {\href{https://arxiv.org/abs/2601.08303}{{arXiv:2601.08303}}}

\bibitem[{Hu et~al.(2022)Hu, Shen, Wallis, Allen-Zhu, Li, Wang, Wang, Chen et~al.}]{Efficient_Training4}
Hu EJ, Shen Y, Wallis P, et~al (2022) {LoRA}: Low-rank adaptation of large language models. In: International Conference on Learning Representations, \urlprefix\url{https://openreview.net/forum?id=nZeVKeeFYf9}

\bibitem[{Hu et~al.(2025)Hu, Ye, Kang, Wu, and Yu}]{hu2024cloud}
Hu Y, Ye D, Kang J, et~al (2025) A cloud--edge collaborative architecture for multimodal {LLM}-based advanced driver assistance systems in {IoT} networks. IEEE Internet of Things Journal 12(10):13208--13221. \doi{10.1109/JIOT.2024.3509628}

\bibitem[{Huang et~al.(2026)Huang, Zou, Zhong, Kang, and Xie}]{huang2025diffusion}
Huang X, Zou R, Zhong W, et~al (2026) Diffusion-based deep reinforcement learning for service scheduling in serverless vehicular edge computing. IEEE Internet of Things Journal 13(5):8424--8435. \doi{10.1109/JIOT.2025.3641386}

\bibitem[{Huh et~al.(2026)Huh, Kang, and Choi}]{huh2026markov}
Huh Y, Kang J, Choi W (2026) Markov-enforced discrete diffusion model for digital semantic symbol error correction. \urlprefix\url{https://arxiv.org/abs/2603.22983}, {\href{https://arxiv.org/abs/2603.22983}{{arXiv:2603.22983}}}

\bibitem[{Huo et~al.(2026)Huo, Zhou, Hao, Hu, Mo, Chen, and Humar}]{huo2025multi}
Huo D, Zhou Y, Hao Y, et~al (2026) Multi-modal model partition strategy for end-edge collaborative inference. Journal of Parallel and Distributed Computing 208:105189. \doi{10.1016/j.jpdc.2025.105189}

\bibitem[{Janapa~Reddi et~al.(2022)Janapa~Reddi, Kanter, Mattson, Duke, Nguyen, Chukka, Shiring, Tan, Charlebois, Chou, El-Khamy, Hong, St~John, Trinh, Buch, Mazumder, Markovic, Atta, Cakir, Charkhabi, Chen, Chiang, Dexter, Heo, Schmuelling, Shabani, and Zika}]{9janapa2022mlperf}
Janapa~Reddi V, Kanter D, Mattson P, et~al (2022) {MLPerf} mobile inference benchmark: An industry-standard open-source machine learning benchmark for on-device {AI}. In: Proceedings of Machine Learning and Systems, pp 352--369

\bibitem[{Jiang et~al.(2023)Jiang, Sablayrolles, Mensch, Bamford, Chaplot, de~las Casas, Bressand, Lengyel, Lample, Saulnier, Lavaud, Lachaux, Stock, Scao, Lavril, Wang, Lacroix, and Sayed}]{jiang2023mistral7b}
Jiang AQ, Sablayrolles A, Mensch A, et~al (2023) Mistral {7B}. \urlprefix\url{https://arxiv.org/abs/2310.06825}, {\href{https://arxiv.org/abs/2310.06825}{{arXiv:2310.06825}}}

\bibitem[{Jiang et~al.(2024)Jiang, Sablayrolles, Roux, Mensch, Savary, Bamford, Chaplot, de~las Casas, Hanna, Bressand, Lengyel, Bour, Lample, Lavaud, Saulnier, Lachaux, Stock, Subramanian, Yang, Antoniak, Scao, Gervet, Lavril, Wang, Lacroix, and Sayed}]{jiang2024mixtralexperts}
Jiang AQ, Sablayrolles A, Roux A, et~al (2024) Mixtral of experts. \urlprefix\url{https://arxiv.org/abs/2401.04088}, {\href{https://arxiv.org/abs/2401.04088}{{arXiv:2401.04088}}}

\bibitem[{Jin et~al.(2026)Jin, Zhang, Li, Wang, Yang, Wu, and Zhang}]{jin2026moe}
Jin L, Zhang Y, Li Y, et~al (2026) {MoE}$^2$: Optimizing collaborative inference for edge large language models. IEEE Transactions on Networking 34:4637--4651. \doi{10.1109/TON.2026.3677371}

\bibitem[{Jin et~al.(2024)Jin, Wang, Ma, Chu, Zhang, Shi, Chen, Liang, Li, Pan, and Wen}]{autoregressive_models4}
Jin M, Wang S, Ma L, et~al (2024) {Time-LLM}: Time series forecasting by reprogramming large language models. In: Kim B, Yue Y, Chaudhuri S, et~al (eds) International Conference on Learning Representations, pp 23857--23880

\bibitem[{Jung et~al.(2026)Jung, Kim, Kim, Cho, and Lee}]{jung2026accelerating}
Jung E, Kim B, Kim H, et~al (2026) Accelerating diffusion via hybrid data-pipeline parallelism based on conditional guidance scheduling. In: Proceedings of the IEEE/CVF Conference on Computer Vision and Pattern Recognition (CVPR), pp 9374--9383

\bibitem[{Karras et~al.(2022)Karras, Aittala, Aila, and Laine}]{Diffusion_Models3}
Karras T, Aittala M, Aila T, et~al (2022) Elucidating the design space of diffusion-based generative models. In: Advances in Neural Information Processing Systems, pp 26565--26577

\bibitem[{Karunanayake et~al.(2025)Karunanayake, Khalil, Yi, and Lam}]{karunanayake2025toward}
Karunanayake B, Khalil I, Yi X, et~al (2025) Toward {LLM}-driven adaptive policy orchestration for host-based intrusion detection systems in {IoT} environments. IEEE Network 39(5):66--73. \doi{10.1109/MNET.2025.3579532}

\bibitem[{Kim et~al.(2025)Kim, Lee, Jeong, Cheon, Lee, and Lee}]{kim2025device}
Kim B, Lee K, Jeong I, et~al (2025) On-device {Sora}: Enabling training-free diffusion-based text-to-video generation for mobile devices. \urlprefix\url{https://arxiv.org/abs/2502.04363}, {\href{https://arxiv.org/abs/2502.04363}{{arXiv:2502.04363}}}

\bibitem[{Kim et~al.(2026)Kim, Xu, Hooper, Singh, Athiwaratkun, Zhang, Keutzer, and Gholami}]{17kim2025cdlm}
Kim M, Xu C, Hooper C, et~al (2026) {CDLM}: Consistency diffusion language models for faster sampling. \urlprefix\url{https://arxiv.org/abs/2511.19269}, {\href{https://arxiv.org/abs/2511.19269}{{arXiv:2511.19269}}}

\bibitem[{Kodavanti et~al.(2026)Kodavanti, Arveti, Vajrala, Miriyala, and R}]{kodavanti2026edgedit}
Kodavanti S, Arveti M, Vajrala S, et~al (2026) {EdgeDiT}: Hardware-aware diffusion transformers for efficient on-device image generation. In: Proceedings of the IEEE/CVF Conference on Computer Vision and Pattern Recognition (CVPR) Workshops, pp 3801--3809

\bibitem[{Kong et~al.(2026)Kong, Yang, Qin, Zhou, and Shen}]{kong2025distributed}
Kong Y, Yang P, Qin X, et~al (2026) Distributed and controllable mobile text-to-image generation with user preference guarantee. IEEE Transactions on Mobile Computing 25(3):3712--3727. \doi{10.1109/TMC.2025.3620352}

\bibitem[{Lai et~al.(2024)Lai, He, Kang, Li, Xu, Zhang, and Xie}]{Efficient_Training8}
Lai B, He J, Kang J, et~al (2024) On-demand quantization for green federated generative diffusion in mobile edge networks. In: ICC 2024 - IEEE International Conference on Communications, pp 2883--2888, \doi{10.1109/ICC51166.2024.10622695}

\bibitem[{Leviathan et~al.(2023)Leviathan, Kalman, and Matias}]{leviathan2023fast}
Leviathan Y, Kalman M, Matias Y (2023) Fast inference from transformers via speculative decoding. In: Proceedings of the 40th International Conference on Machine Learning, vol 202. PMLR, pp 19274--19286

\bibitem[{Li(2026)}]{li2026fastusp}
Li G (2026) {FastUSP}: A multi-level collaborative acceleration framework for distributed diffusion model inference. \urlprefix\url{https://arxiv.org/abs/2602.10940}, {\href{https://arxiv.org/abs/2602.10940}{{arXiv:2602.10940}}}

\bibitem[{Li et~al.(2025{\natexlab{a}})Li, Li, Tian, Tang, Xu, Chen, Hu, Dong, Li, and Chen}]{li2024survey}
Li H, Li Y, Tian A, et~al (2025{\natexlab{a}}) A survey on large language model acceleration based on {KV} cache management. \urlprefix\url{https://arxiv.org/abs/2412.19442}, {\href{https://arxiv.org/abs/2412.19442}{{arXiv:2412.19442}}}

\bibitem[{Li et~al.(2024)Li, Cai, Cao, Zhang, Cai, Bai, Jia, Li, and Han}]{li2024distrifusion}
Li M, Cai T, Cao J, et~al (2024) {DistriFusion}: Distributed parallel inference for high-resolution diffusion models. In: 2024 IEEE/CVF Conference on Computer Vision and Pattern Recognition (CVPR), pp 7183--7193, \doi{10.1109/CVPR52733.2024.00686}

\bibitem[{Li et~al.(2025{\natexlab{b}})Li, Yang, Siew, Xiong, Chen, Mao, and Lam}]{li2025edge}
Li N, Yang W, Siew M, et~al (2025{\natexlab{b}}) Edge-assisted collaborative fine-tuning for multi-user personalized artificial intelligence generated content (aigc). \urlprefix\url{https://arxiv.org/abs/2508.04745}, {\href{https://arxiv.org/abs/2508.04745}{{arXiv:2508.04745}}}

\bibitem[{Li et~al.(2025{\natexlab{c}})Li, Dong, Qian, Zhou, and Wu}]{li2024flexgen}
Li P, Dong H, Qian L, et~al (2025{\natexlab{c}}) {FlexGen}: Efficient on-demand generative {AI} service with flexible diffusion model in mobile edge networks. IEEE Transactions on Cognitive Communications and Networking 11(2):961--973. \doi{10.1109/TCCN.2024.3522084}

\bibitem[{Li et~al.(2025{\natexlab{d}})Li, Zheng, Wang, Wang, Zhao, Liu, Zhan, Zhan, and Lang}]{li2025reflectdrive}
Li P, Zheng Y, Wang Y, et~al (2025{\natexlab{d}}) Discrete diffusion for reflective vision-language-action models in autonomous driving. \urlprefix\url{https://arxiv.org/abs/2509.20109}, {\href{https://arxiv.org/abs/2509.20109}{{arXiv:2509.20109}}}

\bibitem[{Li et~al.(2026)Li, Qian, Niyato, Mao, and Wu}]{li2026toward}
Li P, Qian L, Niyato D, et~al (2026) Toward resource-efficient collaboration of large {AI} models in mobile edge networks. IEEE Network 40(3):51--59. \doi{10.1109/MNET.2025.3650049}

\bibitem[{Li et~al.(2025{\natexlab{e}})Li, Kallidromitis, Bansal, Gokul, Kato, Kozuka, Kuen, Lin, Chang, and Grover}]{NEURIPS2025_975affbe}
Li S, Kallidromitis K, Bansal H, et~al (2025{\natexlab{e}}) {LaViDa}: A large diffusion language model for multimodal understanding. In: Advances in Neural Information Processing Systems, pp 105101--105134, \urlprefix\url{https://proceedings.neurips.cc/paper_files/paper/2025/file/975affbe7b5f3b55fba62247b6877b1c-Paper-Conference.pdf}

\bibitem[{Li et~al.(2025{\natexlab{f}})Li, Chen, Guo, and Shen}]{li2025survey}
Li T, Chen M, Guo B, et~al (2025{\natexlab{f}}) A survey on diffusion language models. \urlprefix\url{https://arxiv.org/abs/2508.10875}, {\href{https://arxiv.org/abs/2508.10875}{{arXiv:2508.10875}}}

\bibitem[{Li et~al.(2022)Li, Thickstun, Gulrajani, Liang, and Hashimoto}]{li2022diffusion}
Li X, Thickstun J, Gulrajani I, et~al (2022) {Diffusion-LM} improves controllable text generation. In: Advances in Neural Information Processing Systems, pp 4328--4343

\bibitem[{Li et~al.(2023{\natexlab{a}})Li, Liu, Lian, Yang, Dong, Kang, Zhang, and Keutzer}]{li2023q}
Li X, Liu Y, Lian L, et~al (2023{\natexlab{a}}) {Q-Diffusion}: Quantizing diffusion models. In: Proceedings of the IEEE/CVF International Conference on Computer Vision, pp 17535--17545

\bibitem[{Li et~al.(2023{\natexlab{b}})Li, Wang, Jin, Hu, Chemerys, Fu, Wang, Tulyakov, and Ren}]{li2023snapfusion}
Li Y, Wang H, Jin Q, et~al (2023{\natexlab{b}}) {SnapFusion}: Text-to-image diffusion model on mobile devices within two seconds. In: Advances in Neural Information Processing Systems, pp 20662--20678

\bibitem[{Liang et~al.(2023)Liang, Bommasani, Lee, Tsipras, Soylu, Yasunaga, Zhang, Narayanan, Wu, Kumar et~al.}]{2liang2022holistic}
Liang P, Bommasani R, Lee T, et~al (2023) Holistic evaluation of language models. Transactions on Machine Learning Research

\bibitem[{Liang et~al.(2025{\natexlab{a}})Liang, Yang, Chen, Cao, Yu, Debbah, Niyato, Poor, and Yuen}]{liang2025gdsg}
Liang R, Yang B, Chen P, et~al (2025{\natexlab{a}}) {GDSG}: Graph diffusion-based solution generator for optimization problems in {MEC} networks. IEEE Transactions on Mobile Computing 24(10):10264--10277. \doi{10.1109/TMC.2025.3568248}

\bibitem[{Liang et~al.(2025{\natexlab{b}})Liang, Yang, Chen, Li, Xue, Yu, Cao, Zhang, Debbah, Poor, and Yuen}]{liang2025diffusion}
Liang R, Yang B, Chen P, et~al (2025{\natexlab{b}}) Diffusion models as network optimizers: Explorations and analysis. IEEE Internet of Things Journal 12(10):13183--13193. \doi{10.1109/JIOT.2025.3528955}

\bibitem[{Liang et~al.(2025{\natexlab{c}})Liang, Li, Yang, Wu, Mao, Nian, Pei, Zhou, Yang, Pang, Mu, and Luo}]{liang2025discrete_vla}
Liang Z, Li Y, Yang T, et~al (2025{\natexlab{c}}) Discrete diffusion {VLA}: Bringing discrete diffusion to action decoding in vision-language-action policies. \urlprefix\url{https://arxiv.org/abs/2508.20072}, {\href{https://arxiv.org/abs/2508.20072}{{arXiv:2508.20072}}}

\bibitem[{Lin et~al.(2022)Lin, Hilton, and Evans}]{3lin2022truthfulqa}
Lin S, Hilton J, Evans O (2022) {T}ruthful{QA}: Measuring how models mimic human falsehoods. In: Proceedings of the 60th Annual Meeting of the Association for Computational Linguistics (Volume 1: Long Papers), pp 3214--3252, \doi{10.18653/v1/2022.acl-long.229}

\bibitem[{Lipman et~al.(2023)Lipman, Chen, Ben-Hamu, Nickel, and Le}]{Diffusion_Models1}
Lipman Y, Chen R, Ben-Hamu H, et~al (2023) Flow matching for generative modeling. In: International Conference on Learning Representations

\bibitem[{Liu et~al.(2024{\natexlab{a}})Liu, Wu, Zhuang, Wu, and Gao}]{liu2024joint}
Liu H, Wu J, Zhuang X, et~al (2024{\natexlab{a}}) Joint communication and computation scheduling for {MEC}-enabled {AIGC} services based on generative diffusion model. In: 2024 22nd International Symposium on Modeling and Optimization in Mobile, Ad Hoc, and Wireless Networks (WiOpt), pp 345--352

\bibitem[{Liu et~al.(2024{\natexlab{b}})Liu, Yu, Zhang, Xu, Lei, Lai, Gu, Ding, Men, Yang, Zhang, Deng, Zeng, Du, Zhang, Shen, Zhang, Su, Sun, Huang, Dong, and Tang}]{5liu2023agentbench}
Liu X, Yu H, Zhang H, et~al (2024{\natexlab{b}}) {AgentBench}: Evaluating {LLMs} as agents. In: Kim B, Yue Y, Chaudhuri S, et~al (eds) International Conference on Learning Representations, pp 52989--53046

\bibitem[{Liu et~al.(2019)Liu, Ott, Goyal, Du, Joshi, Chen, Levy, Lewis, Zettlemoyer, and Stoyanov}]{liu2019roberta}
Liu Y, Ott M, Goyal N, et~al (2019) {RoBERTa}: A robustly optimized {BERT} pretraining approach. \urlprefix\url{https://arxiv.org/abs/1907.11692}, {\href{https://arxiv.org/abs/1907.11692}{{arXiv:1907.11692}}}

\bibitem[{Liu et~al.(2026{\natexlab{a}})Liu, Ding, Jiang, Wang, Song, Lin, Zhao, Zhang, Zhuang, Zhao, Huang, Shi, and Wang}]{liu2026mmada_vla}
Liu Y, Ding P, Jiang T, et~al (2026{\natexlab{a}}) {MMaDA-VLA}: Large diffusion vision-language-action model with unified multi-modal instruction and generation. \urlprefix\url{https://arxiv.org/abs/2603.25406}, {\href{https://arxiv.org/abs/2603.25406}{{arXiv:2603.25406}}}

\bibitem[{Liu et~al.(2026{\natexlab{b}})Liu, Xiao, Wang, Lv, Zhan, Xiao, and Yang}]{liu2025reinforcement}
Liu Y, Xiao L, Wang C, et~al (2026{\natexlab{b}}) Reinforcement learning-based edge-assisted inference with multimodal data. IEEE Transactions on Mobile Computing 25(4):5016--5031. \doi{10.1109/TMC.2025.3627276}

\bibitem[{Liu et~al.(2026{\natexlab{c}})Liu, Du, Hou, Huang, Hosseinalipour, Niyato, and Letaief}]{liu2025two}
Liu Z, Du H, Hou X, et~al (2026{\natexlab{c}}) Two-timescale model caching and resource allocation for edge-enabled {AI}-generated content services. IEEE Transactions on Mobile Computing 25(4):4822--4838. \doi{10.1109/TMC.2025.3627263}

\bibitem[{Lou et~al.(2024)Lou, Meng, and Ermon}]{Language_diffusion5}
Lou A, Meng C, Ermon S (2024) Discrete diffusion modeling by estimating the ratios of the data distribution. In: International Conference on Machine Learning. JMLR.org, ICML'24

\bibitem[{Lu et~al.(2022)Lu, Zhou, Bao, Chen, Li, and Zhu}]{lu2022dpmsolver}
Lu C, Zhou Y, Bao F, et~al (2022) {DPM-Solver}: A fast {ODE} solver for diffusion probabilistic model sampling in around 10 steps. In: Advances in Neural Information Processing Systems, pp 5775--5787

\bibitem[{Luo et~al.(2023)Luo, Tan, Huang, Li, and Zhao}]{luo2023latent}
Luo S, Tan Y, Huang L, et~al (2023) Latent consistency models: Synthesizing high-resolution images with few-step inference. \urlprefix\url{https://arxiv.org/abs/2310.04378}, {\href{https://arxiv.org/abs/2310.04378}{{arXiv:2310.04378}}}

\bibitem[{Luong et~al.(2025)Luong, Hai, Le, Nguyen, Vu, Huynh-The, Zhang, Anh, Niyato, Renzo, Kim, and Pham}]{luong2025diffusion}
Luong NC, Hai ND, Le DV, et~al (2025) Diffusion models for future networks and communications: A comprehensive survey. \urlprefix\url{https://arxiv.org/abs/2508.01586}, {\href{https://arxiv.org/abs/2508.01586}{{arXiv:2508.01586}}}

\bibitem[{Ma et~al.(2025)Ma, Zhang, Jia, Zhao, Ma, Ma, Liu, Zhang, Ding, Li, and Zhou}]{ma2025efficient}
Ma Z, Zhang Y, Jia G, et~al (2025) Efficient diffusion models: A comprehensive survey from principles to practices. IEEE Transactions on Pattern Analysis and Machine Intelligence 47(9):7506--7525. \doi{10.1109/TPAMI.2025.3569700}

\bibitem[{Mendieta et~al.(2025)Mendieta, Sun, and Chen}]{Efficient_Training7}
Mendieta M, Sun G, Chen C (2025) Navigating heterogeneity and privacy in one-shot federated learning with diffusion models. In: 2025 IEEE/CVF Winter Conference on Applications of Computer Vision (WACV), pp 2601--2610, \doi{10.1109/WACV61041.2025.00258}

\bibitem[{Meng et~al.(2023)Meng, Rombach, Gao, Kingma, Ermon, Ho, and Salimans}]{meng2023distillation}
Meng C, Rombach R, Gao R, et~al (2023) On distillation of guided diffusion models. In: Proceedings of the IEEE/CVF conference on computer vision and pattern recognition, pp 14297--14306

\bibitem[{{Meta AI}(2024)}]{meta2024llama31}
{Meta AI} (2024) {Llama 3.1 Model Card}. \url{https://github.com/meta-llama/llama-models/blob/main/models/llama3_1/MODEL_CARD.md}, accessed: 2026-05-18

\bibitem[{Miles et~al.(2026)Miles, Toker, Oncescu, Xu, Deng, and Elezi}]{miles2026test}
Miles R, Toker A, Oncescu AM, et~al (2026) Test-time scaling with diffusion language models via reward-guided stitching. \urlprefix\url{https://arxiv.org/abs/2602.22871}, {\href{https://arxiv.org/abs/2602.22871}{{arXiv:2602.22871}}}

\bibitem[{Nie et~al.(2025)Nie, Zhu, You, Zhang, Ou, Hu, Zhou, Lin, Wen, and Li}]{nie2025large}
Nie S, Zhu F, You Z, et~al (2025) Large language diffusion models. In: Advances in Neural Information Processing Systems, pp 50608--50646

\bibitem[{OpenAI et~al.(2024)OpenAI, Achiam, Adler, Agarwal, Ahmad, Akkaya, Aleman, Almeida, Altenschmidt, Altman, Anadkat et~al.}]{achiam2023gpt}
OpenAI, Achiam J, Adler S, et~al (2024) {GPT-4} technical report. \urlprefix\url{https://arxiv.org/abs/2303.08774}, {\href{https://arxiv.org/abs/2303.08774}{{arXiv:2303.08774}}}

\bibitem[{Pan et~al.(2025)Pan, Lin, Liu, Liang, and Yuen}]{pan2025diffusion}
Pan H, Lin B, Liu Y, et~al (2025) Diffusion-model-enhanced multiobjective optimization for improving forest monitoring efficiency in {UAV}-enabled {Internet of Things}. IEEE Internet of Things Journal 12(19):40980--40996. \doi{10.1109/JIOT.2025.3590507}

\bibitem[{Park et~al.(2025)Park, Go, Kim, Woo, Ham, and Kim}]{Efficient_Model5}
Park B, Go H, Kim JY, et~al (2025) Switch diffusion transformer: Synergizing denoising tasks with sparse mixture-of-experts. In: Computer Vision -- ECCV 2024, Cham, pp 461--477, \doi{10.1007/978-3-031-43546-9_27}

\bibitem[{Peng et~al.(2025)Peng, Liu, Dong, Cheng, Li, Tang, Wang, and Zhao}]{peng2025efficient}
Peng H, Liu P, Dong Z, et~al (2025) How efficient are diffusion language models? {A} critical examination of efficiency evaluation practices. \urlprefix\url{https://arxiv.org/abs/2510.18480}, {\href{https://arxiv.org/abs/2510.18480}{{arXiv:2510.18480}}}

\bibitem[{Qin et~al.(2026)Qin, Dai, Lu, Shao, Wang, Xu, Zhang, Zhang, and Letaief}]{qin2025generative}
Qin HL, Dai J, Lu G, et~al (2026) Generative {AI} meets {6G} and beyond: Diffusion models for semantic communications. IEEE Communications Surveys \& Tutorials 28:6241--6281. \doi{10.1109/COMST.2026.3690542}

\bibitem[{Qin et~al.(2025)Qin, Wu, Du, and Huang}]{qin2025optimal}
Qin S, Wu H, Du H, et~al (2025) Optimal expert selection for distributed mixture-of-experts at the wireless edge. \urlprefix\url{https://arxiv.org/abs/2503.13421}, {\href{https://arxiv.org/abs/2503.13421}{{arXiv:2503.13421}}}

\bibitem[{Rawles et~al.(2025)Rawles, Clinckemaillie, Chang, Waltz, Lau, Fair, Li, Bishop, Li, Campbell-Ajala, Toyama, Berry, Tyamagundlu, Lillicrap, and Riva}]{8rawles2024androidworld}
Rawles C, Clinckemaillie S, Chang Y, et~al (2025) {AndroidWorld}: A dynamic benchmarking environment for autonomous agents. In: International Conference on Learning Representations, pp 406--441

\bibitem[{Reyes-Ram{\'i}rez et~al.(2024)Reyes-Ram{\'i}rez, Arag{\'o}n, S{\'a}nchez-Vega, and L{\'o}pez-Monroy}]{reyes2024improving}
Reyes-Ram{\'i}rez AD, Arag{\'o}n ME, S{\'a}nchez-Vega F, et~al (2024) Improving aggressiveness detection using a data augmentation technique based on a diffusion language model. In: Proceedings of the 8th Workshop on Online Abuse and Harms (WOAH 2024), pp 171--177, \doi{10.18653/v1/2024.woah-1.13}

\bibitem[{Rombach et~al.(2022)Rombach, Blattmann, Lorenz, Esser, and Ommer}]{11rombach2022high}
Rombach R, Blattmann A, Lorenz D, et~al (2022) High-resolution image synthesis with latent diffusion models. In: Proceedings of the IEEE/CVF conference on computer vision and pattern recognition, pp 10684--10695

\bibitem[{Saheed and Chukwuere(2026)}]{saheed2025autonomous}
Saheed YK, Chukwuere JE (2026) Autonomous {LLM} agent: A memory-augmented, edge-optimized {SHAP} explanations with zero-day attack resilience in {IoT}/industrial {IoT} networks. IEEE Internet of Things Journal 13(7):14213--14228. \doi{10.1109/JIOT.2025.3648649}

\bibitem[{Sahoo et~al.(2024)Sahoo, Arriola, Schiff, Gokaslan, Marroquin, Chiu, Rush, and Kuleshov}]{sahoo2024simple}
Sahoo SS, Arriola M, Schiff Y, et~al (2024) Simple and effective masked diffusion language models. Advances in Neural Information Processing Systems 37:130136--130184

\bibitem[{Salimans and Ho(2022)}]{salimans2022progressive}
Salimans T, Ho J (2022) Progressive distillation for fast sampling of diffusion models. \urlprefix\url{https://arxiv.org/abs/2202.00512}, {\href{https://arxiv.org/abs/2202.00512}{{arXiv:2202.00512}}}

\bibitem[{Shen et~al.(2025)Shen, Zhang, Xiong, Hu, Chen, Wan, Wang, Zhang, Gong, Bao, Tao, Huang, Yuan, and Zhang}]{shen2025efficient}
Shen H, Zhang J, Xiong B, et~al (2025) Efficient diffusion models: A survey. Transactions on Machine Learning Research

\bibitem[{Shi et~al.(2026)Shi, Wan, Dong, Jiang, Liu, and Huang}]{shi2026simpletool}
Shi X, Wan J, Dong L, et~al (2026) {SimpleTool}: Parallel decoding for real-time {LLM} function calling. \urlprefix\url{https://arxiv.org/abs/2603.00030}, {\href{https://arxiv.org/abs/2603.00030}{{arXiv:2603.00030}}}

\bibitem[{Song et~al.(2022)Song, Meng, and Ermon}]{song2020denoising}
Song J, Meng C, Ermon S (2022) Denoising diffusion implicit models. \urlprefix\url{https://arxiv.org/abs/2010.02502}, {\href{https://arxiv.org/abs/2010.02502}{{arXiv:2010.02502}}}

\bibitem[{Song et~al.(2021)Song, Sohl-Dickstein, Kingma, Kumar, Ermon, and Poole}]{Diffusion_Models5}
Song Y, Sohl-Dickstein J, Kingma DP, et~al (2021) Score-based generative modeling through stochastic differential equations. In: International Conference on Learning Representations

\bibitem[{Song et~al.(2023)Song, Dhariwal, Chen, and Sutskever}]{song2023consistency}
Song Y, Dhariwal P, Chen M, et~al (2023) Consistency models. In: Proceedings of the 40th International Conference on Machine Learning

\bibitem[{Sun et~al.(2024)Sun, Khalid, Mendieta, Wang, and Chen}]{sun2024exploring}
Sun G, Khalid U, Mendieta M, et~al (2024) Exploring parameter-efficient fine-tuning to enable foundation models in federated learning. In: 2024 IEEE International Conference on Big Data (BigData), pp 8015--8024, \doi{10.1109/BigData62323.2024.10825712}

\bibitem[{Sun et~al.(2026)Sun, Liu, Li, and Xiao}]{sun2026hillinfer}
Sun H, Liu S, Li L, et~al (2026) {HillInfer}: Efficient long-context {LLM} inference on the edge with hierarchical {KV} eviction using {SmartSSD}. \urlprefix\url{https://arxiv.org/abs/2602.18750}, {\href{https://arxiv.org/abs/2602.18750}{{arXiv:2602.18750}}}

\bibitem[{Sung et~al.(2025)Sung, Palakonda, Im, Moon, Kim, Yun, and Kang}]{sung2025memory}
Sung M, Palakonda V, Im S, et~al (2025) Memory- and latency-constrained inference of large language models via adaptive split computing. \urlprefix\url{https://arxiv.org/abs/2511.04002}, {\href{https://arxiv.org/abs/2511.04002}{{arXiv:2511.04002}}}

\bibitem[{Sánchez-Carballo et~al.(2026)Sánchez-Carballo, Melgarejo-Meseguer, and Rojo-Álvarez}]{sanchez2026latent}
Sánchez-Carballo E, Melgarejo-Meseguer FM, Rojo-Álvarez JL (2026) Latent diffusion for {Internet of Things} attack data generation in intrusion detection. \urlprefix\url{https://arxiv.org/abs/2601.16976}, {\href{https://arxiv.org/abs/2601.16976}{{arXiv:2601.16976}}}

\bibitem[{Tian et~al.(2024)Tian, Jiang, Yuan, Peng, and Wang}]{autoregressive_models3}
Tian K, Jiang Y, Yuan Z, et~al (2024) Visual autoregressive modeling: Scalable image generation via next-scale prediction. In: Advances in Neural Information Processing Systems, pp 84839--84865, \doi{10.52202/079017-2694}

\bibitem[{Tian et~al.(2025{\natexlab{a}})Tian, Liu, Hou, Qiu, Wang, Niyato, and Leung}]{11075534}
Tian M, Liu Z, Hou C, et~al (2025{\natexlab{a}}) Accelerating {AI}-generated content collaborative inference via transfer reinforcement learning in dynamic edge networks. IEEE Transactions on Cloud Computing 13(3):1011--1025. \doi{10.1109/TCC.2025.3586878}

\bibitem[{Tian et~al.(2025{\natexlab{b}})Tian, Yang, Yang, Wang, Tian, Zheng, Wang, Teng, Wang, Wang, Tong, Wang, and Li}]{tian2025mmada}
Tian Y, Yang L, Yang J, et~al (2025{\natexlab{b}}) {MMaDA-Parallel}: Multimodal large diffusion language models for thinking-aware editing and generation. \urlprefix\url{https://arxiv.org/abs/2511.09611}, {\href{https://arxiv.org/abs/2511.09611}{{arXiv:2511.09611}}}

\bibitem[{Touvron et~al.(2023)Touvron, Martin, Stone, Albert, Almahairi, Babaei, Bashlykov, Batra, Bhargava, Bhosale et~al.}]{autoregressive_models1}
Touvron H, Martin L, Stone K, et~al (2023) Llama 2: Open foundation and fine-tuned chat models. \urlprefix\url{https://arxiv.org/abs/2307.09288}, {\href{https://arxiv.org/abs/2307.09288}{{arXiv:2307.09288}}}

\bibitem[{Wang et~al.(2026)Wang, Wang, Cui, Li, Lu, Wang, Wang, Tang, Zhang, and Zhan}]{wang2026reflectdrive2}
Wang H, Wang Y, Cui B, et~al (2026) {ReflectDrive-2}: Reinforcement-learning-aligned self-editing for discrete diffusion driving. \urlprefix\url{https://arxiv.org/abs/2605.04647}, {\href{https://arxiv.org/abs/2605.04647}{{arXiv:2605.04647}}}

\bibitem[{Wang et~al.(2025{\natexlab{a}})Wang, Chen, Jiang, Pan, Cai, Yang, and Yang}]{wang2024parameter}
Wang L, Chen S, Jiang L, et~al (2025{\natexlab{a}}) Parameter-efficient fine-tuning in large language models: A survey of methodologies. Artificial Intelligence Review 58:227. \doi{10.1007/s10462-025-11236-4}

\bibitem[{Wang et~al.(2025{\natexlab{b}})Wang, Jiang, Yang, Liu, Li, Cao, and Liu}]{wang2025efficient}
Wang Q, Jiang S, Yang Y, et~al (2025{\natexlab{b}}) Efficient and adaptive diffusion model inference through lookup table on mobile devices. IEEE Transactions on Mobile Computing 24(9):8729--8746. \doi{10.1109/TMC.2025.3558203}

\bibitem[{Wang et~al.(2025{\natexlab{c}})Wang, Xu, Jin, Jin, Zhang, and Deng}]{wang2025diffusion}
Wang X, Xu C, Jin Y, et~al (2025{\natexlab{c}}) Diffusion {LLMs} can do faster-than-{AR} inference via discrete diffusion forcing. \urlprefix\url{https://arxiv.org/abs/2508.09192}, {\href{https://arxiv.org/abs/2508.09192}{{arXiv:2508.09192}}}

\bibitem[{Wei et~al.(2025)Wei, Tang, Zhao, and Yang}]{wei2025diffusionx}
Wei Y, Tang S, Zhao L, et~al (2025) {DiffusionX}: Efficient edge-cloud collaborative image generation with multi-round prompt evolution. \urlprefix\url{https://arxiv.org/abs/2510.16326}, {\href{https://arxiv.org/abs/2510.16326}{{arXiv:2510.16326}}}

\bibitem[{Wen et~al.(2025{\natexlab{a}})Wen, Zhu, Liu, Liu, Yang, Zhang, Zhang, Zhu, and Xu}]{wen2025dvla}
Wen J, Zhu M, Liu J, et~al (2025{\natexlab{a}}) {dVLA}: Diffusion vision-language-action model with multimodal chain-of-thought. \urlprefix\url{https://arxiv.org/abs/2509.25681}, {\href{https://arxiv.org/abs/2509.25681}{{arXiv:2509.25681}}}

\bibitem[{Wen et~al.(2025{\natexlab{b}})Wen, Li, Gu, Zhao, Wang, and Sun}]{wen2025llada_vla}
Wen Y, Li H, Gu K, et~al (2025{\natexlab{b}}) {LLaDA-VLA}: Vision language diffusion action models. \urlprefix\url{https://arxiv.org/abs/2509.06932}, {\href{https://arxiv.org/abs/2509.06932}{{arXiv:2509.06932}}}

\bibitem[{Wu et~al.(2025)Wu, Zhang, Xue, Liu, Diao, Zhu, Luo, Han, and Xie}]{wu2025fast}
Wu C, Zhang H, Xue S, et~al (2025) {Fast-dLLM}: Training-free acceleration of diffusion {LLM} by enabling {KV} cache and parallel decoding. \urlprefix\url{https://arxiv.org/abs/2505.22618}, {\href{https://arxiv.org/abs/2505.22618}{{arXiv:2505.22618}}}

\bibitem[{Wu et~al.(2026)Wu, Hu, Zhu, Liu, Zhang, Li, Chen, Pan, Xu, and Jin}]{Efficient_Model4}
Wu J, Hu M, Zhu J, et~al (2026) Evo: Autoregressive-diffusion large language models with evolving balance. \urlprefix\url{https://arxiv.org/abs/2603.06617}, {\href{https://arxiv.org/abs/2603.06617}{{arXiv:2603.06617}}}

\bibitem[{Wu et~al.(2023)Wu, Fan, Liu, Zheng, Gong, Jiao, Li, Guo, Duan, Chen et~al.}]{wu2023ar}
Wu T, Fan Z, Liu X, et~al (2023) {AR-Diffusion}: Auto-regressive diffusion model for text generation. Advances in Neural Information Processing Systems 36:39957--39974

\bibitem[{Xiao et~al.(2025)Xiao, Kantarci, Kang, Niyato, and Guizani}]{xiao2024efficient}
Xiao B, Kantarci B, Kang J, et~al (2025) Efficient prompting for {LLM}-based generative {Internet of Things}. IEEE Internet of Things Journal 12(1):778--791. \doi{10.1109/JIOT.2024.3470210}

\bibitem[{Xie et~al.(2023)Xie, Yao, Shi, Liu, Zhou, Liu, Li, and Li}]{Efficient_Training5}
Xie E, Yao L, Shi H, et~al (2023) {DiffFit}: Unlocking transferability of large diffusion models via simple parameter-efficient fine-tuning. In: Proceedings of the IEEE/CVF International Conference on Computer Vision (ICCV), pp 4230--4239

\bibitem[{Xie et~al.(2024)Xie, Zhang, Chen, Li, Zhao, Cao, Hua, Cheng, Shin, Lei et~al.}]{6xie2024osworld}
Xie T, Zhang D, Chen J, et~al (2024) {OSWorld}: Benchmarking multimodal agents for open-ended tasks in real computer environments. Advances in Neural Information Processing Systems 37:52040--52094

\bibitem[{Xie and Fang(2025)}]{xie2025energy}
Xie Y, Fang Q (2025) An energy-aware generative {AI} edge inference framework for low-power {IoT} devices. Electronics 14(20). \urlprefix\url{https://doi.org/10.3390/electronics14204086}

\bibitem[{Xiong et~al.(2025)Xiong, Wu, Yang, Xing, and Yeh}]{xiong2025diffguess}
Xiong H, Wu X, Yang X, et~al (2025) {DiffGuess}: Password strength assessment with diffusion models in intelligent vehicle-to-everything ({V2X}) communications. IEEE Transactions on Intelligent Transportation Systems pp 1--11. \doi{10.1109/TITS.2025.3623319}

\bibitem[{Xu(2024)}]{xu2024phd}
Xu C (2024) {PhD} forum abstract: Diffusion-based task scheduling for efficient {AI}-generated content in edge networks. In: 2024 23rd ACM/IEEE International Conference on Information Processing in Sensor Networks (IPSN), pp 333--334, \doi{10.1109/IPSN61024.2024.10697426}

\bibitem[{Xu et~al.(2026)Xu, Mu, Liu, Kim, and Nallanathan}]{xu2025large}
Xu X, Mu X, Liu Y, et~al (2026) Large model at edge: An optimal mobile edge generation ({MEG}) design. IEEE Transactions on Wireless Communications 25:5355--5369. \doi{10.1109/TWC.2025.3617845}

\bibitem[{Yan et~al.(2024)Yan, Liu, Liu, Peng, Wang, Chen, Fu, and Mei}]{yan2024hybridsdedgecloudcollaborative}
Yan C, Liu S, Liu H, et~al (2024) Hybrid {SD}: Edge-cloud collaborative inference for {Stable Diffusion} models. \urlprefix\url{https://arxiv.org/abs/2408.06646}, {\href{https://arxiv.org/abs/2408.06646}{{arXiv:2408.06646}}}

\bibitem[{Yan et~al.(2026)Yan, Zeng, and Hou}]{yan2025xdiff}
Yan P, Zeng H, Hou YT (2026) {xDiff}: Online diffusion model for collaborative inter-cell interference management in {5G} {O-RAN}. IEEE Transactions on Networking 34:1363--1376. \doi{10.1109/TON.2025.3622520}

\bibitem[{Yang et~al.(2023{\natexlab{a}})Yang, Zhang, Song, Hong, Xu, Zhao, Zhang, Cui, and Yang}]{1yang2023diffusion}
Yang L, Zhang Z, Song Y, et~al (2023{\natexlab{a}}) Diffusion models: A comprehensive survey of methods and applications. ACM Comput Surv 56(4). \doi{10.1145/3626235}

\bibitem[{Yang et~al.(2025{\natexlab{a}})Yang, Tian, Li, Zhang, Shen, Tong, and Wang}]{yang2025mmada}
Yang L, Tian Y, Li B, et~al (2025{\natexlab{a}}) {MMaDA}: Multimodal large diffusion language models. In: Advances in Neural Information Processing Systems, pp 138867--138907

\bibitem[{Yang et~al.(2025{\natexlab{b}})Yang, Xiong, Guo, Mao, Kim, and Debbah}]{yang2025efficient}
Yang W, Xiong Z, Guo S, et~al (2025{\natexlab{b}}) Efficient multi-user offloading of personalized diffusion models: A {DRL}-convex hybrid solution. IEEE Transactions on Mobile Computing 24(9):9092--9109. \doi{10.1109/TMC.2025.3560582}

\bibitem[{Yang et~al.(2023{\natexlab{b}})Yang, Zhou, Feng, and Wang}]{yang2023diffusion}
Yang X, Zhou D, Feng J, et~al (2023{\natexlab{b}}) Diffusion probabilistic model made slim. In: 2023 IEEE/CVF Conference on Computer Vision and Pattern Recognition (CVPR), pp 22552--22562, \doi{10.1109/CVPR52729.2023.02160}

\bibitem[{Yang and Chen(2025)}]{yang2025diffusion}
Yang Y, Chen Z (2025) Diffusion modeling meets deep reinforcement learning for edge task scheduling. In: 2025 7th International Conference on Next Generation Data-driven Networks (NGDN), pp 116--121, \doi{10.1109/NGDN66208.2025.11182116}

\bibitem[{Yang et~al.(2025{\natexlab{c}})Yang, Ma, Sun, He, Fu, Yuen, and Zhang}]{11193881}
Yang Y, Ma W, Sun W, et~al (2025{\natexlab{c}}) Diffusion-based multi-agent reinforcement learning for semantic vehicular edge computing. IEEE Transactions on Services Computing 18(6):3668--3681

\bibitem[{Yao et~al.(2025)Yao, Tang, Yang, and Jia}]{yao2025enhancing}
Yao Z, Tang Z, Yang W, et~al (2025) Enhancing {LLM} {QoS} through cloud-edge collaboration: A diffusion-based multi-agent reinforcement learning approach. IEEE Transactions on Services Computing 18(3):1412--1427. \doi{10.1109/TSC.2025.3562362}

\bibitem[{Ye et~al.(2025)Ye, Xie, Zheng, Gao, Wu, Jiang, Li, and Kong}]{ye2025dream7bdiffusionlarge}
Ye J, Xie Z, Zheng L, et~al (2025) Dream {7B}: Diffusion large language models. \urlprefix\url{https://arxiv.org/abs/2508.15487}, {\href{https://arxiv.org/abs/2508.15487}{{arXiv:2508.15487}}}

\bibitem[{You et~al.(2026)You, Nie, Zhang, Zhou, Lu, Wen, and Li}]{You_2026_CVPR}
You Z, Nie S, Zhang X, et~al (2026) {LLaDA-V}: Large language diffusion models with visual instruction tuning. In: Proceedings of the IEEE/CVF Conference on Computer Vision and Pattern Recognition (CVPR), pp 10093--10105

\bibitem[{Yu et~al.(2025{\natexlab{a}})Yu, Li, and Wang}]{yu2025discrete}
Yu R, Li Q, Wang X (2025{\natexlab{a}}) Discrete diffusion in large language and multimodal models: A survey. \urlprefix\url{https://arxiv.org/abs/2506.13759}, {\href{https://arxiv.org/abs/2506.13759}{{arXiv:2506.13759}}}

\bibitem[{Yu et~al.(2025{\natexlab{b}})Yu, Ma, and Wang}]{yu2025dimplediscretediffusionmultimodal}
Yu R, Ma X, Wang X (2025{\natexlab{b}}) Dimple: Discrete diffusion multimodal large language model with parallel decoding. \urlprefix\url{https://arxiv.org/abs/2505.16990}, {\href{https://arxiv.org/abs/2505.16990}{{arXiv:2505.16990}}}

\bibitem[{Zeng et~al.(2025{\natexlab{a}})Zeng, Hu, Song, and Song}]{zeng2025diffusion}
Zeng Q, Hu C, Song M, et~al (2025{\natexlab{a}}) Diffusion model quantization: A review. \urlprefix\url{https://arxiv.org/abs/2505.05215}, {\href{https://arxiv.org/abs/2505.05215}{{arXiv:2505.05215}}}

\bibitem[{Zeng et~al.(2025{\natexlab{b}})Zeng, Zheng, Gao, Niu, Ren, Wang, Cao, and Ji}]{zeng2025generative}
Zeng W, Zheng J, Gao L, et~al (2025{\natexlab{b}}) Generative {AI}-aided multimodal parallel offloading for {AIGC} metaverse service in {IoT} networks. IEEE Internet of Things Journal 12(10):13273--13285. \doi{10.1109/JIOT.2025.3535623}

\bibitem[{Zhao et~al.(2024)Zhao, Xu, Xiao, Jia, and Hou}]{zhao2024mobilediffusion}
Zhao Y, Xu Y, Xiao Z, et~al (2024) {MobileDiffusion}: Instant text-to-image generation on mobile devices. In: European Conference on Computer Vision, Springer, pp 225--242, \doi{10.1007/978-3-031-73033-7_13}

\bibitem[{Zheng(2025)}]{zheng2025diffusion}
Zheng D (2025) Diffusion models on the edge: Challenges, optimizations, and applications. \urlprefix\url{https://arxiv.org/abs/2504.15298}, {\href{https://arxiv.org/abs/2504.15298}{{arXiv:2504.15298}}}

\bibitem[{Zheng et~al.(2025)Zheng, Chen, Qian, Shi, Shu, and Chen}]{zheng2025review}
Zheng Y, Chen Y, Qian B, et~al (2025) A review on edge large language models: Design, execution, and applications. ACM Comput Surv 57(8). \doi{10.1145/3719664}

\bibitem[{Zhou et~al.(2025)Zhou, Yu, Babu, Tirumala, Yasunaga, Shamis, Kahn, Ma, Zettlemoyer, and Levy}]{Efficient_Training1}
Zhou C, Yu L, Babu A, et~al (2025) Transfusion: Predict the next token and diffuse images with one multi-modal model. In: International Conference on Learning Representations, pp 6446--6469

\bibitem[{Zhou et~al.(2022)Zhou, Jiang, Liu, Li, and Leung}]{zhou2021deep}
Zhou H, Jiang K, Liu X, et~al (2022) Deep reinforcement learning for energy-efficient computation offloading in mobile-edge computing. IEEE Internet of Things Journal 9(2):1517--1530. \doi{10.1109/JIOT.2021.3091142}

\bibitem[{Zhu et~al.(2024)Zhu, Li, Liu, Ma, and Wang}]{zhu2024survey}
Zhu X, Li J, Liu Y, et~al (2024) A survey on model compression for large language models. Transactions of the Association for Computational Linguistics 12:1556--1577. \doi{10.1162/tacl_a_00704}

\bibitem[{Zhu et~al.(2026)Zhu, Ren, Tan, and Ma}]{zhu2026esdllm}
Zhu Z, Ren F, Tan Z, et~al (2026) {ES-dLLM}: Efficient inference for diffusion large language models by early-skipping. In: The Fourteenth International Conference on Learning Representations

\end{thebibliography}

\end{document}